\documentclass[final,5p,times,twocolumn]{elsarticle}

\usepackage{graphicx}
\usepackage{amssymb}
\usepackage{caption}
\usepackage{subcaption}
\usepackage{booktabs}
\usepackage{lscape}
\usepackage{graphics}

\usepackage{array}
\usepackage[table,xcdraw]{xcolor}
\newcolumntype{L}[1]{>{\raggedright\arraybackslash\hspace{0pt}}m{#1}}
\newcolumntype{C}[1]{>{\centering\arraybackslash\hspace{0pt}}m{#1}}
\newcolumntype{R}[1]{>{\raggedleft\arraybackslash\hspace{0pt}}m{#1}}

\usepackage{siunitx}
\usepackage{float}
\usepackage{calc}
\usepackage{gensymb}

\usepackage{textcomp}

\newlength\heightfiga\newlength\heightcapa
\newlength\heightfigb\newlength\heightcapb
\newlength\heightfigc\newlength\heightcapc
\newlength\heightfig

\usepackage{lineno}

\usepackage{tocloft}

\usepackage[colorlinks=true]{hyperref}

\biboptions{authoryear}

\journal{Remote Sensing of Environment}

\begin{document}

\begin{frontmatter}

\title{From parcel to continental scale - A first European crop type map based on Sentinel-1 and LUCAS Copernicus in-situ observations}





\newcommand{\orcidauthorRD}{0000-0002-7326-7684} 
\newcommand{\orcidauthorGL}{0000-0002-2734-4538} 
\newcommand{\orcidauthorMV}{0000-0002-9103-7081} 
\newcommand{\orcidauthorAV}{0000-0002-4234-0586} 


\author{Rapha\"{e}l d'Andrimont $^{1,*}$, Astrid Verhegghen$^{1,*}$, Guido Lemoine $^{1}$, Pieter Kempeneers$^{1}$ ,  Michele Meroni$^{1}$ and Marijn van der Velde $^{1}$}
\address{ $^{1}$\quad European Commission, Joint Research Centre (JRC), Ispra , Italy \\
 $^{*}$ These authors contributed equally to this work.
}

\begin{abstract}
Detailed parcel-level crop type mapping for the whole European Union (EU) is necessary for the evaluation of agricultural policies in the European Union (EU). The Copernicus program, and Sentinel-1 (S1) in particular, offers the opportunity to monitor agricultural land at a continental scale and in a timely manner. However, so far the potential of S1 has not been explored at such a scale.
Capitalizing on the unique LUCAS 2018 Copernicus module, we present the first continental crop type map at 10-m spatial resolution for the European Union (EU) based on S1A and S1B Synthetic Aperture Radar observations for the year 2018. 
We combined time series of 10-day average VV and VH S1 sigma backscattering coefficients from January to end of July with crop type information from the LUCAS 2018 survey. 
Random forest classification algorithms are tuned for two strata (a northern continental and a southern Mediterranean) and a two-phase classification procedure that distinguishes three main vegetation classes in the first phase and 19 different crop types (or crop type groups) in the second phase.  Non-vegetation land cover classes are masked out using auxiliary information and areas above 1,000 m and with a steep slope are discarded.
We assess the accuracy of this EU crop map with three approaches. First, the accuracy is assessed with independent LUCAS core in-situ observations across the EU. Second, an accuracy assessment is done specifically for main crop types from farmers declarations (GSAA, geospatial aid application) from 6 EU member countries or regions totaling $>$3M parcels and 8.21 Mha. Finally, for the main crops, the crop areas derived by classification are compared to the subnational (NUTS 2) area statistics reported by Eurostat.
The overall accuracy for the map is reported as 80.3\% when grouping main crop classes and 76\% when considering all 19 crop type classes separately. 
Highest accuracies are obtained for rape and turnip rape with user and produced accuracies higher than 96\%. Crop specific and phenology driven features determine the classification accuracy throughout the growing season. We highlight confusion between common wheat and barley, mainly due to the crop structure and phenology. A confusion between sugar beet and maize is also observed and likely due to the early time constraint we imposed for the classification production, i.e. end of July. The correlation between the remotely sensed estimated and Eurostat reported crop area ranges from 0.93 (potatoes) to 0.99 (rape and turnip rape). 
Finally, we discuss how the framework presented here can underpin the operational delivery of in-season high-resolution based crop mapping for the EU. 
\end{abstract}


\end{frontmatter}







\renewcommand{\thetable}{Table \arabic{table}}
\renewcommand{\thefigure}{Fig. \arabic{figure}}
\renewcommand{\figurename}{}
\renewcommand{\tablename}{}


\section{Introduction}

The European Union (EU) is the largest global exporter of agri-food products, with exports reaching \texteuro 138 billion in 2018 \citep{agrifoodtrade2018}. Farmlands are an important feature of the EU landscape with about 42\% of the EU's land area dedicated to agriculture. At the same time, the EU's Common Agricultural Policy (CAP) makes up about 37\% of the European Commission's (EC) budget (\texteuro 58.8 billion in 2018). Besides underpinning resilient food production, the CAP could contribute to mitigate climate change, improve the sustainable management of natural resources, and preserve biodiversity and landscapes. This is in line with the European Green Deal, and the Farm to Fork and Biodiversity strategies\footnote{See the respective regulations COM/2019/640, COM/2020/381 and COM/2020/380.}. Updated EU-wide information on crop types is essential to evaluate the implementation of these agricultural policies. The EU's Copernicus space programme and technological advances provide the opportunity to transform the ways in which we monitor crops, agricultural practices, as well as CAP expenditure. Moreover, in order to forecast crop yields and production in the EU, independent and timely information is needed to support decisions regarding agricultural markets of cereals and other crops. \citep{van2019european}. 

The Copernicus Sentinel-1 (S1) mission adds unique features to our global Earth Observation (EO) capacity. The frequency of observations is guaranteed by the S1A and S1B polar-orbiting satellites acquiring all-weather (not affected by atmospheric conditions) C-band Synthetic Aperture Radar imagery. Here we exploit this capability for continental scale agricultural monitoring at 10-m resolution. The use of microwave detection for large scale crop type mapping remains underexploited, opposed to applications using optical imagery, e.g. \cite{belgiu2018sentinel}, \cite{immitzer2016first} and \cite{defourny2019near}. Recently, the use of high performance cloud computing on platforms such as Google Earth Engine (GEE, \cite{gorelick2017google}) has been demonstrated for detailed continental scale crop-type mapping. \cite{massey2018integrating} and \cite{teluguntla201830} used large collections of 30-m optical LANDSAT imagery to respectively classify crops over North America, and over China and Australia. 

Besides the recent developments in computational capacity to handle such large data streams and related algorithms, complementary \textit{in-situ} observations remain essential for training and validation purposes. In the EU, continental \textit{in-situ} data collection is organised through the Land Use/Cover Area frame Survey (LUCAS) \citep{gallego2010european,LUCAS:online}. In 2018, the Copernicus module - a survey protocol specifically tailored to EO - was introduced \citep{dandrimont-essd}, providing a unique set of polygons with homogeneous land cover (covering up to  0.52 ha) with land cover as well as crop specific information. Another source of \textit{in-situ} data is the parcel-level crop declarations by farmers in the context of the CAP, which are increasingly publicly available. 
Here, we bring these elements together to create the first validated crop type map for the EU at 10-m resolution. After reviewing the pertinent literature in detail, we present the specific objectives of this study.

\subsection{Crop mapping with Sentinel-1}

The SAR S1 constellation provides minimum consistent 6-day revisits over the EU since 2016. The negligible dependence of microwave backscatter on atmospheric conditions provides consistent calibrated time series that are particularly well-suited for automated processing and machine learning. The sensitivity of microwave backscattering to crop canopy structure \citep{dobson1981microwave,ulaby1981microwave,veloso2017understanding,d2020detecting,meroni2021comparing}, along with its sensitivity to soil surface structure and moisture content \citep{dobson1981microwave} makes it an ideal data source for crop mapping. One of the overall principles in electromagnetic scattering models of crop canopies is that (back)scattering depends on leaf size and shape, relative to the microwave wavelength, their orientation in space, relative to the polarization plane of the microwaves, and the water content of those leafs. Leaf geometry and orientation determines, to a large extent, the preferential (back)scattering direction. This explains the large differences in vertically structured crop canopies (e.g. cereals) versus broad-leaf crops (e.g. sugarbeet, potatoes). It also implies that backscattering changes are expected for phenological stages of the crop that cause structural change in the canopy (ear formation in cereals, fruit formation after flowering). The leaf water content relates to the effective dielectric properties of the canopy, which determines (back)scattering intensity. Vigorous, well-watered crop canopies are more effective scatterers than stressed or senescent crops, which scattering behavior may be modulated by simultaneous change in leaf orientation. Scattering of bare soil is primarily dependent on surface roughness and soil moisture content, which has parallels to canopy scattering because the relative size and orientation of the surface structure is important, as well as row direction of the soil cultivation and soil water content, which determines scattering effectivity via the dielectric properties. It follows that scattering of partially vegetated soil, e.g., at the crop emergence stage, is related to the ensemble of canopy and soil scattering effects and that variation over time is related to the relative contributions of each component. Polarization of the microwaves determines the (Fresnel) reflection coefficients of the canopy and soil material.

Synthetic Aperture Radar (SAR) has been used as a data source for crop type mapping in different studies \citep{mcnairn2004application,kenduiywo2018crop,clauss2018mapping}, as well as in synergistic combination with optical data \citep{van2018synergistic,gao2018crop,kussul2018crop,orynbaikyzy2020crop,chakhar2021improving}. Despite all the advantages of SAR, relatively few studies have focused on crop mapping based on S1 only. \citet{kussul2017deep} and \citet{van2018synergistic} derived a crop type map from a combination of S1 with optical data over Ukraine and Belgium respectively. Other local studies over the Chinese Fuyu City \citep{wei2019multi} and the French Camargue region \citep{ndikumana2018deep}, demonstrate the potential of using S1 for crop type mapping. The consistent availability of S1 observations is particularly relevant in machine learning contexts. Such consistency is difficult to achieve with optical Sentinel-2 data, due to unpredictable cloud cover gaps in a multi-annual monitoring setting. Early results with the use of S1 time series in a deep neural network approach \citep{lemoine2019dnn} demonstrated this successfully in a national EU crop mapping context.  

\subsection{In-situ data in the European Union}

There is growing interest in using LUCAS \textit{in-situ} data for land cover and land use research \citep{pflugmacher2019mapping, weigand2020spatial}. However, until the recent harmonization of the five past LUCAS surveys by \cite{d2020harmonised}, use of different LUCAS survey data remained difficult. In addition, the LUCAS protocol was designed for EU-wide standardized reporting of land cover and land use area statistics and not for Earth Observation (EO) applications.  

Past remote sensing studies that incorporated LUCAS core data have been limited to statistical area estimates \citep{gallego2010european}. The LUCAS sampling frame was used to assess availability of crowd-sourced pictures potentially relevant for agriculture across the EU \citep{d2018crowdsourced}. Little experience exists in using LUCAS data for large scale land cover generation processes, especially on the usability of the LUCAS data set as training data base for supervised classification approaches \citep{mack2017semi}. The 2015 survey was used in several land cover and land use mapping exercises. \cite{Close2018} provided a Sentinel-2 LUCAS-based classification over southern Belgium in the context of Land Use, Land-Use Change, and Forestry (LULUCF) monitoring. \citet{pflugmacher2019mapping} demonstrated the potential of using LUCAS survey 2015 to map pan-European land cover (13 classes) with Landsat data. \cite{weigand2020spatial} showed the suitability of the LUCAS 2015 survey as reference information for high resolution land cover mapping using Sentinel-2 imagery at the scale of Germany (7 classes).

In 2018, a new LUCAS module (the Copernicus module) specifically tailored to EO was introduced, see \cite{dandrimont-essd}. A specific protocol was designed to collect \textit{in-situ} information with specific characteristics fitting EO processing requirements. As a result, a total of 58,428 polygons are provided with a level-3 land cover (66 specific classes including crop type) and land use (38 classes) information. This represent a unique set of data opening the possibility for applications with higher thematic detail as compared to previous LUCAS surveys, such as crop type mapping.

\subsection{Objectives}
The overall aim of this study is to assess the potential of combining the LUCAS Copernicus module with S1 data to generate an EU-wide crop type map at 10-m spatial resolution for crops in the year 2018. Beyond the production of the map, this study aims to set a benchmark for continental crop type mapping at 10-m spatial resolution.
More specifically, we address the following research objectives:
\begin{itemize}
\setlength\itemsep{0em}
    \item To highlight the untapped potential of the LUCAS Copernicus survey as training data for land monitoring applications in general and crop type mapping in particular;
    \item To demonstrate the value of S1 data for continental crop type map production;
    \item To evaluate the effectiveness of various S1 indices and time periods for the discrimination of specific crop types; 
    \item To benchmark accuracies against independent LUCAS core point data and EU farmers' declarations at parcel level;
    \item To determine the crop specific evolution of mapping accuracy throughout the growing season relevant for prospective operational tasks;
    \item To assess crop surface areas derived from the 10-m crop type map against reported EU statistics at subnational level. 
    
\end{itemize}



\section{Materials and Methods}
\label{sec:method}

\begin{figure}[!!!ht]
 \centering 
        \includegraphics[]{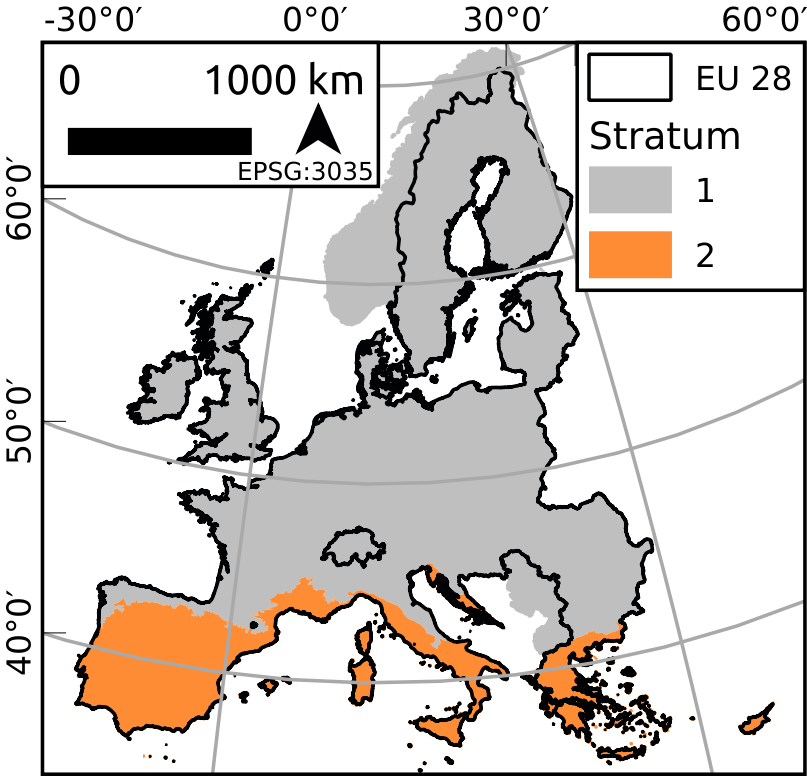}
        \caption{The study area is the European Union 28 (EU-28). The study area is divided in 2 strata based on a combination of biomes from \cite{dinerstein2017ecoregion} as described in Section \ref{sec:mehtod-hierachical-classif}.}
        \label{fig:study_area}
\end{figure}

The study area is the EU-28 as in 2018, covering \num{4469169} km$^2$ (\ref{fig:study_area}). A 10-km buffer was applied around the EU-28 to avoid mapping issues at the border.

\begin{figure*}[htb]
        \centering
\includegraphics[trim=50 200 50 0,clip,width=\textwidth]{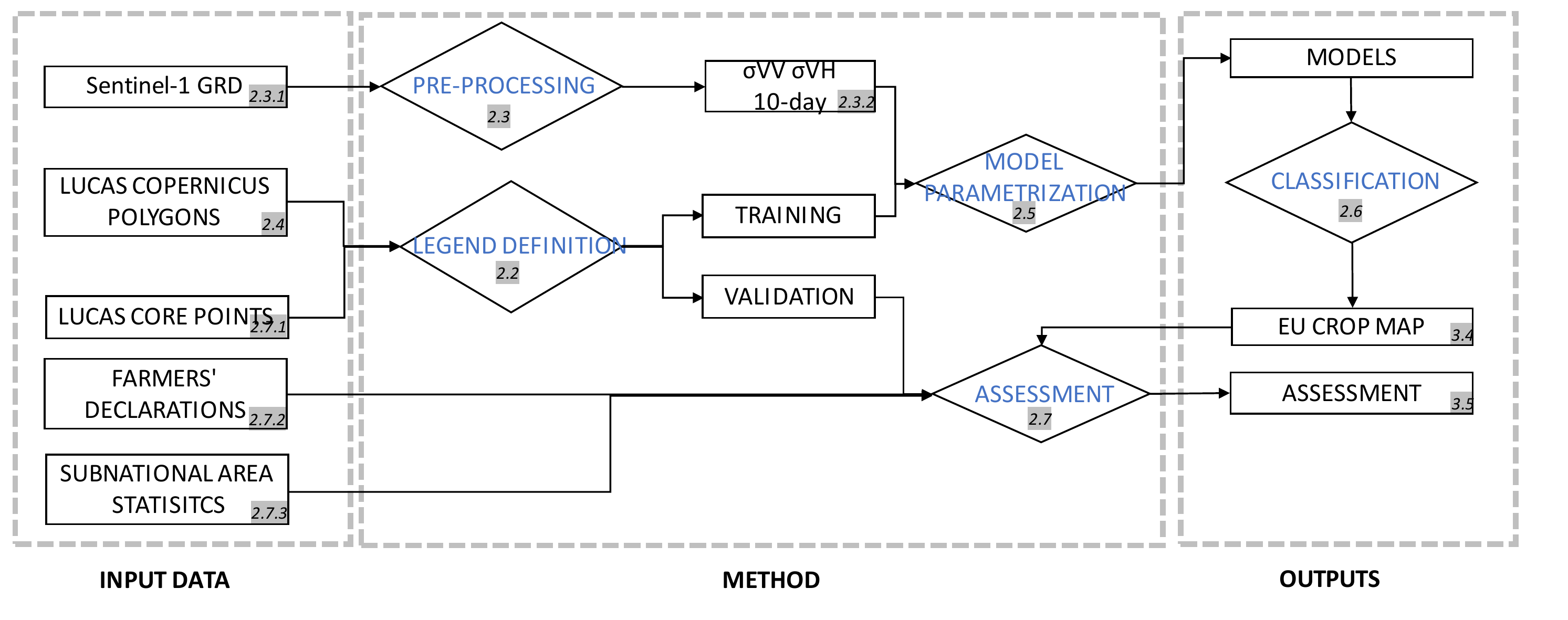} 
    \caption{General overview (references to section number are highlighted in grey). }
    \label{fig:overview}
\end{figure*}

 \ref{fig:overview} provides a flow chart illustrating the general approach and specific methodological steps of the study. 
 First, the input data is detailed: S1 in section \ref{sec:method_3_S1} and \textit{in-situ} data in Section \ref{sec:method_4_LUCAS}. Then, Section \ref{sec:method_2_legend} defines the legend used for the study. While Section \ref{sec:method_5_crop_type_classif_benchmarking} sets out how the features and the parameters for the crop classification models are selected, Section \ref{sec:method_6_image_classif} describes how the classification is applied at the continental scale. Finally, Section \ref{sec:method_7_accuracy_assessment} presents the three accuracy assessment approaches.

\subsection{Sentinel-1 data}
\label{sec:method_3_S1}

The S1 C-band operates at a central frequency of 5.404 Ghz in the microwave portion of the electromagnetic spectrum which corresponds to a wavelength of 5.55 cm. S1 acquisitions over the European land mass are in the so-called interferometric wave (IW) mode, which registers the backscattering of a vertically transmitted microwave signal in a vertical and horizontal receiver, creating a VV and VH polarized band, respectively. 

Over Europe, S1 acquires data both in descending and ascending orbits, as it does not rely on solar illumination (as opposed to optical sensors). This leads to a doubling of the revisit cycle, though from two distinct sensor viewing configurations due to different azimuth and look directions. Furthermore, the S1 swath width is designed to guarantee gap free coverage at the equator. At higher latitudes, this design leads to considerable overlap between adjacent orbits, and an effective higher revisit, more than daily at latitude above 55 degrees North.

\subsubsection{Geocoded backscattering coefficients} 

 Application ready S1 data are accessed on the GEE platform. On GEE, S1 data are processed from the GRD Level-1 product with the S1 Toolbox to remove thermal noise and to create geocoded and radiometrically calibrated backscattering coefficients ($\sigma^0$) at 10-m pixel spacing. 

Due to the side-looking nature of the SAR sensor, the incidence angle varies over the swath in the range direction, for S1 IW mode between 25-40 degrees. Normalization of the incidence effect is normally desired, and simple corrections can be applied to convert $\sigma^0$ to $\gamma^0$. A more robust approach would be to apply a correction that takes the local terrain height variation into account, known as terrain flattening (\cite{small2011flattening}), which can also partially compensate for the look angle differences in ascending and descending orbit acquisitions. However, terrain flattening is not applied in the GEE processing workflow and our analysis is thus based on $\sigma^0$ as most of the crops are typically on flat areas and the steep slopes are masked out as described in Section \ref{sec:method_6_image_classif}.


To be ingested in the machine learning classification algorithm, a spatially and temporally consistent time series needs to be derived from the S1 VV and VH $\sigma^0$ scenes. 
In addition, the backscattering coefficient ratio VH/VV (cross polarization ratio, CR) is computed from the VV and VH scenes. Several studies highlighted the advantages of the CR for mapping and monitoring of crops (\cite{veloso2017understanding, meroni2021comparing}).
We proceeded as follows to generate the required archive:

\begin{enumerate}

\item  \textit{Edge masking}.
The edge of each scene should be masked before the compositing. This is done by masking groups of neighboring pixels with values lower than -25 dB in the VV polarization (as described in \cite{d2018targeted}). 

\item  \textit{Averaging over a 10-day period}.
  $\sigma^0$ values are then averaged over subsequent time periods of 10 days for each pixel and for all ascending and descending acquisitions available in that period, separately for the VV and VH polarizations. The averaged $\sigma^0$ value is converted to decibels (dB). 

\item  \textit{Computing the cross polarization ratio}.
The CR is computed for each scene and averaged to the same 10-days periods.

\end{enumerate}

In this way, we generate a regularly timed, gridded, equal sized set of temporal features for VV, VH and CR to use as input for the subsequent machine learning runs (n = 36 10-day composites over the period 2018-01-01 to 2018-12-31), independent of the number of actual S1 acquisitions in the 10-day period. The S1 data were processed in GEE and then downloaded.

\subsection{LUCAS 2018 in-situ data}
\label{sec:method_4_LUCAS}

This section describes the LUCAS 2018 survey data that were used for training the classification models and assessing the accuracy of the crop type map. 

\subsubsection{LUCAS 2018 survey}

The LUCAS 2018 survey collected 97 variables at \num{337854} core points. Most of the points surveyed fall in a homogeneous area for which the minimum mapping unit is about 7 m$^2$(a circle of 1.5 m radius). When the land cover is not homogeneous, for example when it is composed of trees or shrubs interspersed with grass, the scale of observation is extended to classify it. In these cases, a systematic observation of the “environment” in the vicinity of the point, which in LUCAS is called the extended window of observation, has to be adopted. The extended window of observation expands to a radius of 20 meters from the point (representing an area of 0.13 ha) for forest and shrublands. Detailed information about the survey can be found in \cite{c12018}. The land cover surveyed is classified according to a harmonized 3-level legend system \cite{c32018}. Additionally to the \textit{core} variables collected, other specific modules were carried out on demand on a subset of points, such as (i) the topsoil module and (iii) the grassland module.
The LUCAS 2018 \textit{core} data are available in an harmonised open database in \cite{d2020harmonised}.

\subsubsection{LUCAS Copernicus module}

The LUCAS 2018 Copernicus module was applied to a subset of points to collect the land cover extent up to 51 meters in four cardinal directions around a point of observation, offering \textit{in-situ} data compatible with the spatial resolution of high-resolution sensors (see \cite{dandrimont-essd} for the open ready-to-use dataset). The LUCAS Copernicus dataset contains \num{63287} polygons at level-2. When filtering the data for which a level-3 legend is available,  \num{58426} polygons with a level-3 land cover (66 specific classes including crop type) and land use (38 classes) are available. The minimum mapping unit (MMU) of the in-situ data is 78.53 m² (i.e. a circle of 5-m radius) as the Copernicus module survey is not executed for smaller areas. The data from the LUCAS Copernicus module is used to train the classification models.

\subsection{Legend definition}
\label{sec:method_2_legend}
 
The LUCAS survey legend is re-organised to serve the purpose of the study (for details of the LUCAS legend classification see \cite{c32018} and Supplementary material). As the objective of the study is to classify the main EU-28 crop types based on LUCAS \textit{in-situ} data, the class definitions differ slightly from the LUCAS legend. The four vegetation classes of LUCAS (B-cropland, C-woodland, D-shrubland and E-grassland) are organized here in three main classes: arable land, woodlands and shrubland, and grasslands. The arable class is further divided in 19 specific classes of crop types or crop groups. The final legend is presented in \ref{tab:legend}. 

The main differences with the LUCAS legend are the following. The woodlands and shrublands classes are regrouped in a single class. The LUCAS cropland (B) class is re-organized to keep the arable land classes separated from the grasslands and permanent crops. First, the temporary grasslands (B55) are grouped with the grasslands class (E). Second, the permanent crops (B70 and B80) are grouped with the woody vegetation class. Third, the bare arable land (F40) with an agricultural land use (U111/112/113) is regrouped with the arable land classes. The grassland class encompasses grasslands from natural and agricultural land uses. 

\begin{table*}[htb]
\caption{The legend of the EU crop map. For the two-phase hierarchical classification procedure, the legend is divided into two levels: level 1 with seven broad land cover classes, and level 2 containing 19 crop types or crop groups.}
\label{tab:legend}
\footnotesize
\centering
\begin{tabular}{clclll}
\hline
\textbf{Level 1} &                  & \textbf{Level 2} &                                     &                      &  \\ \hline
100 &
   &
   &
  Artificial land &
  \begin{tabular}[c]{@{}l@{}}A11, A12, A13, A21, \\ A22\end{tabular} &
   \\ \hline
200             &                  &                 & Arable land                         & See below            &  \\ \hline
                & Cereals          & 211             & Common wheat                        & B11                  &  \\
                &                  & 212             & Durum wheat                         & B12                  &  \\
                &                  & 213             & Barley                              & B13                  &  \\
                &                  & 214             & Rye                                 & B14                  &  \\
                &                  & 215             & Oats                                & B15                  &  \\
                &                  & 216             & Maize                               & B16                  &  \\
                &                  & 217             & Rice                                & B17                  &  \\
                &                  & 218             & Triticale                           & B18                  &  \\
                &                  & 219             & Other cereals                       & B19                  &  \\ \cline{2-6} 
                & Root crops       & 221             & Potatoes                            & B21                  &  \\
                &                  & 222             & Sugar beet                          & B22                  &  \\
                &                  & 223             & Other roots crops                   & B23                  &  \\ \cline{2-6} 
 &
  \begin{tabular}[c]{@{}l@{}}Non permanent \\ industrial crops\end{tabular} &
  230 &
  \begin{tabular}[c]{@{}l@{}}Other non permanent \\ industrial crops\end{tabular} &
  B34, B35, B36, B37 &
   \\
                &                  & 231             & Sunflower                           & B31                  &  \\
                &                  & 232             & Rape and turnip rape                & B32                  &  \\
                &                  & 233             & Soya                                & B33                  &  \\ \cline{2-6} 
 &
  \begin{tabular}[c]{@{}l@{}}Dry pulses, \\ vegetables and flowers\end{tabular} &
  240 &
  Dry pulses, vegetables and flowers &
  \begin{tabular}[c]{@{}l@{}}B41, B42, B43, B44, \\ B45\end{tabular} &
   \\ \cline{2-6} 
 &
  Fodder crops &
  250 &
  \begin{tabular}[c]{@{}l@{}}Other fodder crops \\ (excl. temp. grasslands)\end{tabular} &
  B51, B52, B53, B54 &
   \\ \cline{2-6} 
                & Bare arable land & 290             & Bare arable land                    & F40*                 &  \\ \hline
300 &
   &
   &
  \begin{tabular}[c]{@{}l@{}}Woodland and shrubland \\ type of vegetation\end{tabular} &
  \begin{tabular}[c]{@{}l@{}}B71-B77, B81-B84, \\ C10, C21, C22, C23, \\ C31, C32, C33, D10, \\ D20\end{tabular} &
   \\ \hline
500             &                  &                 & Grassland (permanent and temporary) & B55, E10, E20, E30   &  \\ \hline
600             &                  &                 & Bare land and lichens/moss          & F10, F20, F30, F40** &  \\ \hline
\end{tabular}

\tiny
\centering
\textit{*U111/112/113 (agriculture),
**other than U111/U112/U113}

\end{table*}

\subsection{Crop type classification}
\label{sec:method_5_crop_type_classif_benchmarking}
This section describes the training data (input reference data and classification features) selection and the parametrization of the classification models used for the continental crop type mapping.

\subsubsection{Reference training data}
The classifiers are trained with the \textit{in-situ} information collected in the LUCAS Copernicus module.
The polygons corresponding to the LUCAS labels mentioned in \ref{tab:legend} are selected and reclassified according to the legend of the study. In total, \num{58,178} polygons from the \num{63,287} Copernicus polygons available are selected to train the classifier. 

\subsubsection{Stratification}
 
To take into account climatic and ecological spatial gradients, the EU-28 is stratified in two strata based on the main biomes defined by \cite{dinerstein2017ecoregion}. One main continental northern stratum and one southern Mediterranean stratum (see map in \ref{fig:study_area}):
\begin{itemize}
\setlength\itemsep{0em}
  \item Stratum 1 (Str1): Temperate Broadleaf \& Mixed Forests and Temperate Grasslands, Savannas \& Shrublands, Boreal Forests/Taiga  and  Tundra, Temperate Conifer Forests 
  \item Stratum 2 (Str2): Mediterranean Forests, Woodlands \& Shrub
\end{itemize}

 
More detailed stratification schemes (e.g. country- or a regional-level) have been adopted in previous crop type mapping exercises in Europe \citep{defourny2019near,Wang2019}. However, this is feasible where crop type in-situ data are abundant, e.g. provided by national crop data survey. In this study, we had to face  a trade-off between stratification detail and sample size available for the resulting strata and we opted for a simple and broad stratification.

\subsubsection{Two-phase classification}
\label{sec:mehtod-hierachical-classif}

For each stratum, a two-phase classification procedure is followed. The first step distinguishes between five broad land cover classes: Artificial land, Arable land, Woodland and shrubland type of vegetation, Grassland and Bare land (Level 1 in \ref{tab:legend}). In a second step, the Arable land class is further classified into 19 different crop types or crop groups (Level 2 in \ref{tab:legend}). Therefore, a total of four supervised classification (2 strata x 2 hierarchical levels) models are trained.

For the supervised classifications, we trained Random forests (RF) models on the S1 time series to map level 1 and 2 classes. RF is an ensemble machine learning method known to be robust against multi-colinearity and overfitting \citep{breiman2001random} and currently is a standard classification algorithm in the remote sensing domain \citep{belgiu2016random}. RF was reported to yield classification accuracies comparable to more sophisticated algorithms such as support vector machines, but with a much lower computational complexity \citep{inglada2017operational}. RF models are also stable with respect to the choice of parameters \citep{pelletier2016assessing} and easy to implement, making them suited for operational processing chains (such as Sen2-Agri \citep{defourny2019near}).

\subsubsection{Features selection}
\label{sec:Features_selection}


In order to produce a single crop type map at the continental map, the most significant features in term of S1 backscatter and time period are selected. To do so, different combinations of the 10-day averaged S1 indices, and different time periods from January to December are evaluated using the training reference dataset. 

The first part of the features selection evaluates, throughout the growing season period, the progression of the overall accuracy (O.A.) of the classifications and the F-score for the main crop types. The accuracy is evaluated using hindcasting with monthly time step from January to December, thus extending the analysed period beyond the end of the average European time of harvest. For the two classification levels and in the two strata, RF models are trained for twelve different time period using a 80/20 split of the reference (polygon) training data. The resulting O.A. and F-score are evaluated at the continental and the stratum level to select the best time period. From the analysis presented in section \ref{sec:results-features selection}, we consider the period from January to July (included) as a good trade-off between accuracy and timeliness. The period of January to end of July is used for all further processing.

Second, the performance of the two S1 $\sigma^0$ (VV and VH), one index (VH/VV), and their combinations are evaluated in terms of overall accuracy over the period January to end of July (included) . 


\subsection{Classification at scale}
\label{sec:method_6_image_classif}
A final map is produced at the scale of EU-28 using the best features defined by the previous analysis. This final map derived with S1 information from January to the end of July is referred to as the EU crop map in the manuscript.

The RF models are trained for the two strata with the best features in term of indices and time period.


A random search cross validation is used to define the best hyperparameters for each of the four RF models (level-1 and level-2 for each strata). We are tuning seven parameters (\ref{tab:appendix_Hyperparameters}) and define a grid of ranges, and randomly sample from this grid of possible parameter values to perform a 3-fold cross-validation with 100 possible combination of values, totalling 300 fits. Among the seven parameters, two are considered as key parameters: the number of decision trees (Ntrees) and the number of features to consider when looking for the best split when growing the trees (Mtry) \cite{belgiu2016random}. For Ntrees, we defined a range from 300 to 1200 with a step size of 100 and and for Mtry, we tested two options: the square root of the number of features and the binary logarithm of the number of features.

The RF classification algorithms and the hyperparameter tuning is implemented using the Python's scikit-learn (\cite{scikit-learn}) packages RandomForestClassifier and RandomizedSearchCV. The final hyperparameters used for each RF model are summarized in the supplementary materials (\ref{tab:appendix_Hyperparameters}).

The classification was performed at the continental scale on the JRC Big Data Analytics Platform (BDAP) using an HTCondor environment (\cite{Soille2018}). 


The final map is further processed for distribution. The main focus of the map being the arable land, areas located at an altitude above 1000m and on a slope of more than  10 $^{\circ}$ is masked out. The ALOS Global Digital Surface Model "ALOS World 3D - 30m (AW3D30)" is used to mask steep slopes. In addition, classes that were poorly represented in the LUCAS Copernicus module and consequently poorly classified (i.e. built-up, water, wetlands, and bare lands classes) are masked out with auxiliary products. Built-up areas are masked with the JRC GHSL European Settlement Map for 2015 at 10 m \citep{corbaneESR,sabo2020}. Water is masked with the permanent water from the 2019 version of the JRC Global Surface Water product\citep{Pekel2016}. Then, a number of classes are masked with the Corine Land Cover 2018 map \citep{eea2018corine}  (inland and coastal wetlands, bare land, moors and heathlands) see \ref{tab:Appendix_Masking} for details.


The EU crop map is re-projected to the Lambert azimuthal equal area (ETRS89-LAEA, EPSG:3035) projection for the analysis.

\subsection{Accuracy assessment}
\label{sec:method_7_accuracy_assessment}

Three approaches to estimate the accuracy of the EU crop map are tested. The first approach is taking as reference data the high quality LUCAS core points not surveyed by the Copernicus module. The second approach is comparing the EU crop map with a selection of GSAA data based on farmers' declarations. The third approach compares the area of several main crops obtained from the EU crop map to the corresponding official subnational statistics.

\subsubsection{LUCAS 2018 core points}

In the first approach, validation was focused on all the vegetation classes (i.e. crop types, woodland and shrubland, and grassland). To obtain data relevant for the assessment, LUCAS core points were filtered to keep only high quality information. While the LUCAS core points were not initially designed for EO analysis, \cite{weigand2020spatial} showed that the data can be used in land cover applications although no detailed analysis for crop type mapping applications were done to date. We therefore defined four criteria to select only high-quality points suitable for crop type mapping applications: keep only direct and in-situ observations, remove parcels smaller than 0.1 ha and only keep data with an homogeneous land cover class.  
In addition, the subset of (63,364) core points also surveyed for the Copernicus module and used for training were not included in this validation set. This screening resulted in a total of 87,853 LUCAS core points covering the selected vegetation classes (spatial distribution overview in Figure~\ref{fig:ValidationPOint}). 

The reference data is used in combination with the EU crop map to report the confusion matrix, the overall, user, and producer accuracy. LUCAS is a two-phase sampling scheme. The LUCAS 2018 survey followed a stratified random sample design \citep{scarno2018lucas}. The first-phase is a systematic sampling scheme and we treat the second-phase, the 2018 samples, as collected under a stratified random sampling, due to the large amount of collected data. Accordingly, the accuracy metrics are reported based on the estimated area proportion of correctly and mis-classified classes according to the equations of \cite{Olofsson2014a} 
for a 95\% confidence level. The accuracy metrics are provided first for a set of grouped crop classes (\ref{tab:legend}) and second considering all the classes.



\subsubsection{Comparison with GSAA}

In the second approach, a subset of the classes are assessed and compared wall-to-wall with vector parcels from the GSAA for specific regions. The GSAA refers to the annual crop declarations made by EU farmers for CAP area-aid support measures. The electronic GSAA records include a spatial delineation of the parcels. A GSAA element is always a polygon of an agricultural parcel with one crop (or a single crop group with the same payment eligibility). The GSAA is operated at the region or country level in the EU-28, resulting in about 65 different designs and implementation schemes over the EU. Since these infrastructures are set up in each region, at the moment data is not interoperable, nor are legends semantically harmonized. Furthermore, most GSAA data is not publicly available, although several countries are increasingly opening the data for public use. In this study, six regions with publicly available GSAA are selected representing a contrasting gradient across the EU (\ref{tab:LPIS_summary}). The selected regions along with the acronym used in this study are: bevl2018 (Flanders in BE), dk2018 (DK), frcv2018 (Centre - Val de Loire in FR), nld2018 (NL), nrw2018 (North Rhine-Westphalia in DE)) and si2018 (SI). The six selected regions with GSAA data have comparable sizes, ranging from 401,833 parcels in Flanders to 659,302 parcels in the Netherlands, facilitating the comparison. As each national GSAA records set can contain more than 300 distinct classes, only the parcels for crop classes covering at least 1\% cumulative GSAA area of a given region were selected (see \ref{fig:LPIS_distribution_area} for the distribution and \ref{fig:LPIS_distribution_count} for the count distribution). Excluding marginal classes with percentage cover smaller than 1\% results in a coverage of the total area ranging from from 84\% (Denmark) to 92\% (Slovenia) of the total parcels for each respective GSAA. In total, 3,149,334 parcels were extracted covering 8.21 Millions of hectares. Of our pixel-based classification predicted classes within each GSAA parcels, we extracted the class mode and compared it to the declared crop, mapped to the LUCAS legend (semantic mapping in \ref{tab:GSAA-LUCAS_legendMatching}). See also the Discussion Section \ref{sec:discussion_recommendation} about semantic harmonization of GSAA. 

 For each of the selected parcels, the majority predicted class was computed for the enclosed pixel ensemble and then compared to the declared crop label, mapped to the LUCAS legend. Then Producer Accuracy (PA) and User Accuracy (UA) were computed. The extraction of the data was implemented with the Python "Checks by Monitoring" (CbM) package (\url{https://pypi.org/project/cbm/}). This package provides Big Data Analytics routines for parcel crop monitoring using the Sentinel sensors and cloud compute infrastructure.

\begin{table*}[htb]
\centering
\caption{Summary of GSAA data used in the study.} 
\label{tab:LPIS_summary}
\footnotesize
\begin{tabular}{lC{4.5cm}C{3cm}C{3cm}C{3cm}}
  \hline
\textbf{Region} & \textbf{Description}  & \textbf{Area (Mha)} & \textbf{Area Ratio of the total (\%)} & \textbf{Parcels (\#)} \\ 
  \hline
 \textbf{bevl2018} & Flanders (Belgium) & 0.59 & 85 & 409,884 \\ 
  \textbf{dk2018} & Denmark & 2.23 & 84 & 445122 \\ 
  \textbf{frcv2018} & Centre - Val de Loire (France) & 1.95 & 85 & 336,212 \\ 
  \textbf{nld2018} & Netherlands & 1.62 & 87 & 659,302 \\ 
  \textbf{nrw2018} & North Rhine-Westphalia (Germany) & 1.39 & 91 & 613,012 \\ 
  \textbf{si2018} & Slovenia & 0.43 & 92 & 685,802 \\
   \hline
   \textbf{Total} &  & 8.21 & - & 3,149,334 \\
   \hline
\end{tabular}

\end{table*}

\subsubsection{Comparison with reported subnational statistics}

In the third and final validation approach, the area of the crop obtained from the EU crop map is compared to the official subnational statistics for the main crops in EU-28. In order perform do the comparison, subnational area data is obtained from the "Area (cultivation/harvested/production) (1000 ha)"\footnote{From the tab of the APRO\_CPSHR table reporting on crop production in EU standard humidity. Data is downloaded  from the Eurostat Data Browser portal (\url{https://ec.europa.eu/eurostat/databrowser/view/APRO_CPSHR/default/table}). Detailed description along with metadata can be found in \url{https://ec.europa.eu/eurostat/cache/metadata/en/apro_cp_esms.htm}.} by NUTS 2 regions. The legend of the Eurostat crop classes is then semantically matched with the EU crop map legend as described in \ref{tab:legend_convergence_Estat}.

The areas estimated from the EU crop map are obtained with a zonal histogram per administrative zone and then compared to the one reported by Eurostat at the finest spatial detail available (i.e. NUTS2 regions for most of the EU-28 except for the UK and DE where only NUTS1 is reported). For each crop with enough data available, the Pearson correlation coefficient is calculated. Additionally, the relative percentage difference in area is computed for each administrative region and mapped, highlighting the difference between the reported and here estimated surface area for each crop. The distribution of these differences for each crop is finally summarized to draw conclusion at the crop level. 



\section{Results}

\subsection{Features selection}
\label{sec:results-features selection}

\begin{figure*}[ht]
        \centering
    \includegraphics[width=0.8\textwidth]{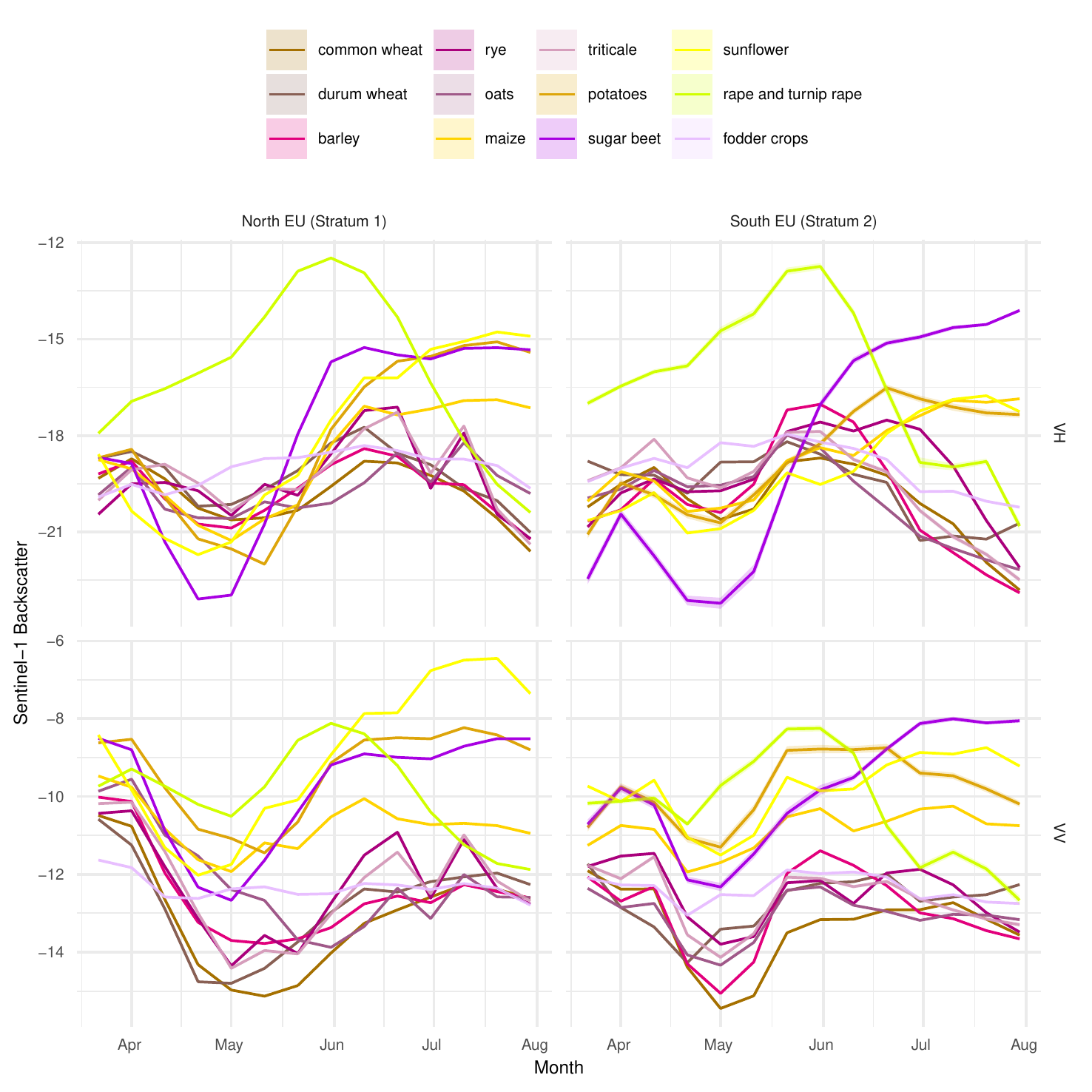}
    \caption{Evolution of VV and VH backscatter for the growing season for the 12 main crops. The value are averaged per stratum. The ribbon around the line corresponds to one standard deviation. }
    \label{fig:s1_time_serie}
\end{figure*}

The reference data consists of S1 time series retrieved from \num{58178} polygons derived from the LUCAS Copernicus polygons corresponding to 2,296,889 S1 10-m pixels (\ref{tab:LUCAScopernicus_per_stratum}). The training data are distributed into 23 thematic classes and two geographical strata (Str1 and Str2) as described in the methodology. Each of the pixels thus have time-series of synthesized 10-day - or 36 dekads per year - with VV, VH and VH/VV data. The VV and VH temporal $ \sigma^0 $ signatures for the 12 main crops are averaged per stratum and presented in \ref{fig:s1_time_serie} for the crop growing season.

\begin{figure}[ht]
        \centering
\includegraphics[width=0.45\textwidth]{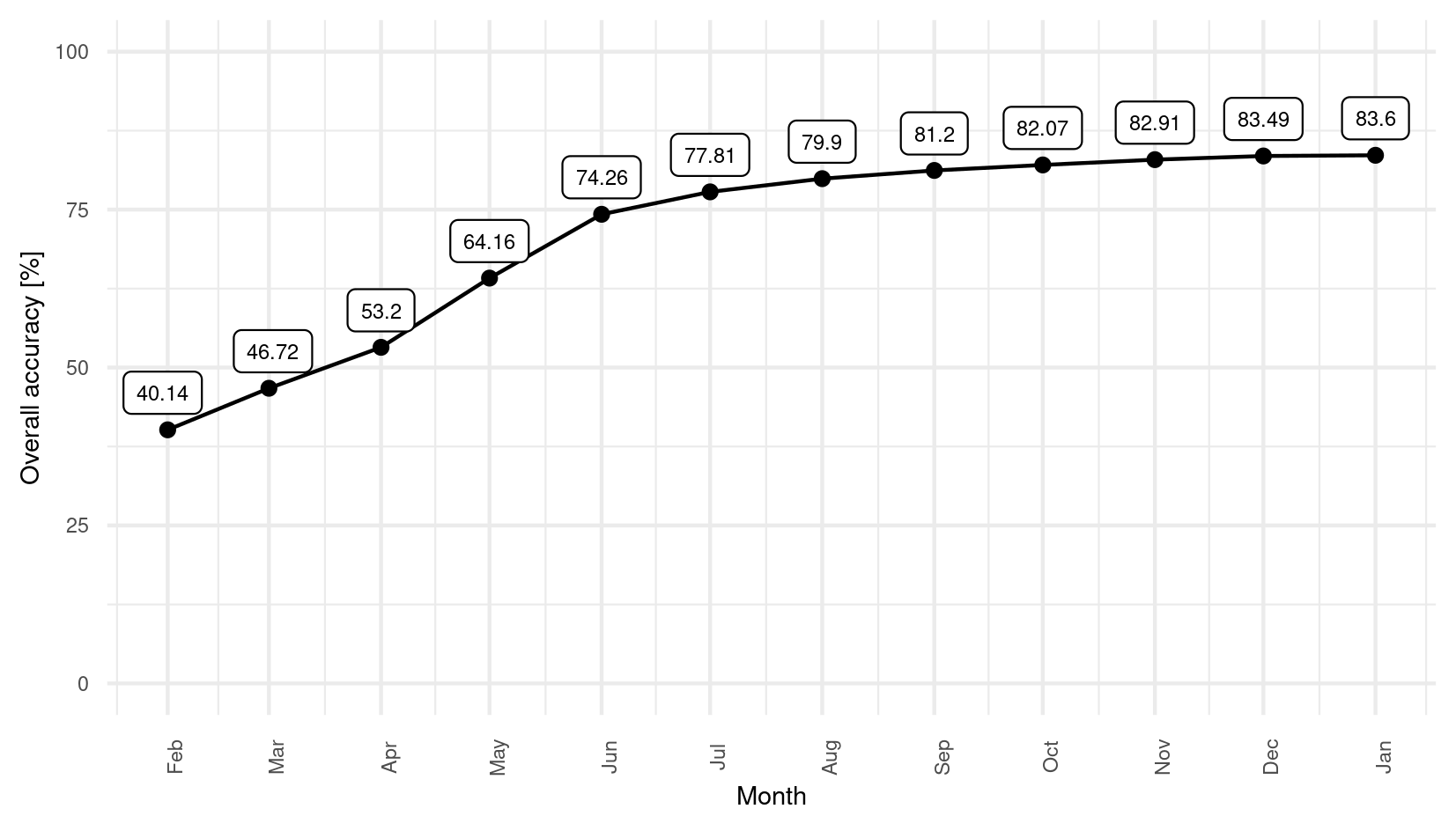}
    \caption{Overall accuracy at a monthly time step for the crop type level and the two strata combined }
    \label{fig:benchmark_date_combined_oa_crop}
\end{figure}

\ref{fig:benchmark_date_combined_oa_crop} summarizes the overall accuracy through time. An initial steep increase in accuracy is observed from February to June when only very small increases are achieved by adding more data. By the end of July the accuracy reaches a value of 78.8\% (\ref{fig:benchmark_date_combined_oa_crop}). 

\begin{figure}[ht]
        \centering
    \includegraphics[width=0.45\textwidth]{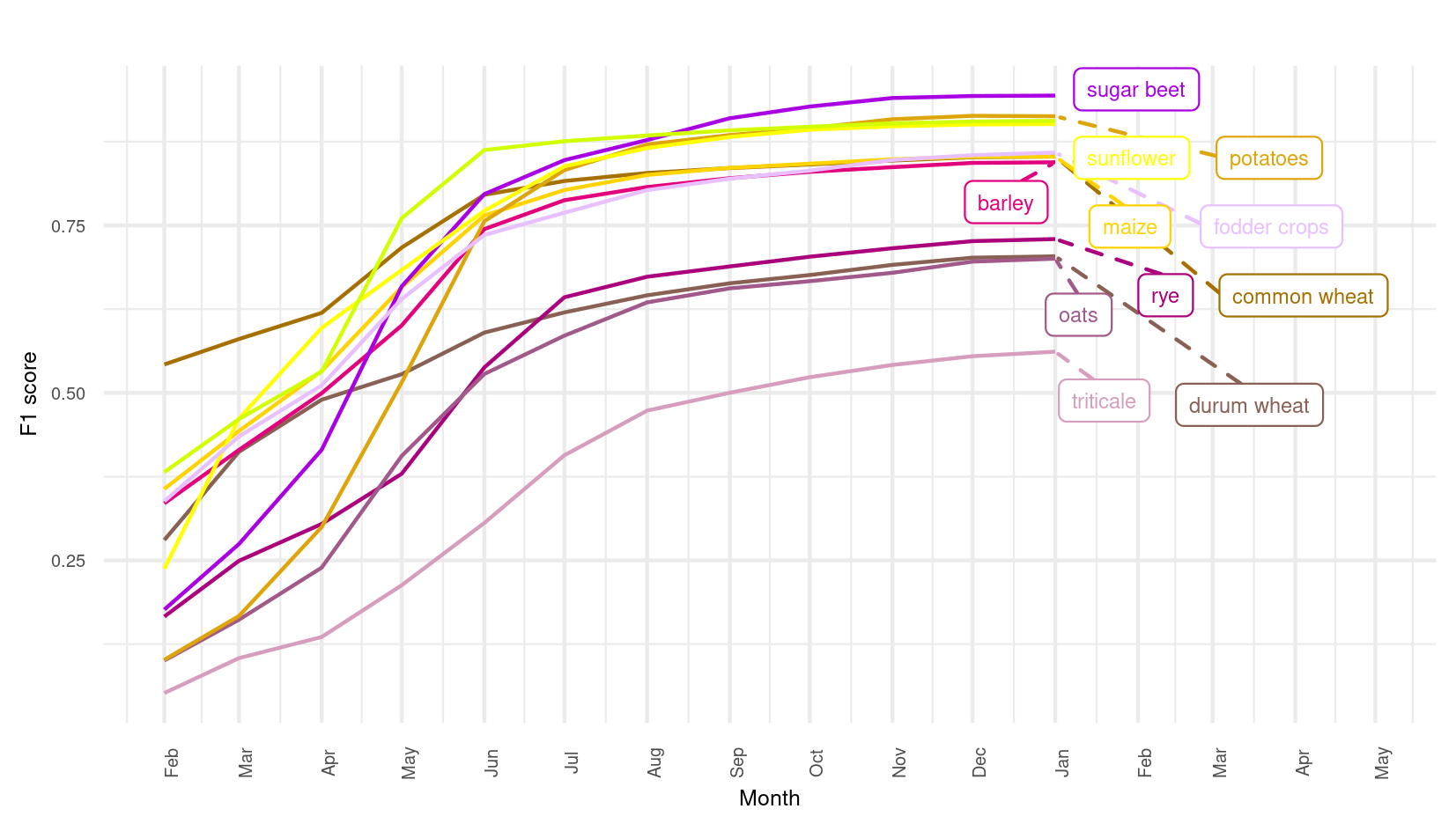}
    \caption{Evolution of The F-score through time for the 12 mains crops over the EU-28. }
    \label{fig:benchmark_date_combined_Fscore_crop}
\end{figure}

The dynamic of the Fscore through time for each crop separately (\ref{fig:benchmark_date_combined_Fscore_crop}) shows expected patterns. The rape and turnip rape Fscore is booming when the crop is flowering in May (Fscore 0.86) indicating that the model could identify them quite early in the growing season (March). The figure also shows that common wheat  has a steep increase in Fscore in April which coincides with the reproductive growth period. Grain maize, a summer crop which is harvested later in the season than winter wheat and rape and turnip rape, is reaching the plateau in Fscore at the beginning of July. Interestingly, the Fscore for sugar beet continuous to increase until October which is logical as sugar beet is harvested in late autumn in Europe. Finally, as expected, we observe a very low Fscore value for mixed classes (such as other fodder crops).

As detailed earlier, and considering also any future operational application, we opted for using time-series from January until the end of July for further processing (evaluation of indices and final training of the RF algorithms).

From the comparison (\ref{tab:benchmarkingIndices}) of the overall accuracy for the polarization backscattering coefficients (VV and VH), the cross-ratio index (VH/VV) along with their combinations, we observe that the combination of VV and VH gives the highest accuracy (79.89 \%) and is selected for the EU crop map. 
The training data for the EU crop map therefore consists of 10-day time series of VV and VH backscatter coefficients from 1st of January to 31st of July (22 dekads). 

\begin{table}[ht]
\centering
\caption{Overall accuracy for S1 indices (dates are fixed from Jan to end of July).} 
\label{tab:benchmarkingIndices}
\begin{tabular}{lr}
  \hline
 Indice & Overal Accuracy \\ 
  \hline
  VV & 71.96 \\ 
  VH & 72.70 \\ 
  VH/VV & 50.90 \\ 
  VV and VH & 79.89 \\ 
  VV and VH and VH/VV & 77.91 \\ 
   \hline
\end{tabular}
\end{table}

\subsection{Crop type over EU}

\subsubsection{Visual assessment}
\begin{figure*}[htb]
  \centering 
         \includegraphics[width=0.8\textwidth]{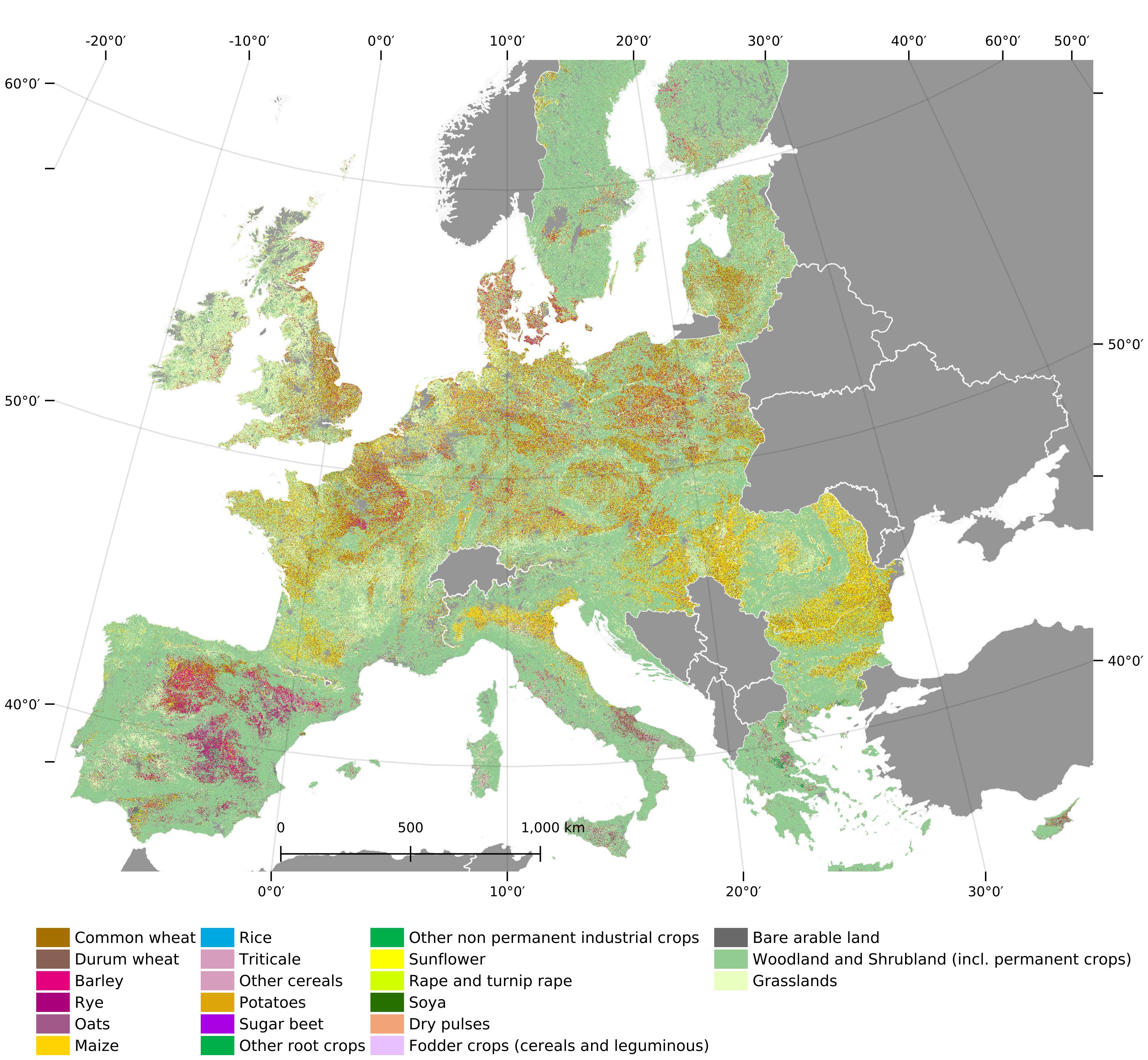}
         \caption{EU crop type map for 2018.}
         \label{fig:EUcroptypemap}
 \end{figure*}

The EU crop map presented in \ref{fig:EUcroptypemap} is representative of a number of crop types and main vegetation classes and 
covers 91 Mha of cropland (9,174 M 10 m pixels). 
The map is available for download and visualization (see section \ref{sec:app_data_dissemination}).

The resulting map is consistent over different landscapes at the level of the main land cover classes. A naive area estimation by pixel counting for each class is provided in  \ref{tab:Appendix_Area}. Woodlands and shrublands are regrouped in a single class covering more than 2 M km\(^2\); i.e. more than half of the area mapped because it is a very broad class covering broadleaved and needleleaved natural and managed forests as well as permanent crops. The arable land class has an extend of about 0.9 million km\(^2\). The grassland class, grouping the permanent and temporary grasslands covers 0.7 million km\(^2\).  The  woodland and shrubland class covers forests patches but also tree lines and hedges in agricultural landscapes.

\begin{figure*}[htb]
  \centering 
         \includegraphics[width=0.8\textwidth]{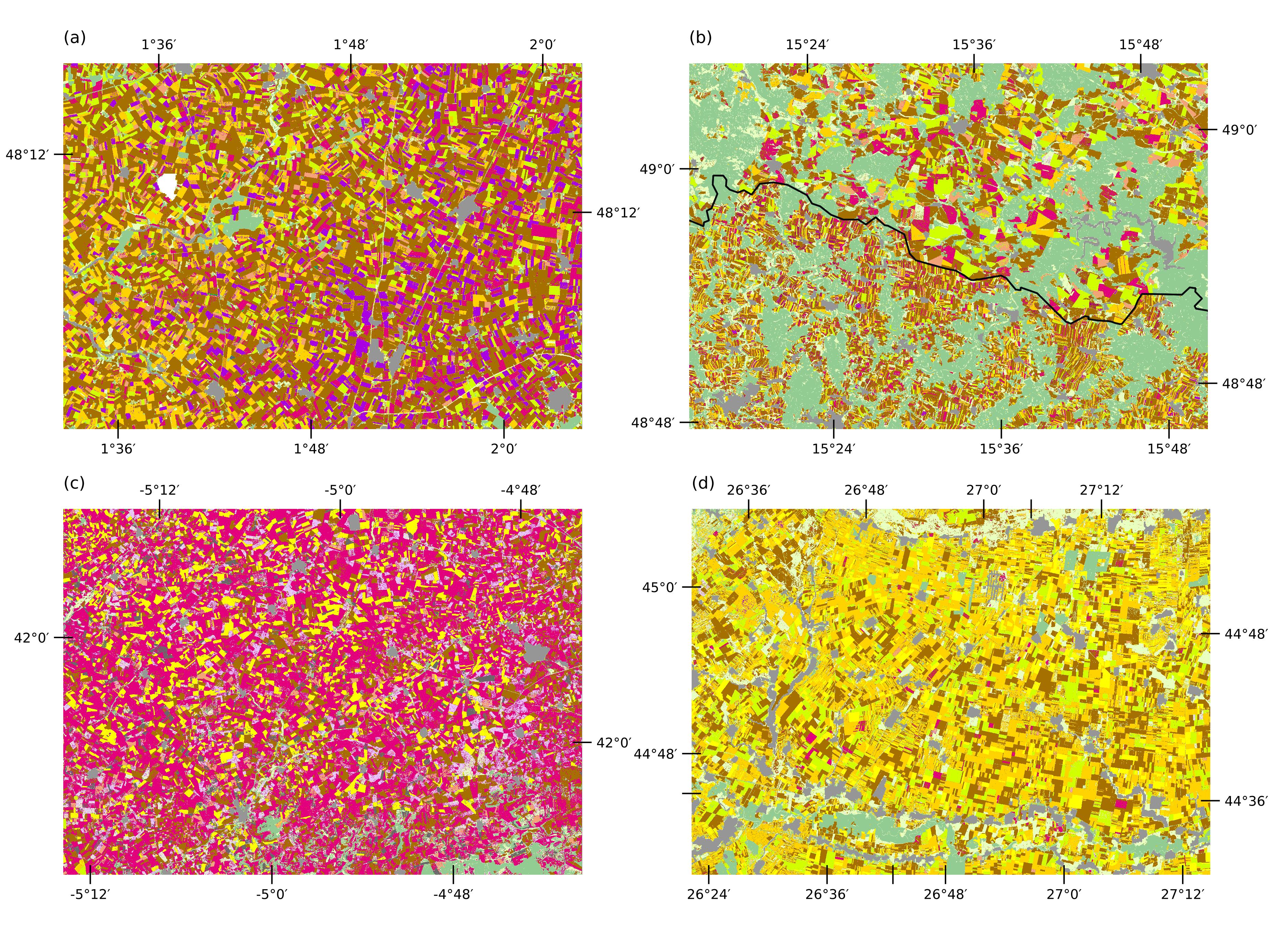}
         \caption{EU crop map for (a) Orleans (France), (b) border between Austria and Czech Republic, (c) Castilla and Leon (Spain), (d) (Romania).}
         \label{fig:zooms}
 \end{figure*}

Over the EU-28, the three most detected crop types in term of area are common wheat, maize and barley. Despite using a pixel based classification approach, separated parcels can clearly be recognized. The difference in the size of parcels is well captured, as illustrated in \ref{fig:zooms}, by comparing  \ref{fig:zooms} (b) Austria and the Czech Republic \ref{fig:zooms} (c). The crop types are identified over the EU without a priori knowledge of country practices. The ability of the map to represent different agricultural landscapes is well illustrated in the examples of \ref{fig:zooms} for North of Orleans, France (a) where the diversity of crop types cultivated in the area are distinctly classified (wheat, barley, sugar beet, maize, potatoes, rape and turnip rape, dry pulses). North West of Valladolid, Spain \ref{fig:zooms}(c) the major crops cultivated in Castilla y Leon: barley and sunflower are clearly detected together with some parcels of wheat, dry pulses and bare arable land. In Romania, North East of Bucharest \ref{fig:zooms} (d), the diversity of small and large parcels of wheat, maize, rape and turnip rape, sunflower and barley are well captured.

\subsubsection{F-score per crop type and administrative unit}

\begin{figure*}[htb]
  \centering 
         \includegraphics[width=0.8\textwidth]{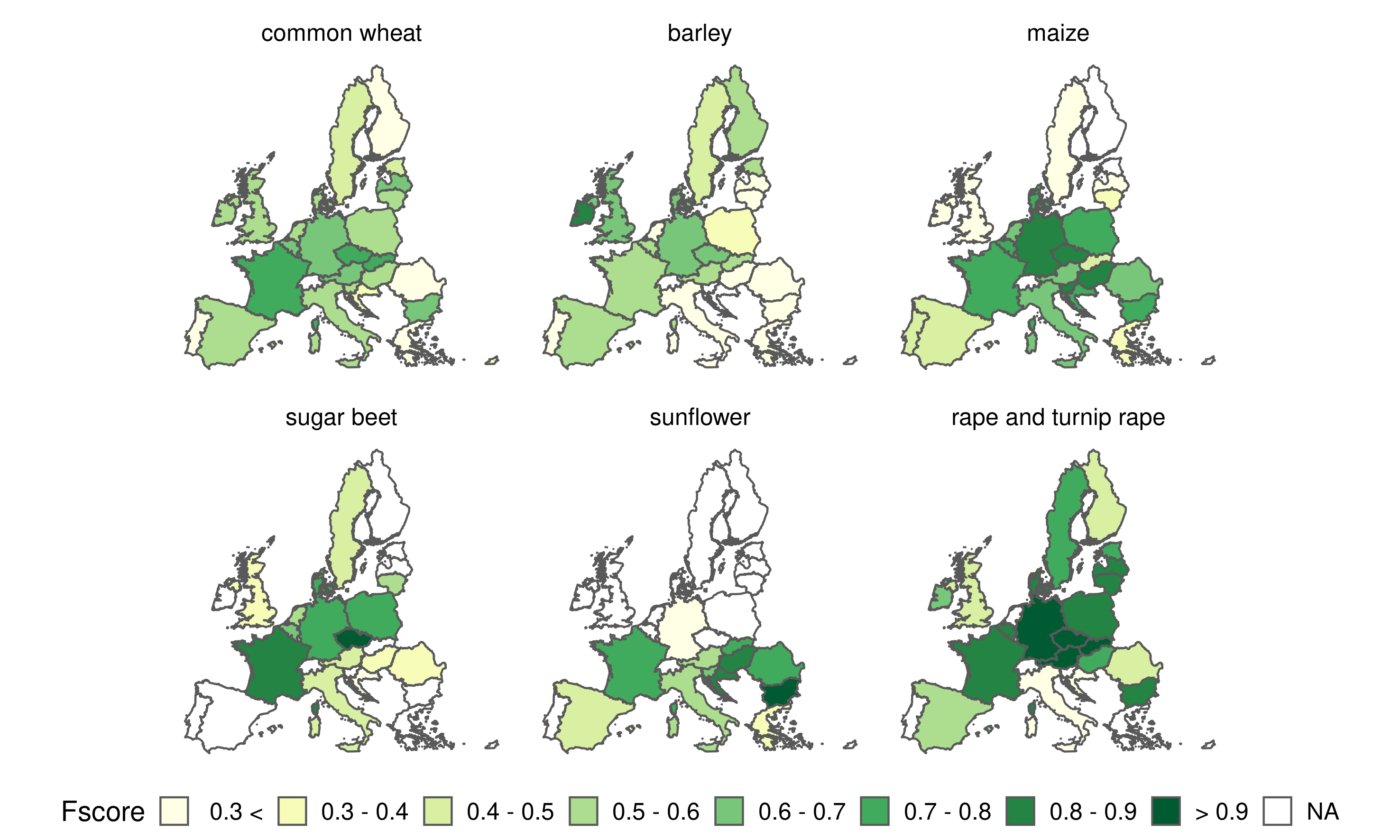}
         \caption{The F-score per country for the 6 main crops in Europe.}
         \label{fig:F1scorePerCountry6classMapped}
 \end{figure*}

As agricultural practices vary across Europe, in terms of crop type cultivated and crop calendar, we expect our mapping approach using a minimal stratification to represent the field reality with different fidelity in different areas. To investigate this aspect, a first assessment of spatial discrepancy for arable land class accuracy is done by calculating the F-score at country level, when enough LUCAS points were available, for the six main crops over EU, i.e. common wheat, barley, maize, sugar beet, sunflower, rape and turnip rape. This allows to quickly grasp how well the classification is performing spatially as highlighted in \ref{fig:F1scorePerCountry6classMapped} (see \ref{tab:fscore_country_6maincrops} for number details). The F-score for most crops is higher in the northern part of the study area than in the southern part. The F-score for common wheat is about 0.7 for Belgium, Bulgaria, Czech republic, Denmark, France, Luxembourg and Slovakia. Maize reaches high scores in the North (above 0.8 for Germany, Czech Republic, Hungary and Slovenia) and rape and turnip rape have an F-score higher than 0.9 in Austria, Czech Republic, Germany, Lithuania, Luxembourg, Latvia, Poland and Slovakia.


\subsection{Accuracy Assessment}
\label{sec:results_accuracy}

\subsubsection{LUCAS core points} 
\label{sec:results_accuracy_lucas}

The accuracy of the EU crop map is evaluated using \num{87,853} points selected from the LUCAS 2018 survey. An overall accuracy of 76.1\% (\ref{tab:LUCAS Confusion matrix}) is achieved over the EU when considering the 19 crop type classes with the woodland and shrubland and grassland classes. An accuracy of 80.3\% (\ref{tab:LUCAS Confusion matrix group}) is achieved when grouping the crops by the main crop type classes (groups defined in the \ref{tab:legend}). The woodland and shrubland class is well discriminated from the rest (PA of 97\% and UA of 83\%). A confusion is observed between the grasslands class and the woodland and shrubland class, as highlighted by a PA of 58\% for grasslands indicating an omission error. The UA of grassland is 82\% indicating a commission error for grassland and again a confusion between the grassland and woodland and shrubland class. The pixels mapped as woodland and shrubland instead of grassland and vice versa could be linked with the class definition of woodland and shrubland including open tree/shrub cover canopy (10\%). In terms of land cover some classes of grasslands are very similar to open canopy class of woodland and shrubland. The point information provided by the LUCAS core dataset is also not the most suitable to assess more complex land cover classes characterised by discontinuous canopies. 

Regarding arable land, cereals are well discriminated, reaching a producer accuracy (PA) of 84\% and user accuracy (UA) of 72\%. Omission errors for the cereals class are mainly occurring with the woodland and grassland classes while commission errors are happening mainly with grasslands. Root crops have a lower PA of 59\%, with many points labelled as root crops being mapped as cereals instead. However the root crops present a high UA of 90\% showing little commission errors with the other classes. The non permanent industrial crops have high PA (89\%) and UA (79\%) with good discrimination of sunflower and rape and turnip rape. The fodder crops class including cereals and leguminous cultivated for fodder is not well discriminated and is confused with the cereals and grassland class (PA of 7\% and UA of 29\%).  

Regarding specific crop type (\ref{tab:LUCAS PA UA main crops}), the LUCAS accuracy assessment highlights a difference between the northern and the southern stratum with an OA of respectively 78\% and 70.8\%. The overall best-performing crop classes are rape and turnip rape, sunflower and maize. In the northern stratum, common wheat, maize, sunflower and rape and turnip rape have PA above 80\% but the value of their UA is lower, indicating commission errors. The commission error is specifically high for the common wheat.


\begin{table*}[htb]
\small
\addtolength{\tabcolsep}{-1pt}
\caption{Confusion matrix for crop type group class for the EU-28. The values represent weighted area UA  Users accuracy PA Producer Accuracy.}
\label{tab:LUCAS Confusion matrix group}
\resizebox{18cm}{!}{
\begin{tabular}{lllllllllllllllllllllllll}
 & \multicolumn{2}{l}{} & \multicolumn{22}{l}{\textbf{Reference class (LUCAS point)}} \\ \hline
 & \multicolumn{2}{l}{\textbf{Map Class}} & \multicolumn{2}{l}{210} & \multicolumn{2}{l}{220} & \multicolumn{2}{l}{230} & \multicolumn{2}{l}{240} & \multicolumn{2}{l}{250} & \multicolumn{2}{l}{290} & \multicolumn{2}{l}{300} & \multicolumn{2}{l}{500} & \multicolumn{2}{l}{UA (\%)} & \multicolumn{2}{l}{SE (\%)} & \multicolumn{2}{l}{OA(\%)} \\ \hline
210 & \multicolumn{2}{l}{Cereals} & \multicolumn{2}{l}{0.13} & \multicolumn{2}{l}{0.003} & \multicolumn{2}{l}{0.002} & \multicolumn{2}{l}{0.002} & \multicolumn{2}{l}{0.005} & \multicolumn{2}{l}{0.009} & \multicolumn{2}{l}{0.002} & \multicolumn{2}{l}{0.026} & \multicolumn{2}{l}{72.3} & \multicolumn{2}{l}{0.6} & \multicolumn{2}{l}{80.3} \\
220 & \multicolumn{2}{l}{Root Crops} & \multicolumn{2}{l}{0} & \multicolumn{2}{l}{0.006} & \multicolumn{2}{l}{0} & \multicolumn{2}{l}{0} & \multicolumn{2}{l}{0} & \multicolumn{2}{l}{0} & \multicolumn{2}{l}{0} & \multicolumn{2}{l}{0} & \multicolumn{2}{l}{90.1} & \multicolumn{2}{l}{2} & \multicolumn{2}{l}{} \\
230 & \multicolumn{2}{l}{Non permanent industrial crops} & \multicolumn{2}{l}{0.001} & \multicolumn{2}{l}{0} & \multicolumn{2}{l}{0.024} & \multicolumn{2}{l}{0.002} & \multicolumn{2}{l}{0} & \multicolumn{2}{l}{0.001} & \multicolumn{2}{l}{0} & \multicolumn{2}{l}{0.001} & \multicolumn{2}{l}{79.5} & \multicolumn{2}{l}{6.2} & \multicolumn{2}{l}{} \\
240 & \multicolumn{2}{l}{Dry pulses, Vegetables and Flowers} & \multicolumn{2}{l}{0} & \multicolumn{2}{l}{0} & \multicolumn{2}{l}{0} & \multicolumn{2}{l}{0.002} & \multicolumn{2}{l}{0} & \multicolumn{2}{l}{0} & \multicolumn{2}{l}{0} & \multicolumn{2}{l}{0.001} & \multicolumn{2}{l}{43.6} & \multicolumn{2}{l}{4.4} & \multicolumn{2}{l}{} \\
250 & \multicolumn{2}{l}{Fodder Crops} & \multicolumn{2}{l}{0.001} & \multicolumn{2}{l}{0} & \multicolumn{2}{l}{0} & \multicolumn{2}{l}{0} & \multicolumn{2}{l}{0.001} & \multicolumn{2}{l}{0} & \multicolumn{2}{l}{0} & \multicolumn{2}{l}{0.002} & \multicolumn{2}{l}{28.8} & \multicolumn{2}{l}{3.3} & \multicolumn{2}{l}{} \\
290 & \multicolumn{2}{l}{Bare Arable Land} & \multicolumn{2}{l}{0.001} & \multicolumn{2}{l}{0} & \multicolumn{2}{l}{0} & \multicolumn{2}{l}{0} & \multicolumn{2}{l}{0} & \multicolumn{2}{l}{0.006} & \multicolumn{2}{l}{0.001} & \multicolumn{2}{l}{0.003} & \multicolumn{2}{l}{50.3} & \multicolumn{2}{l}{2.7} & \multicolumn{2}{l}{} \\
300 & \multicolumn{2}{l}{Tree and Shrub Cover} & \multicolumn{2}{l}{0.01} & \multicolumn{2}{l}{0} & \multicolumn{2}{l}{0} & \multicolumn{2}{l}{0.001} & \multicolumn{2}{l}{0.004} & \multicolumn{2}{l}{0.004} & \multicolumn{2}{l}{0.483} & \multicolumn{2}{l}{0.077} & \multicolumn{2}{l}{83.4} & \multicolumn{2}{l}{0.4} & \multicolumn{2}{l}{} \\
500 & \multicolumn{2}{l}{Grassland} & \multicolumn{2}{l}{0.01} & \multicolumn{2}{l}{0} & \multicolumn{2}{l}{0} & \multicolumn{2}{l}{0} & \multicolumn{2}{l}{0.009} & \multicolumn{2}{l}{0.001} & \multicolumn{2}{l}{0.013} & \multicolumn{2}{l}{0.151} & \multicolumn{2}{l}{82.1} & \multicolumn{2}{l}{0.5} & \multicolumn{2}{l}{} \\ \hline
 & \multicolumn{2}{l}{Producer accuracy (\%)} & \multicolumn{2}{l}{84.4} & \multicolumn{2}{l}{58.6} & \multicolumn{2}{l}{89.3} & \multicolumn{2}{l}{23.6} & \multicolumn{2}{l}{6.9} & \multicolumn{2}{l}{25.7} & \multicolumn{2}{l}{96.9} & \multicolumn{2}{l}{58.1} & \multicolumn{2}{l}{} & \multicolumn{2}{l}{} & \multicolumn{2}{l}{} \\
 & \multicolumn{2}{l}{Standard Error (\%)} & \multicolumn{2}{l}{0.8} & \multicolumn{2}{l}{3} & \multicolumn{2}{l}{1.3} & \multicolumn{2}{l}{3.9} & \multicolumn{2}{l}{0.8} & \multicolumn{2}{l}{1.7} & \multicolumn{2}{l}{0.1} & \multicolumn{2}{l}{0.6} & \multicolumn{2}{l}{} & \multicolumn{2}{l}{} & \multicolumn{2}{l}{} \\ \hline
\end{tabular}}
\end{table*}

\begin{table}[htb]
\centering
\footnotesize
\caption{Weighted area accuracy metrics, overall accuracy for the crop types classification (level-2) and producer and user accuracies for the main crop type classes (for the whole study area and for the two strata)}
\label{tab:LUCAS PA UA main crops}
\begin{tabular}{lrrrrrr}
\hline
 & \multicolumn{2}{l}{All} & \multicolumn{2}{l}{stratum 1} & \multicolumn{2}{l}{stratum 2} \\ \hline
class & P.A. & U.A & P.A. & U.A & P.A. & U.A \\ \hline
common wheat & 78.2 & 49.6 & 82.5 & 49.9 & 44.5 & 46.0 \\
durum wheat & 25.9 & 49.7 & 1.8 & 92.3 & 45.2 & 49.0 \\
barley & 53.8 & 51.6 & 50.0 & 57.8 & 67.2 & 44.2 \\
maize & 84.0 & 58.5 & 87.6 & 58.3 & 22.7 & 66.7 \\
potatoes & 37.3 & 73.5 & 39.8 & 74.1 & \textbackslash{} & \textbackslash{} \\
sugar beet & 57.1 & 75.0 & 57.9 & 75.0 & \textbackslash{} & \textbackslash{} \\
sunflower & 86.7 & 62.2 & 90.7 & 72.0 & 64.3 & 32.4 \\
rape and turnip rape & 82.7 & 80.2 & 83.5 & 80.2 & 17.9 & 78.9 \\ \hline
Overall   accuracy & \multicolumn{2}{l}{76.1} & \multicolumn{2}{l}{78.0} & \multicolumn{2}{l}{70.8} \\ \hline
\end{tabular}
\end{table}

\subsubsection{GSAA}
\label{sec:results_accuracy_gsaa}
\begin{figure}[htb]
  \centering
  \begin{subfigure}{.9\linewidth}
    \vspace{0.1cm}
    \centering
    \includegraphics[width = \linewidth]{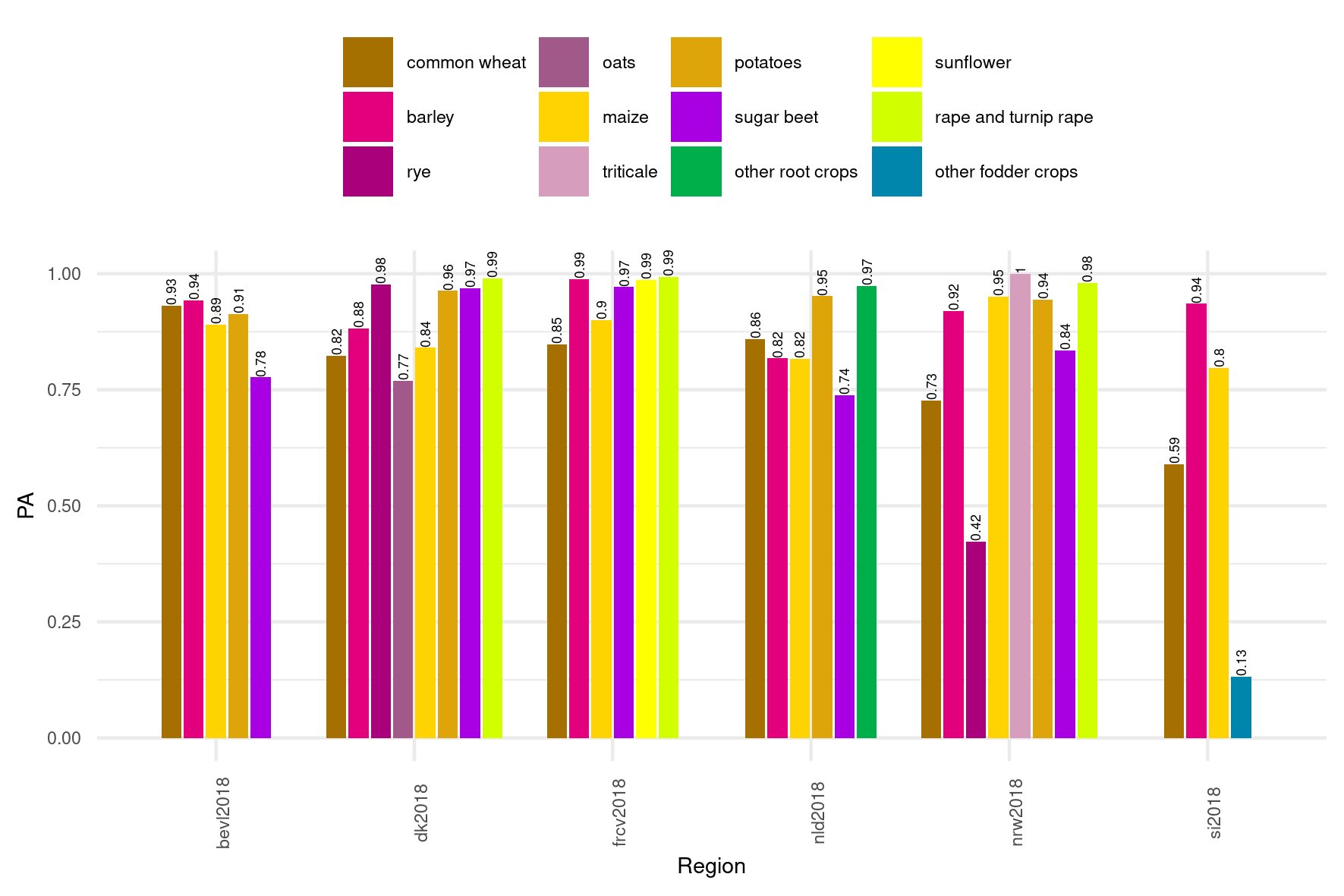}
    \caption{PA}
    \label{fig:LPIS_PA}
  \end{subfigure}

  \begin{subfigure}{.9\linewidth}
    \centering
    \includegraphics[width = \linewidth]{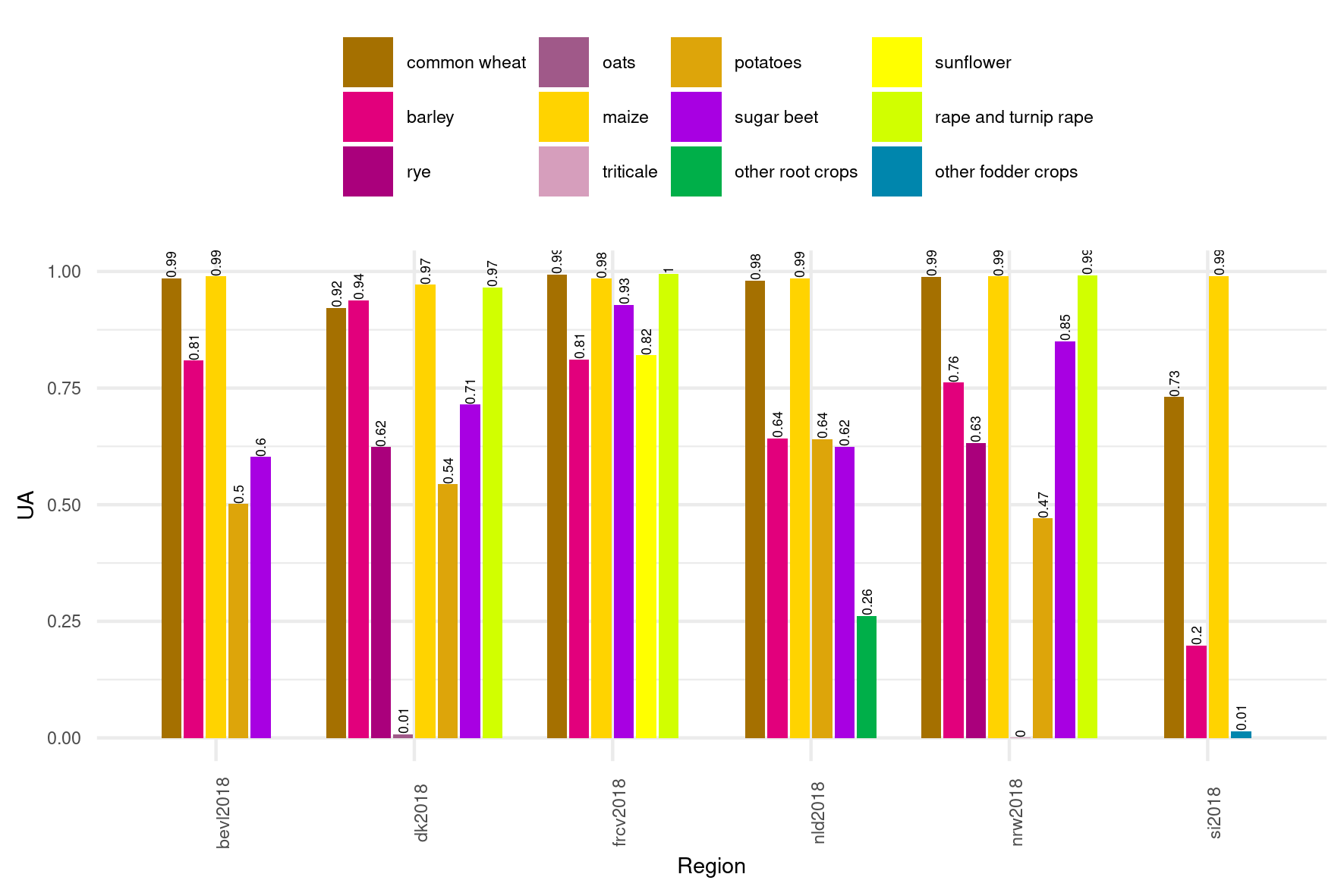}
    \caption{UA}
    \label{fig:LPIS_UA}
  \end{subfigure}
  \caption{User and Producer accuracy obtained from the comparison of the EU crop map and the parcel data from the GSAA. The detailed confusion matrices are in \ref{AppendixA}.}
\end{figure}

The LUCAS core points are giving a first evaluation of the accuracy of the crop type at the continental level. A more detail assessment is done with the GSAA data for specific countries or regions. For each region, the classified crop of each parcel from a crop representing an area greater than 1\% of the region is compared to the one from the GSAA. After removing the grassland and woodland classes the confusion matrices are computed. The confusion matrices along with the PA and UA for each region are presented in \ref{AppendixA} see (\ref{tab:bevl2018} for bevl2018, \ref{tab:dk2018} for dk2018, \ref{tab:nld2018} for nld2018, \ref{tab:nrw2018} and \ref{tab:si2018} for si2018). The PA and UA for the main crops are summarized in \ref{fig:LPIS_PA} and \ref{fig:LPIS_UA} respectively.

The analysis shows very promising results for the detection of common wheat, barley, maize, sunflower and rape and turnip rape in all the regions analysed. 
Rape and turnip rape are detected without almost any confusion in the three regions where it is present (dk2018, frcv2018 and nrw2018), reaching PA above 98\% and UA above 97\%. 
It is worth noting that common wheat and barley have much better accuracies for both PA and UA compared to the LUCAS points based assessment, with the exception of si2018. Except for nrw2018 and si2018, common wheat has a PA above 80\%, while the UAs are above 92\% in all regions except si2018. In nrw2018, the most common confusion for common wheat are omission error towards  barley and triticale classes. On the other hand any parcels of triticale in GSAA are correctly mapped as triticale. The confusion of common wheat with triticale is due to the similarity of the two crops, both in general crop structure and seasonality. 
In si2018, omission errors are observed towards barley and the other cereals class.
Barley is present in the six regions with a PA above 92\% except for dk2018 (88\%) and nld2018 (82\%). In Denmark, this is explained by a confusion between common wheat and barley while in nld2018 the confusion is mainly with maize. The UA for barley are a bit lower than the PA value but still above 80\% for bevl2018, dk2018 and frcv2018. Both in nrw2018 (UA 76\%) and nld2018 (UA 64\%), the main commission error is due to a confusion with common wheat.
Sugar beet presents very promising results in bevl2018 (PA 97\% and UA 93\%) and nrw2018 (PA 84\% and UA 85\%). In bevl2018, dk2018 and nld2018 where UA are lower, the confusion is mainly with maize and to a lesser extent with potatoes.
The class potatoes, presents in four regions, shows high PA (91\% in bevl2018, 96\% in dk2018, 95\% in nld2018 and 94\% in nrw2018) and relatively low UA. In nrw2018, dk2018 and bevl2018, the UA (respectively 47\%, 54\% and 50\%) of potatoes is explained by commission errors with maize. However, the UA of maize is showing little commission errors (99\%). In nld2018, confusion is happening with maize and sugar beet and probably linked to the time period made available to the classifier. Maize and sugar beet maturity phases are reached in autumn, between September and November. A better discrimination would thus be expected later in the season.

\subsubsection{Subnational  statistics}

\label{sec:results_accuracy_eurostat}

\begin{figure*}[htb]
  \centering 
         \includegraphics[width=0.8\textwidth]{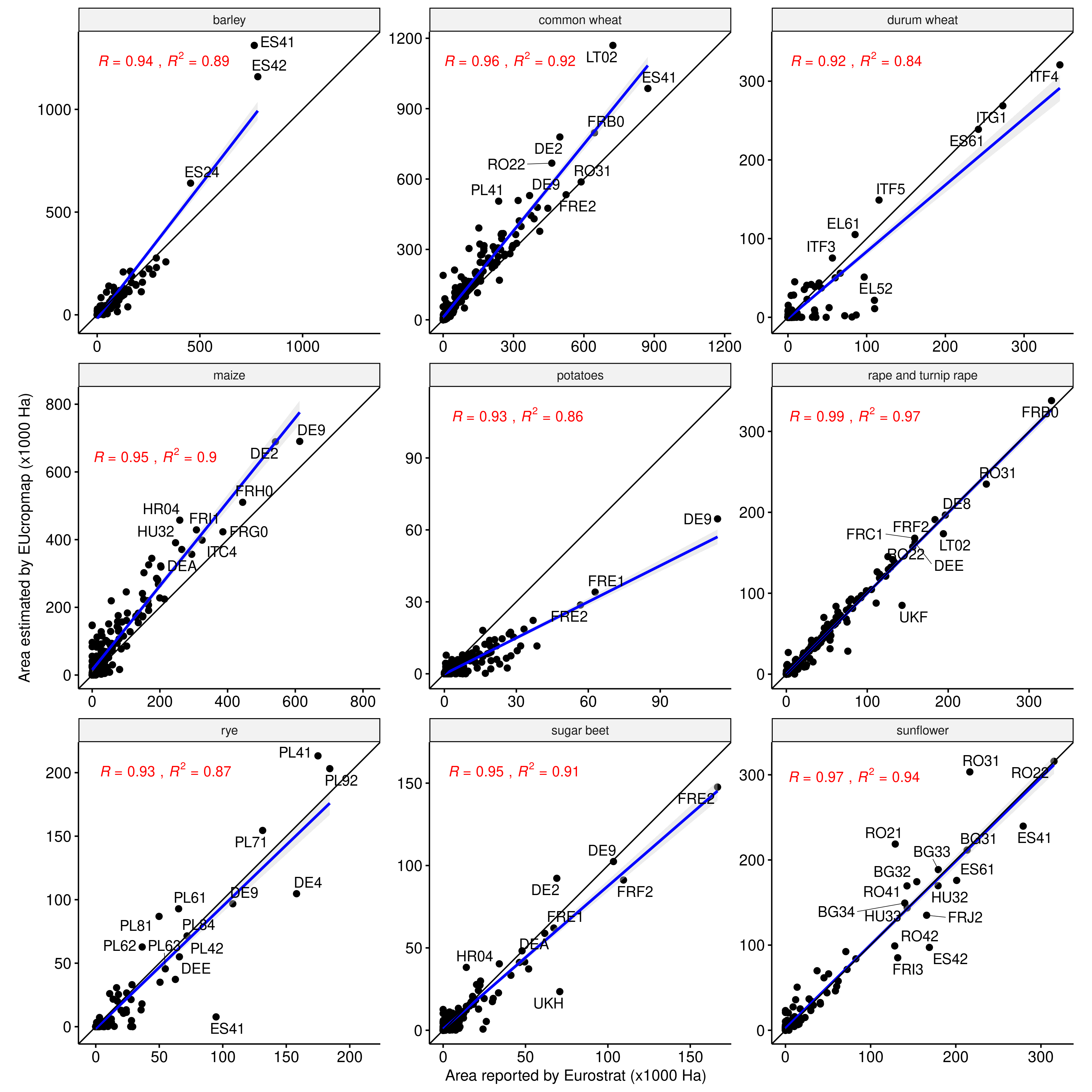}
         \caption{The areas reported by Eurostat at NUTS2 level (except for UK and DE at NUTS1) are compared with the area retrieved from the EU crop map. $R$ is the Pearson correlation coefficient, $R^2$ is the  the coefficient of determination.}
         \label{fig:Nuts2Comparison}
 \end{figure*}

The crop area data is extracted and compared for 220 administrative regions covering most of the EU-28. The matching with the reported statistics was done at the NUTS2 level(\ref{tab:legend_convergence_Estat}). For DE and UK no data is available at NUTS2 level, the comparison was thus done at NUTS1 level for which statistics are available. \ref{fig:Nuts2Comparison} shows how the area estimated by the EU crop map compared with the area reported by Eurostat. The calculated Pearson's correlations (r) range from 0.93 for potatoes and rye to 0.99 for rape and turnip rape.
Additionally, the relative difference in area (\%) is computed for each region and presented in \ref{fig:Nuts2Comparison_map} (\ref{fig:Nuts2Comparison_hist} provides the histogram of the differences for each crop). The maps show that while for common wheat and maize the EU crop map estimations are generally higher compared to Eurostat without a distinct spatial pattern in observed differences, this is not the case for other crops, whose overestimation and underestimation are more localised. Barley for example is underestimated in the center of Europe but overestimated in both the north and the south of Europe.
The distribution of the difference shows that common wheat and maize are underestimated in our EU crop map compared to Eurostat. Barley is under-estimated in the northern part and over-estimated in the southern part. The other main crop area estimations are close to the official Eurostat statistics.

\section{Discussion}

\subsection{Timeliness and accuracy of the crop type classification}

\begin{figure*}[htb]
  \centering 
         \includegraphics[width=0.8\textwidth]{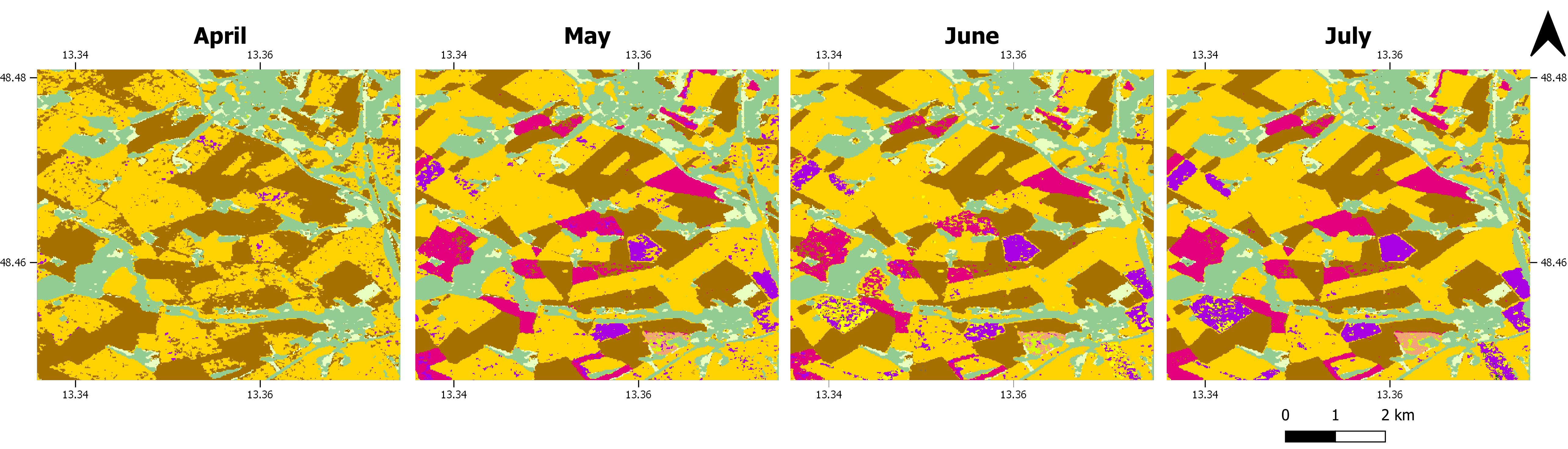}
         \caption{Crop type mapping results for different time periods, considering the data from January to respectively April, May, June, and July (the EU crop map presented in this study) in South East Germany.}
         \label{fig:EUcropmapTime}
 \end{figure*}

The final map presented in this study is built with S1 observations from January to end of July. The overall accuracy at the end of July across all crops is relatively high at 76\%. As the season progresses, discrimination of crop types and parcels keeps improving as shown in \ref{fig:EUcropmapTime}. However, for an operational application specific crops can be mapped earlier during the season depending on the specificity of the region and requirements regarding accuracy for specific crop types. The timing when the F-score is highest for each specific crop is linked to canopy structure and phenological stage as these affect the backscattering signal. For in-season applications, a trade-off can be found between the required level of accuracy and timely map production. 

While covering only 2018, the current study shows potential for developing in season prediction of crop type and confirms the observations of  \citep{veloso2017understanding,you2020examining} at the EU scale. A key element is that the S1 SAR signal is not affected by atmospheric perturbation. Any acquisitions can thus be used for crop type mapping. Although dependent on year to year phenological development, findings regarding the best timing for specific crop types are likely to be valid for future studies. 

\ref{fig:S1TScropGroupStratumComparison} shows that the time series of VV and VH backscatter for specific crop types have a similar shape in the northern and the southern stratum but often appear shifted in time. This relates to phenological development that occurs later when going northward. As a consequence, the F-score for the same class may peak at different times as revealed by the analysis of the temporal evolution of the F-score in the two strata. 


In the northern stratum (\ref{fig:benchmark_date_combined_Fscore_crop_strata}), the F-score reaches a plateau at the end of June for common wheat, rape and turnip rape. Other cereals (rye, barley and maize) can be predicted with a high F-score at the end of July, even if the F-score keeps increasing slightly until the end of the summer. The F-score for sunflower, potatoes and other fodder crops reaches a  plateau one month later, at the end of August, while sugar beet reaches the plateau in September, in agreement with its longer growth cycle that extends into autumn. Winter wheat and rape and turnip rape that are already established in the field at the beginning of the year, present a structure, reflected by the VV and VH backscatter, that is already different for the other classes earlier in the year.

Similar results were found in other studies exploring the potential of S1 for crop type monitoring in northern European countries. For Belgium, \cite{van2018synergistic} illustrates that while winter cereals were already well discriminated at the end of June, end of August was more appropriate for potatoes, maize or sugar beets. For France, \cite{veloso2017understanding} showed that  S1 backscatter clearly separates winter wheat and rapeseed between March and July during the tillering and senescence stages and distinguishes maize, soybean and sunflower during the heading/flowering phase between June and the end of August.


In contrast, the highest accuracies are achieved earlier in the Mediterranean stratum. Already at the end of June, plateaus in F-score are reached for the accuracies of wheat (common and durum wheat), rape and turnip rape, other fodder crops, and the mixed class "other non permanent industrial crops" (including e.g. cotton, fibre and oleaginous crops). 
At the end of July, the plateau is reached for barley, sunflower and root crops while it is reached end of August for maize.
The earlier time at which the highest accuracies are achieved in the southern stratum can be explained by typical timings of the growing season in the two strata. As shown in (Figure S4) \cite{meroni2021comparing} there is latitudinal gradient in the timing of the growing season with earlier  growing seasons in the south of Europe. This gradient justifies the observed lag in achieving the best performances in the two strata

To assess the accuracy of the EU crop type map produced at the end of July, three different and complementary approaches were presented in the Section \ref{sec:results_accuracy}. We discuss here the pros and cons of each.

The first one using high-quality LUCAS points (Section \ref{sec:results_accuracy_lucas}) enables an accuracy assessment for the whole continental area but on the other hand the protocol to collect the in-situ information is not designed for EO applications and we filtered out "low-accuracy" points, working only with a subset of the 2018 survey. In addition, the point information is difficult to reconcile with the assessment of crop type classes that present spatial heterogeneity in a small area and especially where only few acquisitions contribute to the 10-day average (e.g. in the South). Speckle filtering, not used in the present study, may attenuate such undesired heterogeneity.

The information about the crop type is then assessed more precisely with the GSAA (Section \ref{sec:results_accuracy_gsaa}). The main differences between the LUCAS point based and GSAA parcel based accuracies are due to the parcel aggregation of the pixel classification within the GSAA parcel, which uses a majority filter. In this way, the analysis is not deteriorated by possible geometric inaccuracies that can affect the LUCAS points. In addition, speckle-induced variation affecting the pixel classification result within the parcel is removed. However, the GSAA assessment is limited to specific areas in the northern part of the study, due to a lack of openly accessible GSAA sets for 2018. 

Finally, the comparison with the official statistics (Section \ref{sec:results_accuracy_eurostat}
) has the obvious advantage of covering the whole area of interest but the drawback of being a rather coarse evaluation of the classification as the comparison with official statistics can only be performed at some aggregate administrative spatial level. 

Useful observations can be derived from the comparative analysis of of the three complementary level of accuracy assessment. For instance, some specific errors are detected with all approaches.

From the comparison with official statistics (Fig. \ref{fig:Nuts2Comparison_map} and \ref{fig:Nuts2Comparison_hist}) we observe an overestimation of common wheat in the northern part of our study. Similarly, for common wheat, an important commission error with other cereals classes is depicted by the LUCAS points assessments. The same confusion (common wheat versus barley and triticale) is reported by the GSAA assessment even with a smaller severity. 

An over-estimation in the northern stratum is also observed for maize. This is consistent with what is observed in the accuracy assessment based on the LUCAS point(a low UA associated with an important commission error). In contrast, this is not reported in the GSAA assessment where the high UA of maize illustrates a low commission error for that class. Contrastingly, an under-estimation of maize is observed in the southern part of our study area. This is in agreement with the omission errors reported in the LUCAS assessment. We however lack of the GSAA comparison in the South to confirm this observation.

While barley is over-estimated in the southern part of the study, it is under-estimated in the northern part according to the official statistic assessment. The underestimation of barley in the North is in agreement with the LUCAS point assessment reporting low PA and UA. In contrast, the GSAA assessement reports PA are above 80\% in all GSAA analyzed, not confirming an omission error. 
An hypothesis is that the GSAA assessment filters the pixels that are not yet detected as barley at the parcel level. The overestimation of barley in the South is confirmed by the LUCAS point assessment.


The confusion observed at the level of the study area with the LUCAS points assessment (\ref{tab:LUCAS Confusion matrix group}) between the cereals and the root crops and for the GSAA between sugar beet and maize and potatoes is directly related to the time period (January to end of July) selected for EU crop map. The evolution of the VH backscatter coefficient in time \ref{fig:S1TScropGroupStratumComparison} illustrates what is also visible from the F-score \ref{fig:benchmark_date_combined_Fscore_crop_strata}, that the discrimination between broad leaf crops, cereals and maize increases from June onwards in the Northern stratum. 
Extending the period of a few month would alleviate these problems.



\subsection{Recommendation for an operational service}
\label{sec:discussion_recommendation}
More than 6 years after the successful launch of the first Copernicus satellite, S1 A, Europe still lacks access to a full, free and open archive of Level-2 Copernicus Application Ready Data for the S1 (A and B) sensors (S1-CARD). Whereas Sentinel-2 Level-2 data is produced as a routine ESA output, this is not the case for S1. Level 2 S1-CARD generation is a required expensive processing task that should be done once with the results made available to the community.  

The scale of this study requires the use of advanced cloud computing solutions closely coupled to the massive on-line data archives of Sentinel data. The requirements for such cloud solutions are that they provide access to application ready data (ARD) and a programming interface that allows the application of advanced machine learning algorithms, for instance, in support of supervised classification, as was applied in this study. GEE is currently the only platform that provides access to consistently processed S1 ARD (backscattering coefficients only, for now), which is processed with a standard recipes using the open source SNAP s1tbx software\footnote{S1TBX - ESA Sentinel-1 Toolbox - Documentation, \url{http://step.esa.int}.}. 
Nevertheless, although  GEE offers a number of routines for supervised classification (including RF used in this study), it has some limitations in the hyperparameter tuning along with some processing issues when the model is applied to very large areas. This is why the pre-processing of the S1 data into 10-day temporal synthesized mosaics was done in GEE while the results were downloaded on a local facility to perform the classification. For this study, this represented about 10 TB of data. This data transfer is clearly sub-optimal and, in addition, while GEE is free to use for non commercial application, the download of large amount of data has to be done via a paid Google Cloud account. 
In our study, we have processed the annual volume of data for 2018 in a post-hoc fashion (i.e. the classification was done off-line after the growing season). However, with the effective use of cloud based solutions, our method can be run in quasi-real time, for instance as a combination of transfer learning with early season in situ collection and GSAA for training and testing. Cloud based capacities allow integration of daily acquisitions in 10-day compositing with automatic re-run of the ML classifiers. Deep learning approaches that learn to extract discriminative temporal information from the data, such as convolutional Neural Networks and Recurrent Neural Networks could provide a more robust alternative to RF and should be considered for further studies.

Finally, it is interesting to reflect on the actual utility of a EU crop map against other potential sources. We in fact acknowledge that there would be no need for a post-harvest crop map with 70-85\% accuracy for only a limited set of major crops if the annual GSAA declaration set is available in July, with a typical accuracy of 90-95\% (for a far more significant set of crops). However, so far, most of the EU Member States' GSAA are not openly accessible and, when this is the case, they are usually available after the growing season (the Netherlands being an exception, with a preliminary version published in July of the current year). Recently, the Integrated Administration and Control System (IACS) spatial data sharing initiatives\footnote{Art. 65 of the Proposal for a Regulation of the European Parliament and of the Council, COM(2018) 393 final).} is promoting EU member states to share their IACS data within the infrastructure for spatial information in Europe (INSPIRE) portal\footnote{See the discovery portal \url{https://agridata.ec.europa.eu/extensions/iacs/iacs.html} and see \cite{tgiacssdiscovery}.}. The expected outcome of this new regulation is the expanded public availability of  GSAA.

While LUCAS is performed every three years, the GSAA data are collected every year in Europe, making them and ideal source of ground data for crop classification training and validation. Beside the legal and technical challenge to make the data open access already discussed, the legend semantic harmonization is key to use in classification approaches across different regions. While the semantic matching was done for this specific study, a common semantic model would be very useful to integrate GSAA in a more detailed EU-wide classification context. 

Besides LUCAS and GSAA, other information source could be considered. The use of crowd sourcing to collect in-situ data and generate LUCAS-like information could enrich data availability  \citep{laso2020crowdsourcing}. The use of geo-tagged street-view pictures has been already investigated for agriculture monitoring \citep{d2018crowdsourced} and could represent an alternative in-situ dataset. Application like Pl@ntnet \citep{joly2016look} or i-naturalist \citep{van2018inaturalist} collect massive plant and species occurrence information via their portal. Finally, leveraging mobile gaming solutions such as the impact games  (\url{https://world.game/}) could represent an additional alternative to collect massive in-situ data.

\subsection{The EU crop map 2018: perspectives for improvement}


There are a number of limitations in the proposed mapping approach. Some are coming from the lack of information in the LUCAS survey and other from methodological choices.

From the LUCAS data, the limitation comes mainly form the crop type surveyed, the number of samples per classes and details about the agriculture practices such as irrigation. We did not distinguish between rain-fed and irrigated parcels due to a lack of samples in each category. This particularly affects the maize detection in the South of Europe where most of the maize is irrigated. The \ref{fig:Nuts2Comparison_map} shows that the maize is indeed underestimated in Spain, southern Italy and Greece. Another limitation is that the approach proposed doesn't capture rice. While that class is usually easy to discriminate from other crops, it is here missed due to the lacking of training data (i.e. 14 polygons). The west area of northern Italy is particularly affected, with a confusion between rice and maize. Future studies could consider using information about the water management available in the LUCAS survey. Better understanding and taking into account on the backscatter signal of irrigated surfaces would also certainly improve the discrimination of irrigated crops.  

In term of methodological choices, improvements are possible taking into consideration crop classes not considered in this study. Specifically, permanent crops are covering a considerable extent of agricultural land and are currently classified in the broad class Woodland and shrubland. The Copernicus information for the two classes of permanent crops (fruit trees and the other permanent crops such as olive and vineyard) can be used to explore classification workflows identifying them. Distinction between temporary, permanent and natural grasslands would also be worth considering in future developments which would include multi-season temporal profiles.

Although we have not considered using S1 coherence in this study we are well aware of the potential use of this alternative Level 2 product in agricultural use contexts (e.g. \cite{tamm2016relating,de2021grassland}) \footnote{The terrain flattening correction that is recommended for wall-to-wall mapping applications of S1 can be implemented as a per-orbit reference layer, which is due to the excellent orbit stability of the S1 sensors. See \url{https://developers.google.com/earth-engine/tutorials/community/sar-basics}.}. We realize that a future extension of S1 use in national or continental crop mapping requires the introduction of terrain flattening to better compensate for incidence angle and look direction differences in the multi-orbit compositions for 10-day periods. 
Although we have masked terrain with steep slopes ($>$ 10$^{\circ}$) above 1000 m limiting these effects somewhat, we expect improvements especially for crops with an extended bare soil phase (e.g. summer crops) for which angular variability in backscattering can be significant. We have not applied speckle filtering, because speckle variance is already reduced in the averaging step we apply to create the 10-day composites and using polygon averaged values for the RF training. However, the sparser revisit frequency in the Southern areas makes speckle filtering potentially useful. Furthermore, we could consider speckle filtering with a shorter averaging period (e.g. 5 days) in the North, to better capture abrupt temporal changes in the $\sigma^0$ signatures (e.g. at harvest). 

This study demonstrates the potential of S1 as a baseline for continental mapping. However Sentinel-2 (S2) would certainly improve the results as already demonstrated by \cite{van2018synergistic,gao2018crop,kussul2018crop,orynbaikyzy2020crop,chakhar2021improving}. In particular, we expect that the inclusion of S2 data would improve the discrimination of  main cropping practices. For example helping to separate winter crops from permanent grassland, for the detection of bare soil conditions in spring to delineate summer crops, and to highlight crop specific phenology events (e.g. \cite{d2020detecting}).  Additionally, S2 would be particularly useful in the South where there is less cloud coverage and also less S1 acquisitions \citep{lemoine2017GSARS}. More generally, to facilitate S2 ingestion in the machine learning processing, smoothed time series of cloud free biophysical variables would be useful (e.g. the Copernicus pan-European High-Resolution Vegetation Phenology and Productivity products (HR-VPP) \citep{tian2021calibrating}).

The EU crop map provides a coherent and unique continental view with a spatial detail that is relevant for crop management interventions at the parcel level. As such, the information can be relevant for EU wide policy monitoring, e.g. by deriving indicators such as crop diversity \citep{merlos2020scale}, and be a source of input data for EU wide models and assessments requiring detailed information on crop location. 


Another perspective is to derive national or sub-national area statistics from the EU crop map. This was not considered in this study. Area estimation can be done directly from the crop map (pixel counting) but this is not taking into account the bias inherent of remote sensing: the presence of mixed (border) pixels and the misclassification of pure pixels \cite{gallego2004remote}. Estimating area from a satellite-based map typically requires the use of an un-biased estimators \citep{Olofsson2014a,Stehman2014a,gallego2010european} by combining data from a ground survey or from another source of higher quality data in combination with the map. 
LUCAS 2018 survey is a source of reference data, providing information until the NUTS2 level. However,
further exploration is needed in order to use it for area estimation at a finer scale. While the approach of \cite{gallego2008using} for area estimation with CLC as post-stratification could be a first naive approach, a more precise approach would be the one of \cite{Stehman2014a} to take into account the fact that the strata from our map and the strata from the LUCAS survey are not similar. However, for this purpose, the sampling weight of each LUCAS point should be made available by Eurostat\footnote{It is worth mentioning here that two institutional ongoing projects are developing methodologies and good practices to integrate EO in statistic collection process: Sen4Stats (\url{https://www.esa-sen4stat.org/}) and EO4stats  ("Training on EO for statistics and testing of the methodology" PN5C/03/2020/E4).}. In addition, further care should used to account for the fact that a subset of the 2018 survey was used in the training.  

This study sets up a framework that could apply for other years. While the approach demonstrated is robust, using model trained for 2018 on other years could result in not satisfying result. Indeed, it is noted that 2018 was a peculiar year, with droughts and severe heatwaves affecting Europe. Drought conditions in central and northern Europe caused yield reductions up to 50\% for the main crops, yet wet conditions in southern Europe saw yield gains up to 34\%, both with respect to the previous 5-year mean  \citep{buras2019quantifying,toreti2019exceptional}.
To tackle year to year seasonality and the limited availability of \textit{in-situ} data, classification methodological developments are still needed to apply the proposed approach to other years.

\section{Conclusions}
Arable land represents almost half of total EU area and its monitoring is crucial for efficient environmental, agriculture and climate policies' implementation. In this study, a framework to derive a 10-m crop type from S1 at continental level has been designed, implemented and assessed rigorously. The proposed framework opens avenues to develop a synergistic robust information system to monitor agriculture consistently both at fine spatial-and-temporal scale over large areas. This demonstrates the current momentum combining the unique Copernicus Sentinel fleet, high quality massive in-situ data along with cloud computing. The method and data developed will serve communities ranging from scientific modellers to policy makers.



\section{Author contributions}
R.D and A.V conceptualized the study and designed the methodology, R.D., A.V., G.L., P.K. processed the data. R.D., A.V., G.L., P.K., M. M. , M. V.  analyzed the data and wrote the paper.

\section{Acknowledgements}
The authors would like to thank Kostas Anastasakis as jrc-cbm developer,
Javier Sanchez Lopez for his agronomicic advices, the JRC JEODPP colleagues for their continuous support and the Google Earth Engine team for their technical help.

\bibliography{bibliography.bib}

\begin{thebibliography}{67}
\expandafter\ifx\csname natexlab\endcsname\relax\def\natexlab#1{#1}\fi
\providecommand{\url}[1]{\texttt{#1}}
\providecommand{\href}[2]{#2}
\providecommand{\path}[1]{#1}
\providecommand{\DOIprefix}{doi:}
\providecommand{\ArXivprefix}{arXiv:}
\providecommand{\URLprefix}{URL: }
\providecommand{\Pubmedprefix}{pmid:}
\providecommand{\doi}[1]{\href{http://dx.doi.org/#1}{\path{#1}}}
\providecommand{\Pubmed}[1]{\href{pmid:#1}{\path{#1}}}
\providecommand{\bibinfo}[2]{#2}
\ifx\xfnm\relax \def\xfnm[#1]{\unskip,\space#1}\fi
\bibitem[{Belgiu \& Csillik(2018)}]{belgiu2018sentinel}
\bibinfo{author}{Belgiu, M.}, \& \bibinfo{author}{Csillik, O.}
  (\bibinfo{year}{2018}).
\newblock \bibinfo{title}{Sentinel-2 cropland mapping using pixel-based and
  object-based time-weighted dynamic time warping analysis}.
\newblock {\it \bibinfo{journal}{Remote sensing of environment}\/},  {\it
  \bibinfo{volume}{204}\/}, \bibinfo{pages}{509--523}.
\bibitem[{Belgiu \& Dr{\u{a}}gu{\c{t}}(2016)}]{belgiu2016random}
\bibinfo{author}{Belgiu, M.}, \& \bibinfo{author}{Dr{\u{a}}gu{\c{t}}, L.}
  (\bibinfo{year}{2016}).
\newblock \bibinfo{title}{Random forest in remote sensing: A review of
  applications and future directions}.
\newblock {\it \bibinfo{journal}{ISPRS Journal of Photogrammetry and Remote
  Sensing}\/},  {\it \bibinfo{volume}{114}\/}, \bibinfo{pages}{24--31}.
\bibitem[{Breiman(2001)}]{breiman2001random}
\bibinfo{author}{Breiman, L.} (\bibinfo{year}{2001}).
\newblock \bibinfo{title}{Random forests}.
\newblock {\it \bibinfo{journal}{Machine learning}\/},  {\it
  \bibinfo{volume}{45}\/}, \bibinfo{pages}{5--32}.
\bibitem[{Buras et~al.(2019)Buras, Rammig \& Zang}]{buras2019quantifying}
\bibinfo{author}{Buras, A.}, \bibinfo{author}{Rammig, A.}, \&
  \bibinfo{author}{Zang, C.~S.} (\bibinfo{year}{2019}).
\newblock \bibinfo{title}{Quantifying impacts of the drought 2018 on european
  ecosystems in comparison to 2003}.
\newblock {\it \bibinfo{journal}{arXiv preprint arXiv:1906.08605}\/}, .
\bibitem[{Chakhar et~al.(2021)Chakhar, Hern{\'a}ndez-L{\'o}pez, Ballesteros \&
  Moreno}]{chakhar2021improving}
\bibinfo{author}{Chakhar, A.}, \bibinfo{author}{Hern{\'a}ndez-L{\'o}pez, D.},
  \bibinfo{author}{Ballesteros, R.}, \& \bibinfo{author}{Moreno, M.~A.}
  (\bibinfo{year}{2021}).
\newblock \bibinfo{title}{Improving the accuracy of multiple algorithms for
  crop classification by integrating sentinel-1 observations with sentinel-2
  data}.
\newblock {\it \bibinfo{journal}{Remote Sensing}\/},  {\it
  \bibinfo{volume}{13}\/}, \bibinfo{pages}{243}.
\bibitem[{Clauss et~al.(2018)Clauss, Ottinger \&
  K{\"u}nzer}]{clauss2018mapping}
\bibinfo{author}{Clauss, K.}, \bibinfo{author}{Ottinger, M.}, \&
  \bibinfo{author}{K{\"u}nzer, C.} (\bibinfo{year}{2018}).
\newblock \bibinfo{title}{Mapping rice areas with sentinel-1 time series and
  superpixel segmentation}.
\newblock {\it \bibinfo{journal}{International Journal of Remote Sensing}\/},
  {\it \bibinfo{volume}{39}\/}, \bibinfo{pages}{1399--1420}.
\bibitem[{Close et~al.(2018)Close, Benjamin, Petit, Fripiat \&
  Hallot}]{Close2018}
\bibinfo{author}{Close, O.}, \bibinfo{author}{Benjamin, B.},
  \bibinfo{author}{Petit, S.}, \bibinfo{author}{Fripiat, X.}, \&
  \bibinfo{author}{Hallot, E.} (\bibinfo{year}{2018}).
\newblock \bibinfo{title}{{Use of Sentinel-2 and LUCAS Database for the
  Inventory of Land Use, Land Use Change, and Forestry in Wallonia, Belgium}}.
\newblock {\it \bibinfo{journal}{Land}\/},  {\it \bibinfo{volume}{7}\/},
  \bibinfo{pages}{154}. \URLprefix \url{http://www.mdpi.com/2073-445X/7/4/154}.
  \DOIprefix\doi{10.3390/land7040154}.
\bibitem[{Corbane et~al.(2020)Corbane, Sabo, Syrris, Kemper, Politis, Pesaresi,
  Soille \& Osé}]{corbaneESR}
\bibinfo{author}{Corbane, C.}, \bibinfo{author}{Sabo, F.},
  \bibinfo{author}{Syrris, V.}, \bibinfo{author}{Kemper, T.},
  \bibinfo{author}{Politis, P.}, \bibinfo{author}{Pesaresi, M.},
  \bibinfo{author}{Soille, P.}, \& \bibinfo{author}{Osé, K.}
  (\bibinfo{year}{2020}).
\newblock \bibinfo{title}{Application of the symbolic machine learning to
  copernicus vhr imagery: The european settlement map}.
\newblock {\it \bibinfo{journal}{IEEE Geoscience and Remote Sensing
  Letters}\/},  {\it \bibinfo{volume}{17}\/}, \bibinfo{pages}{1153--1157}.
  \DOIprefix\doi{10.1109/LGRS.2019.2942131}.
\bibitem[{d'Andrimont et~al.(2018{\natexlab{a}})d'Andrimont, Lemoine \& Van~der
  Velde}]{d2018targeted}
\bibinfo{author}{d'Andrimont, R.}, \bibinfo{author}{Lemoine, G.}, \&
  \bibinfo{author}{Van~der Velde, M.} (\bibinfo{year}{2018}{\natexlab{a}}).
\newblock \bibinfo{title}{Targeted grassland monitoring at parcel level using
  sentinels, street-level images and field observations}.
\newblock {\it \bibinfo{journal}{Remote Sensing}\/},  {\it
  \bibinfo{volume}{10}\/}, \bibinfo{pages}{1300}.
\bibitem[{d'Andrimont et~al.(2020{\natexlab{a}})d'Andrimont, Taymans, Lemoine,
  Ceglar, Yordanov \& van~der Velde}]{d2020detecting}
\bibinfo{author}{d'Andrimont, R.}, \bibinfo{author}{Taymans, M.},
  \bibinfo{author}{Lemoine, G.}, \bibinfo{author}{Ceglar, A.},
  \bibinfo{author}{Yordanov, M.}, \& \bibinfo{author}{van~der Velde, M.}
  (\bibinfo{year}{2020}{\natexlab{a}}).
\newblock \bibinfo{title}{Detecting flowering phenology in oil seed rape
  parcels with sentinel-1 and-2 time series}.
\newblock {\it \bibinfo{journal}{Remote sensing of environment}\/},  {\it
  \bibinfo{volume}{239}\/}, \bibinfo{pages}{111660}.
\bibitem[{d'Andrimont et~al.(2021)d'Andrimont, Verhegghen, Meroni, Lemoine,
  Strobl, Eiselt, Yordanov, Martinez-Sanchez \& van~der
  Velde}]{dandrimont-essd}
\bibinfo{author}{d'Andrimont, R.}, \bibinfo{author}{Verhegghen, A.},
  \bibinfo{author}{Meroni, M.}, \bibinfo{author}{Lemoine, G.},
  \bibinfo{author}{Strobl, P.}, \bibinfo{author}{Eiselt, B.},
  \bibinfo{author}{Yordanov, M.}, \bibinfo{author}{Martinez-Sanchez, L.}, \&
  \bibinfo{author}{van~der Velde, M.} (\bibinfo{year}{2021}).
\newblock \bibinfo{title}{Lucas copernicus 2018: Earth-observation-relevant in
  situ data on land cover and use throughout the european union}.
\newblock {\it \bibinfo{journal}{Earth System Science Data}\/},  {\it
  \bibinfo{volume}{13}\/}, \bibinfo{pages}{1119--1133}.
\bibitem[{d'Andrimont et~al.(2018{\natexlab{b}})d'Andrimont, Yordanov, Lemoine,
  Yoong, Nikel \& van~der Velde}]{d2018crowdsourced}
\bibinfo{author}{d'Andrimont, R.}, \bibinfo{author}{Yordanov, M.},
  \bibinfo{author}{Lemoine, G.}, \bibinfo{author}{Yoong, J.},
  \bibinfo{author}{Nikel, K.}, \& \bibinfo{author}{van~der Velde, M.}
  (\bibinfo{year}{2018}{\natexlab{b}}).
\newblock \bibinfo{title}{Crowdsourced street-level imagery as a potential
  source of in-situ data for crop monitoring}.
\newblock {\it \bibinfo{journal}{Land}\/},  {\it \bibinfo{volume}{7}\/},
  \bibinfo{pages}{127}.
\bibitem[{d'Andrimont et~al.(2020{\natexlab{b}})d'Andrimont, Yordanov,
  Martinez-Sanchez, Eiselt, Palmieri, Dominici, Gallego, Reuter, Joebges,
  Lemoine et~al.}]{d2020harmonised}
\bibinfo{author}{d'Andrimont, R.}, \bibinfo{author}{Yordanov, M.},
  \bibinfo{author}{Martinez-Sanchez, L.}, \bibinfo{author}{Eiselt, B.},
  \bibinfo{author}{Palmieri, A.}, \bibinfo{author}{Dominici, P.},
  \bibinfo{author}{Gallego, J.}, \bibinfo{author}{Reuter, H.~I.},
  \bibinfo{author}{Joebges, C.}, \bibinfo{author}{Lemoine, G.} et~al.
  (\bibinfo{year}{2020}{\natexlab{b}}).
\newblock \bibinfo{title}{Harmonised lucas in-situ land cover and use database
  for field surveys from 2006 to 2018 in the european union}.
\newblock {\it \bibinfo{journal}{Scientific Data}\/},  {\it
  \bibinfo{volume}{7}\/}, \bibinfo{pages}{1--15}.
\bibitem[{De~Vroey et~al.(2021)De~Vroey, Radoux \& Defourny}]{de2021grassland}
\bibinfo{author}{De~Vroey, M.}, \bibinfo{author}{Radoux, J.}, \&
  \bibinfo{author}{Defourny, P.} (\bibinfo{year}{2021}).
\newblock \bibinfo{title}{Grassland mowing detection using sentinel-1 time
  series: Potential and limitations}.
\newblock {\it \bibinfo{journal}{Remote Sensing}\/},  {\it
  \bibinfo{volume}{13}\/}, \bibinfo{pages}{348}.
\bibitem[{Defourny et~al.(2019)Defourny, Bontemps, Bellemans, Cara, Dedieu,
  Guzzonato, Hagolle, Inglada, Nicola, Rabaute et~al.}]{defourny2019near}
\bibinfo{author}{Defourny, P.}, \bibinfo{author}{Bontemps, S.},
  \bibinfo{author}{Bellemans, N.}, \bibinfo{author}{Cara, C.},
  \bibinfo{author}{Dedieu, G.}, \bibinfo{author}{Guzzonato, E.},
  \bibinfo{author}{Hagolle, O.}, \bibinfo{author}{Inglada, J.},
  \bibinfo{author}{Nicola, L.}, \bibinfo{author}{Rabaute, T.} et~al.
  (\bibinfo{year}{2019}).
\newblock \bibinfo{title}{Near real-time agriculture monitoring at national
  scale at parcel resolution: Performance assessment of the sen2-agri automated
  system in various cropping systems around the world}.
\newblock {\it \bibinfo{journal}{Remote Sensing of Environment}\/},  {\it
  \bibinfo{volume}{221}\/}, \bibinfo{pages}{551--568}.
\bibitem[{Dinerstein et~al.(2017)Dinerstein, Olson, Joshi, Vynne, Burgess,
  Wikramanayake, Hahn, Palminteri, Hedao, Noss
  et~al.}]{dinerstein2017ecoregion}
\bibinfo{author}{Dinerstein, E.}, \bibinfo{author}{Olson, D.},
  \bibinfo{author}{Joshi, A.}, \bibinfo{author}{Vynne, C.},
  \bibinfo{author}{Burgess, N.~D.}, \bibinfo{author}{Wikramanayake, E.},
  \bibinfo{author}{Hahn, N.}, \bibinfo{author}{Palminteri, S.},
  \bibinfo{author}{Hedao, P.}, \bibinfo{author}{Noss, R.} et~al.
  (\bibinfo{year}{2017}).
\newblock \bibinfo{title}{An ecoregion-based approach to protecting half the
  terrestrial realm}.
\newblock {\it \bibinfo{journal}{BioScience}\/},  {\it \bibinfo{volume}{67}\/},
  \bibinfo{pages}{534--545}.
\bibitem[{Dobson \& Ulaby(1981)}]{dobson1981microwave}
\bibinfo{author}{Dobson, M.~C.}, \& \bibinfo{author}{Ulaby, F.}
  (\bibinfo{year}{1981}).
\newblock \bibinfo{title}{Microwave backscatter dependence on surface
  roughness, soil moisture, and soil texture: Part iii-soil tension}.
\newblock {\it \bibinfo{journal}{IEEE Transactions on Geoscience and Remote
  Sensing}\/},  (pp. \bibinfo{pages}{51--61}).
\bibitem[{EEA(2018)}]{eea2018corine}
\bibinfo{author}{EEA} (\bibinfo{year}{2018}).
\newblock \bibinfo{title}{Corine land cover (clc) 2018, version 20b2}.
\newblock {\it \bibinfo{journal}{Release Date: 21-12-2018}\/}, .
\bibitem[{{European Commission}(2019)}]{agrifoodtrade2018}
\bibinfo{author}{{European Commission}} (\bibinfo{year}{2019}).
\newblock \bibinfo{title}{Agri-food trade in 2018: another successful year for
  agri-food trade}.
\newblock
  \bibinfo{howpublished}{\url{https://ec.europa.eu/info/sites/info/files/food-farming-fisheries/news/documents/agri-food-trade-2018_en.pdf}}.
\newblock \bibinfo{note}{(Accessed on 03/16/2021)}.
\bibitem[{Eurostat(2018{\natexlab{a}})}]{LUCAS:online}
\bibinfo{author}{Eurostat} (\bibinfo{year}{2018}{\natexlab{a}}).
\newblock \bibinfo{title}{Lucas web site}.
\newblock
  \bibinfo{howpublished}{\url{https://ec.europa.eu/eurostat/web/lucas}}.
\newblock \bibinfo{note}{(Accessed on 08/30/2018)}.
\bibitem[{Eurostat(2018{\natexlab{b}})}]{c12018}
\bibinfo{author}{Eurostat} (\bibinfo{year}{2018}{\natexlab{b}}).
\newblock \bibinfo{title}{Technical reference document c-1: Instructions for
  surveyors}.
\newblock
  \bibinfo{howpublished}{\url{https://ec.europa.eu/eurostat/documents/205002/8072634/LUCAS2018-C1-Instructions.pdf}}.
\newblock \bibinfo{note}{(Accessed on 07/30/2019)}.
\bibitem[{Eurostat(2018{\natexlab{c}})}]{c32018}
\bibinfo{author}{Eurostat} (\bibinfo{year}{2018}{\natexlab{c}}).
\newblock \bibinfo{title}{Technical reference document c-3: Classification}.
\newblock
  \bibinfo{howpublished}{\url{https://ec.europa.eu/eurostat/documents/205002/8072634/LUCAS2018-C3-Classification.pdf}}.
\newblock \bibinfo{note}{(Accessed on 07/30/2019)}.
\bibitem[{Gallego(2004)}]{gallego2004remote}
\bibinfo{author}{Gallego, F.~J.} (\bibinfo{year}{2004}).
\newblock \bibinfo{title}{Remote sensing and land cover area estimation}.
\newblock {\it \bibinfo{journal}{International Journal of Remote Sensing}\/},
  {\it \bibinfo{volume}{25}\/}, \bibinfo{pages}{3019--3047}.
\bibitem[{Gallego \& Bamps(2008)}]{gallego2008using}
\bibinfo{author}{Gallego, J.}, \& \bibinfo{author}{Bamps, C.}
  (\bibinfo{year}{2008}).
\newblock \bibinfo{title}{Using corine land cover and the point survey lucas
  for area estimation}.
\newblock {\it \bibinfo{journal}{International Journal of Applied Earth
  Observation and Geoinformation}\/},  {\it \bibinfo{volume}{10}\/},
  \bibinfo{pages}{467--475}.
\bibitem[{Gallego \& Delinc{\'e}(2010)}]{gallego2010european}
\bibinfo{author}{Gallego, J.}, \& \bibinfo{author}{Delinc{\'e}, J.}
  (\bibinfo{year}{2010}).
\newblock \bibinfo{title}{The european land use and cover area-frame
  statistical survey}.
\newblock {\it \bibinfo{journal}{Agricultural survey methods}\/},  (pp.
  \bibinfo{pages}{149--168}).
\bibitem[{Gao et~al.(2018)Gao, Wang, Wang, Zhu, Tang, Shen \&
  Zhu}]{gao2018crop}
\bibinfo{author}{Gao, H.}, \bibinfo{author}{Wang, C.}, \bibinfo{author}{Wang,
  G.}, \bibinfo{author}{Zhu, J.}, \bibinfo{author}{Tang, Y.},
  \bibinfo{author}{Shen, P.}, \& \bibinfo{author}{Zhu, Z.}
  (\bibinfo{year}{2018}).
\newblock \bibinfo{title}{A crop classification method integrating gf-3 polsar
  and sentinel-2a optical data in the dongting lake basin}.
\newblock {\it \bibinfo{journal}{Sensors}\/},  {\it \bibinfo{volume}{18}\/},
  \bibinfo{pages}{3139}.
\bibitem[{Gorelick et~al.(2017)Gorelick, Hancher, Dixon, Ilyushchenko, Thau \&
  Moore}]{gorelick2017google}
\bibinfo{author}{Gorelick, N.}, \bibinfo{author}{Hancher, M.},
  \bibinfo{author}{Dixon, M.}, \bibinfo{author}{Ilyushchenko, S.},
  \bibinfo{author}{Thau, D.}, \& \bibinfo{author}{Moore, R.}
  (\bibinfo{year}{2017}).
\newblock \bibinfo{title}{Google earth engine: Planetary-scale geospatial
  analysis for everyone}.
\newblock {\it \bibinfo{journal}{Remote Sensing of Environment}\/}, .
\bibitem[{Immitzer et~al.(2016)Immitzer, Vuolo \&
  Atzberger}]{immitzer2016first}
\bibinfo{author}{Immitzer, M.}, \bibinfo{author}{Vuolo, F.}, \&
  \bibinfo{author}{Atzberger, C.} (\bibinfo{year}{2016}).
\newblock \bibinfo{title}{First experience with sentinel-2 data for crop and
  tree species classifications in central europe}.
\newblock {\it \bibinfo{journal}{Remote Sensing}\/},  {\it
  \bibinfo{volume}{8}\/}, \bibinfo{pages}{166}.
\bibitem[{Inglada et~al.(2017)Inglada, Vincent, Arias, Tardy, Morin \&
  Rodes}]{inglada2017operational}
\bibinfo{author}{Inglada, J.}, \bibinfo{author}{Vincent, A.},
  \bibinfo{author}{Arias, M.}, \bibinfo{author}{Tardy, B.},
  \bibinfo{author}{Morin, D.}, \& \bibinfo{author}{Rodes, I.}
  (\bibinfo{year}{2017}).
\newblock \bibinfo{title}{Operational high resolution land cover map production
  at the country scale using satellite image time series}.
\newblock {\it \bibinfo{journal}{Remote Sensing}\/},  {\it
  \bibinfo{volume}{9}\/}, \bibinfo{pages}{95}.
\bibitem[{Joly et~al.(2016)Joly, Bonnet, Go{\"e}au, Barbe, Selmi, Champ,
  Dufour-Kowalski, Affouard, Carr{\'e}, Molino et~al.}]{joly2016look}
\bibinfo{author}{Joly, A.}, \bibinfo{author}{Bonnet, P.},
  \bibinfo{author}{Go{\"e}au, H.}, \bibinfo{author}{Barbe, J.},
  \bibinfo{author}{Selmi, S.}, \bibinfo{author}{Champ, J.},
  \bibinfo{author}{Dufour-Kowalski, S.}, \bibinfo{author}{Affouard, A.},
  \bibinfo{author}{Carr{\'e}, J.}, \bibinfo{author}{Molino, J.-F.} et~al.
  (\bibinfo{year}{2016}).
\newblock \bibinfo{title}{A look inside the pl@ ntnet experience}.
\newblock {\it \bibinfo{journal}{Multimedia Systems}\/},  {\it
  \bibinfo{volume}{22}\/}, \bibinfo{pages}{751--766}.
\bibitem[{Kenduiywo et~al.(2018)Kenduiywo, Bargiel \&
  Soergel}]{kenduiywo2018crop}
\bibinfo{author}{Kenduiywo, B.~K.}, \bibinfo{author}{Bargiel, D.}, \&
  \bibinfo{author}{Soergel, U.} (\bibinfo{year}{2018}).
\newblock \bibinfo{title}{Crop-type mapping from a sequence of sentinel 1
  images}.
\newblock {\it \bibinfo{journal}{International Journal of Remote Sensing}\/},
  {\it \bibinfo{volume}{39}\/}, \bibinfo{pages}{6383--6404}.
\bibitem[{Kussul et~al.(2017)Kussul, Lavreniuk, Skakun \&
  Shelestov}]{kussul2017deep}
\bibinfo{author}{Kussul, N.}, \bibinfo{author}{Lavreniuk, M.},
  \bibinfo{author}{Skakun, S.}, \& \bibinfo{author}{Shelestov, A.}
  (\bibinfo{year}{2017}).
\newblock \bibinfo{title}{Deep learning classification of land cover and crop
  types using remote sensing data}.
\newblock {\it \bibinfo{journal}{IEEE Geoscience and Remote Sensing
  Letters}\/},  {\it \bibinfo{volume}{14}\/}, \bibinfo{pages}{778--782}.
\bibitem[{Kussul et~al.(2018)Kussul, Mykola, Shelestov \&
  Skakun}]{kussul2018crop}
\bibinfo{author}{Kussul, N.}, \bibinfo{author}{Mykola, L.},
  \bibinfo{author}{Shelestov, A.}, \& \bibinfo{author}{Skakun, S.}
  (\bibinfo{year}{2018}).
\newblock \bibinfo{title}{Crop inventory at regional scale in ukraine:
  developing in season and end of season crop maps with multi-temporal optical
  and sar satellite imagery}.
\newblock {\it \bibinfo{journal}{European Journal of Remote Sensing}\/},  {\it
  \bibinfo{volume}{51}\/}, \bibinfo{pages}{627--636}.
\bibitem[{Laso~Bayas et~al.(2020)Laso~Bayas, See, Bartl, Sturn, Karner, Fraisl,
  Moorthy, Busch, van~der Velde \& Fritz}]{laso2020crowdsourcing}
\bibinfo{author}{Laso~Bayas, J.~C.}, \bibinfo{author}{See, L.},
  \bibinfo{author}{Bartl, H.}, \bibinfo{author}{Sturn, T.},
  \bibinfo{author}{Karner, M.}, \bibinfo{author}{Fraisl, D.},
  \bibinfo{author}{Moorthy, I.}, \bibinfo{author}{Busch, M.},
  \bibinfo{author}{van~der Velde, M.}, \& \bibinfo{author}{Fritz, S.}
  (\bibinfo{year}{2020}).
\newblock \bibinfo{title}{Crowdsourcing lucas: Citizens generating reference
  land cover and land use data with a mobile app}.
\newblock {\it \bibinfo{journal}{Land}\/},  {\it \bibinfo{volume}{9}\/},
  \bibinfo{pages}{446}.
\bibitem[{Lemoine(2017)}]{lemoine2017GSARS}
\bibinfo{author}{Lemoine, G.} (\bibinfo{year}{2017}).
\newblock {\it \bibinfo{title}{Data access and data analysis software.}\/}.
\newblock (\bibinfo{edition}{1st} ed.).
\newblock \bibinfo{address}{Rome}: \bibinfo{publisher}{Handbook on Remote
  Sensing for Agricultural Statistics (Chapter 1). Handbook of the Global
  Strategy to improve Agricultural and Rural Statistics (GSARS)}.
\newblock \DOIprefix\doi{10.13140/RG.2.2.13259.69920}.
\bibitem[{Lemoine et~al.(2019)Lemoine, Devos, Milenov \&
  d'Andrimont}]{lemoine2019dnn}
\bibinfo{author}{Lemoine, G.}, \bibinfo{author}{Devos, W.},
  \bibinfo{author}{Milenov, P.}, \& \bibinfo{author}{d'Andrimont, R.}
  (\bibinfo{year}{2019}).
\newblock \bibinfo{title}{Machine learning for crop type identification using
  country-wide, consistent sentinel-1 time series}.
\newblock In {\it \bibinfo{booktitle}{Proc. 2019 Conference on Big Data from
  Space (BiDS '19)}\/} (pp. \bibinfo{pages}{81--74}).
\bibitem[{Mack et~al.(2017)Mack, Leinenkugel, Kuenzer \& Dech}]{mack2017semi}
\bibinfo{author}{Mack, B.}, \bibinfo{author}{Leinenkugel, P.},
  \bibinfo{author}{Kuenzer, C.}, \& \bibinfo{author}{Dech, S.}
  (\bibinfo{year}{2017}).
\newblock \bibinfo{title}{A semi-automated approach for the generation of a new
  land use and land cover product for germany based on landsat time-series and
  lucas in-situ data}.
\newblock {\it \bibinfo{journal}{Remote Sensing Letters}\/},  {\it
  \bibinfo{volume}{8}\/}, \bibinfo{pages}{244--253}.
\bibitem[{Massey et~al.(2018)Massey, Sankey, Yadav, Congalton \&
  Tilton}]{massey2018integrating}
\bibinfo{author}{Massey, R.}, \bibinfo{author}{Sankey, T.~T.},
  \bibinfo{author}{Yadav, K.}, \bibinfo{author}{Congalton, R.~G.}, \&
  \bibinfo{author}{Tilton, J.~C.} (\bibinfo{year}{2018}).
\newblock \bibinfo{title}{Integrating cloud-based workflows in
  continental-scale cropland extent classification}.
\newblock {\it \bibinfo{journal}{Remote Sensing of Environment}\/},  {\it
  \bibinfo{volume}{219}\/}, \bibinfo{pages}{162--179}.
\bibitem[{McNairn \& Brisco(2004)}]{mcnairn2004application}
\bibinfo{author}{McNairn, H.}, \& \bibinfo{author}{Brisco, B.}
  (\bibinfo{year}{2004}).
\newblock \bibinfo{title}{The application of c-band polarimetric sar for
  agriculture: A review}.
\newblock {\it \bibinfo{journal}{Canadian Journal of Remote Sensing}\/},  {\it
  \bibinfo{volume}{30}\/}, \bibinfo{pages}{525--542}.
\bibitem[{Merlos \& Hijmans(2020)}]{merlos2020scale}
\bibinfo{author}{Merlos, F.~A.}, \& \bibinfo{author}{Hijmans, R.~J.}
  (\bibinfo{year}{2020}).
\newblock \bibinfo{title}{The scale dependency of spatial crop species
  diversity and its relation to temporal diversity}.
\newblock {\it \bibinfo{journal}{Proceedings of the National Academy of
  Sciences}\/},  {\it \bibinfo{volume}{117}\/}, \bibinfo{pages}{26176--26182}.
\bibitem[{Meroni et~al.(2021)Meroni, d'Andrimont, Vrieling, Fasbender, Lemoine,
  Rembold, Seguini \& Verhegghen}]{meroni2021comparing}
\bibinfo{author}{Meroni, M.}, \bibinfo{author}{d'Andrimont, R.},
  \bibinfo{author}{Vrieling, A.}, \bibinfo{author}{Fasbender, D.},
  \bibinfo{author}{Lemoine, G.}, \bibinfo{author}{Rembold, F.},
  \bibinfo{author}{Seguini, L.}, \& \bibinfo{author}{Verhegghen, A.}
  (\bibinfo{year}{2021}).
\newblock \bibinfo{title}{Comparing land surface phenology of major european
  crops as derived from sar and multispectral data of sentinel-1 and-2}.
\newblock {\it \bibinfo{journal}{Remote sensing of environment}\/},  {\it
  \bibinfo{volume}{253}\/}, \bibinfo{pages}{112232}.
\bibitem[{Ndikumana et~al.(2018)Ndikumana, Ho~Tong~Minh, Baghdadi, Courault \&
  Hossard}]{ndikumana2018deep}
\bibinfo{author}{Ndikumana, E.}, \bibinfo{author}{Ho~Tong~Minh, D.},
  \bibinfo{author}{Baghdadi, N.}, \bibinfo{author}{Courault, D.}, \&
  \bibinfo{author}{Hossard, L.} (\bibinfo{year}{2018}).
\newblock \bibinfo{title}{Deep recurrent neural network for agricultural
  classification using multitemporal sar sentinel-1 for camargue, france}.
\newblock {\it \bibinfo{journal}{Remote Sensing}\/},  {\it
  \bibinfo{volume}{10}\/}, \bibinfo{pages}{1217}.
\bibitem[{Olofsson et~al.(2014)Olofsson, Foody, Herold, Stehman, Woodcock \&
  Wulder}]{Olofsson2014a}
\bibinfo{author}{Olofsson, P.}, \bibinfo{author}{Foody, G.~M.},
  \bibinfo{author}{Herold, M.}, \bibinfo{author}{Stehman, S.~V.},
  \bibinfo{author}{Woodcock, C.~E.}, \& \bibinfo{author}{Wulder, M.~A.}
  (\bibinfo{year}{2014}).
\newblock \bibinfo{title}{{Good practices for estimating area and assessing
  accuracy of land change}}.
\newblock {\it \bibinfo{journal}{Remote Sensing of Environment}\/},  {\it
  \bibinfo{volume}{148}\/}, \bibinfo{pages}{42--57}. \URLprefix
  \url{http://www.sciencedirect.com/science/article/pii/S0034425714000704
  http://dx.doi.org/10.1016/j.rse.2014.02.015}.
  \DOIprefix\doi{10.1016/j.rse.2014.02.015}.
\bibitem[{Orynbaikyzy et~al.(2020)Orynbaikyzy, Gessner, Mack \&
  Conrad}]{orynbaikyzy2020crop}
\bibinfo{author}{Orynbaikyzy, A.}, \bibinfo{author}{Gessner, U.},
  \bibinfo{author}{Mack, B.}, \& \bibinfo{author}{Conrad, C.}
  (\bibinfo{year}{2020}).
\newblock \bibinfo{title}{Crop type classification using fusion of sentinel-1
  and sentinel-2 data: Assessing the impact of feature selection, optical data
  availability, and parcel sizes on the accuracies}.
\newblock {\it \bibinfo{journal}{Remote Sensing}\/},  {\it
  \bibinfo{volume}{12}\/}, \bibinfo{pages}{2779}.
\bibitem[{Pedregosa et~al.(2011)Pedregosa, Varoquaux, Gramfort, Michel,
  Thirion, Grisel, Blondel, Prettenhofer, Weiss, Dubourg, Vanderplas, Passos,
  Cournapeau, Brucher, Perrot \& Duchesnay}]{scikit-learn}
\bibinfo{author}{Pedregosa, F.}, \bibinfo{author}{Varoquaux, G.},
  \bibinfo{author}{Gramfort, A.}, \bibinfo{author}{Michel, V.},
  \bibinfo{author}{Thirion, B.}, \bibinfo{author}{Grisel, O.},
  \bibinfo{author}{Blondel, M.}, \bibinfo{author}{Prettenhofer, P.},
  \bibinfo{author}{Weiss, R.}, \bibinfo{author}{Dubourg, V.},
  \bibinfo{author}{Vanderplas, J.}, \bibinfo{author}{Passos, A.},
  \bibinfo{author}{Cournapeau, D.}, \bibinfo{author}{Brucher, M.},
  \bibinfo{author}{Perrot, M.}, \& \bibinfo{author}{Duchesnay, E.}
  (\bibinfo{year}{2011}).
\newblock \bibinfo{title}{Scikit-learn: Machine learning in {P}ython}.
\newblock {\it \bibinfo{journal}{Journal of Machine Learning Research}\/},
  {\it \bibinfo{volume}{12}\/}, \bibinfo{pages}{2825--2830}.
\bibitem[{Pekel et~al.(2016)Pekel, Cottam, Gorelick \& Belward}]{Pekel2016}
\bibinfo{author}{Pekel, J.-F.}, \bibinfo{author}{Cottam, A.},
  \bibinfo{author}{Gorelick, N.}, \& \bibinfo{author}{Belward, A.~S.}
  (\bibinfo{year}{2016}).
\newblock \bibinfo{title}{{High-resolution mapping of global surface water and
  its long-term changes}}.
\newblock {\it \bibinfo{journal}{Nature}\/},  {\it \bibinfo{volume}{540}\/},
  \bibinfo{pages}{418--422}. \URLprefix
  \url{http://www.nature.com/articles/nature20584}.
  \DOIprefix\doi{10.1038/nature20584}.
\bibitem[{Pelletier et~al.(2016)Pelletier, Valero, Inglada, Champion \&
  Dedieu}]{pelletier2016assessing}
\bibinfo{author}{Pelletier, C.}, \bibinfo{author}{Valero, S.},
  \bibinfo{author}{Inglada, J.}, \bibinfo{author}{Champion, N.}, \&
  \bibinfo{author}{Dedieu, G.} (\bibinfo{year}{2016}).
\newblock \bibinfo{title}{Assessing the robustness of random forests to map
  land cover with high resolution satellite image time series over large
  areas}.
\newblock {\it \bibinfo{journal}{Remote Sensing of Environment}\/},  {\it
  \bibinfo{volume}{187}\/}, \bibinfo{pages}{156--168}.
\bibitem[{Pflugmacher et~al.(2019)Pflugmacher, Rabe, Peters \&
  Hostert}]{pflugmacher2019mapping}
\bibinfo{author}{Pflugmacher, D.}, \bibinfo{author}{Rabe, A.},
  \bibinfo{author}{Peters, M.}, \& \bibinfo{author}{Hostert, P.}
  (\bibinfo{year}{2019}).
\newblock \bibinfo{title}{Mapping pan-european land cover using landsat
  spectral-temporal metrics and the european lucas survey}.
\newblock {\it \bibinfo{journal}{Remote Sensing of Environment}\/},  {\it
  \bibinfo{volume}{221}\/}, \bibinfo{pages}{583--595}.
\bibitem[{Sabo et~al.(2019)Sabo, Corbane, Politis, Pesaresi \&
  Kemper}]{sabo2020}
\bibinfo{author}{Sabo, F.}, \bibinfo{author}{Corbane, C.},
  \bibinfo{author}{Politis, P.}, \bibinfo{author}{Pesaresi, M.}, \&
  \bibinfo{author}{Kemper, T.} (\bibinfo{year}{2019}).
\newblock \bibinfo{title}{Update and improvement of the european settlement
  map}.
\newblock In {\it \bibinfo{booktitle}{2019 Joint Urban Remote Sensing Event
  (JURSE)}\/} (pp. \bibinfo{pages}{1--4}).
\newblock \DOIprefix\doi{10.1109/JURSE.2019.8808933}.
\bibitem[{Scarno et~al.(2018)Scarno, Ballin, Barcaroli \&
  Masselli}]{scarno2018lucas}
\bibinfo{author}{Scarno, M.}, \bibinfo{author}{Ballin, M.},
  \bibinfo{author}{Barcaroli, G.}, \& \bibinfo{author}{Masselli, M.}
  (\bibinfo{year}{2018}).
\newblock \bibinfo{title}{{Redesign sample for Land Use/Cover Area frame Survey
  (LUCAS) 2018}}.
\newblock {\it \bibinfo{journal}{Statistical Working Papers}\/}, .
  \DOIprefix\doi{10.2785/132365}.
\bibitem[{Small(2011)}]{small2011flattening}
\bibinfo{author}{Small, D.} (\bibinfo{year}{2011}).
\newblock \bibinfo{title}{Flattening gamma: Radiometric terrain correction for
  sar imagery}.
\newblock {\it \bibinfo{journal}{IEEE Transactions on Geoscience and Remote
  Sensing}\/},  {\it \bibinfo{volume}{49}\/}, \bibinfo{pages}{3081--3093}.
\bibitem[{Soille et~al.(2018)Soille, Burger, {De Marchi}, Kempeneers,
  Rodriguez, Syrris \& Vasilev}]{Soille2018}
\bibinfo{author}{Soille, P.}, \bibinfo{author}{Burger, A.},
  \bibinfo{author}{{De Marchi}, D.}, \bibinfo{author}{Kempeneers, P.},
  \bibinfo{author}{Rodriguez, D.}, \bibinfo{author}{Syrris, V.}, \&
  \bibinfo{author}{Vasilev, V.} (\bibinfo{year}{2018}).
\newblock \bibinfo{title}{{A versatile data-intensive computing platform for
  information retrieval from big geospatial data}}.
\newblock {\it \bibinfo{journal}{Future Generation Computer Systems}\/},  {\it
  \bibinfo{volume}{81}\/}, \bibinfo{pages}{30--40}. \URLprefix
  \url{https://doi.org/10.1016/j.future.2017.11.007}.
  \DOIprefix\doi{10.1016/j.future.2017.11.007}.
\bibitem[{Stehman(2014)}]{Stehman2014a}
\bibinfo{author}{Stehman, S.~V.} (\bibinfo{year}{2014}).
\newblock \bibinfo{title}{{Estimating area and map accuracy for stratified
  random sampling when the strata are different from the map classes}}.
\newblock {\it \bibinfo{journal}{International Journal of Remote Sensing}\/},
  {\it \bibinfo{volume}{35}\/}, \bibinfo{pages}{4923--4939}. \URLprefix
  \url{http://dx.doi.org/10.1080/01431161.2014.930207}.
  \DOIprefix\doi{10.1080/01431161.2014.930207}.
\bibitem[{Tamm et~al.(2016)Tamm, Zalite, Voormansik \&
  Talgre}]{tamm2016relating}
\bibinfo{author}{Tamm, T.}, \bibinfo{author}{Zalite, K.},
  \bibinfo{author}{Voormansik, K.}, \& \bibinfo{author}{Talgre, L.}
  (\bibinfo{year}{2016}).
\newblock \bibinfo{title}{Relating sentinel-1 interferometric coherence to
  mowing events on grasslands}.
\newblock {\it \bibinfo{journal}{Remote Sensing}\/},  {\it
  \bibinfo{volume}{8}\/}, \bibinfo{pages}{802}.
\bibitem[{Teluguntla et~al.(2018)Teluguntla, Thenkabail, Oliphant, Xiong,
  Gumma, Congalton, Yadav \& Huete}]{teluguntla201830}
\bibinfo{author}{Teluguntla, P.}, \bibinfo{author}{Thenkabail, P.},
  \bibinfo{author}{Oliphant, A.}, \bibinfo{author}{Xiong, J.},
  \bibinfo{author}{Gumma, M.~K.}, \bibinfo{author}{Congalton, R.~G.},
  \bibinfo{author}{Yadav, K.}, \& \bibinfo{author}{Huete, A.}
  (\bibinfo{year}{2018}).
\newblock \bibinfo{title}{A 30-m landsat-derived cropland extent product of
  australia and china using random forest machine learning algorithm on google
  earth engine cloud computing platform}.
\newblock {\it \bibinfo{journal}{ISPRS journal of photogrammetry and remote
  sensing}\/},  {\it \bibinfo{volume}{144}\/}, \bibinfo{pages}{325--340}.
\bibitem[{Tian et~al.(2021)Tian, Cai, Jin, Hufkens, Scheifinger, Tagesson,
  Smets, Van~Hoolst, Bonte, Ivits et~al.}]{tian2021calibrating}
\bibinfo{author}{Tian, F.}, \bibinfo{author}{Cai, Z.}, \bibinfo{author}{Jin,
  H.}, \bibinfo{author}{Hufkens, K.}, \bibinfo{author}{Scheifinger, H.},
  \bibinfo{author}{Tagesson, T.}, \bibinfo{author}{Smets, B.},
  \bibinfo{author}{Van~Hoolst, R.}, \bibinfo{author}{Bonte, K.},
  \bibinfo{author}{Ivits, E.} et~al. (\bibinfo{year}{2021}).
\newblock \bibinfo{title}{Calibrating vegetation phenology from sentinel-2
  using eddy covariance, phenocam, and pep725 networks across europe}.
\newblock {\it \bibinfo{journal}{Remote Sensing of Environment}\/},  {\it
  \bibinfo{volume}{260}\/}, \bibinfo{pages}{112456}.
\bibitem[{Toreti et~al.(2019)Toreti, Belward, Perez-Dominguez, Naumann,
  Luterbacher, Cronie, Seguini, Manfron, Lopez~Lozano, Baruth
  et~al.}]{toreti2019exceptional}
\bibinfo{author}{Toreti, A.}, \bibinfo{author}{Belward, A.},
  \bibinfo{author}{Perez-Dominguez, I.}, \bibinfo{author}{Naumann, G.},
  \bibinfo{author}{Luterbacher, J.}, \bibinfo{author}{Cronie, O.},
  \bibinfo{author}{Seguini, L.}, \bibinfo{author}{Manfron, G.},
  \bibinfo{author}{Lopez~Lozano, R.}, \bibinfo{author}{Baruth, B.} et~al.
  (\bibinfo{year}{2019}).
\newblock \bibinfo{title}{The exceptional 2018 european water seesaw calls for
  action on adaptation}.
\newblock {\it \bibinfo{journal}{Earth's Future}\/}, .
\bibitem[{Toth \& Milenov(2020)}]{tgiacssdiscovery}
\bibinfo{author}{Toth, K.}, \& \bibinfo{author}{Milenov, P.}
  (\bibinfo{year}{2020}).
\newblock \bibinfo{title}{Technical guidelines on iacs spatial data sharing.
  part 1, data discovery - publications office of the eu}.
\newblock {\it \bibinfo{journal}{EUR 30330 EN}\/}, .
\bibitem[{Ulaby et~al.(1981)Ulaby, Moore \& Fung}]{ulaby1981microwave}
\bibinfo{author}{Ulaby, F.~T.}, \bibinfo{author}{Moore, R.~K.}, \&
  \bibinfo{author}{Fung, A.~K.} (\bibinfo{year}{1981}).
\newblock {\it \bibinfo{title}{Microwave remote sensing: Active and passive.
  volume 1-microwave remote sensing fundamentals and radiometry}\/}.
\newblock \bibinfo{publisher}{{ }}.
\bibitem[{Van~Horn et~al.(2018)Van~Horn, Mac~Aodha, Song, Cui, Sun, Shepard,
  Adam, Perona \& Belongie}]{van2018inaturalist}
\bibinfo{author}{Van~Horn, G.}, \bibinfo{author}{Mac~Aodha, O.},
  \bibinfo{author}{Song, Y.}, \bibinfo{author}{Cui, Y.}, \bibinfo{author}{Sun,
  C.}, \bibinfo{author}{Shepard, A.}, \bibinfo{author}{Adam, H.},
  \bibinfo{author}{Perona, P.}, \& \bibinfo{author}{Belongie, S.}
  (\bibinfo{year}{2018}).
\newblock \bibinfo{title}{The inaturalist species classification and detection
  dataset}.
\newblock In {\it \bibinfo{booktitle}{Proceedings of the IEEE conference on
  computer vision and pattern recognition}\/} (pp.
  \bibinfo{pages}{8769--8778}).
\bibitem[{Van~Tricht et~al.(2018)Van~Tricht, Gobin, Gilliams \&
  Piccard}]{van2018synergistic}
\bibinfo{author}{Van~Tricht, K.}, \bibinfo{author}{Gobin, A.},
  \bibinfo{author}{Gilliams, S.}, \& \bibinfo{author}{Piccard, I.}
  (\bibinfo{year}{2018}).
\newblock \bibinfo{title}{Synergistic use of radar sentinel-1 and optical
  sentinel-2 imagery for crop mapping: a case study for belgium}.
\newblock {\it \bibinfo{journal}{Remote Sensing}\/},  {\it
  \bibinfo{volume}{10}\/}, \bibinfo{pages}{1642}.
\bibitem[{van~der Velde et~al.(2019)van~der Velde, van Diepen \&
  Baruth}]{van2019european}
\bibinfo{author}{van~der Velde, M.}, \bibinfo{author}{van Diepen, C.}, \&
  \bibinfo{author}{Baruth, B.} (\bibinfo{year}{2019}).
\newblock \bibinfo{title}{The european crop monitoring and yield forecasting
  system: Celebrating 25 years of jrc mars bulletins}.
\newblock {\it \bibinfo{journal}{Agricultural Systems}\/},  {\it
  \bibinfo{volume}{168}\/}, \bibinfo{pages}{56--57}.
\bibitem[{Veloso et~al.(2017)Veloso, Mermoz, Bouvet, Le~Toan, Planells, Dejoux
  \& Ceschia}]{veloso2017understanding}
\bibinfo{author}{Veloso, A.}, \bibinfo{author}{Mermoz, S.},
  \bibinfo{author}{Bouvet, A.}, \bibinfo{author}{Le~Toan, T.},
  \bibinfo{author}{Planells, M.}, \bibinfo{author}{Dejoux, J.-F.}, \&
  \bibinfo{author}{Ceschia, E.} (\bibinfo{year}{2017}).
\newblock \bibinfo{title}{Understanding the temporal behavior of crops using
  sentinel-1 and sentinel-2-like data for agricultural applications}.
\newblock {\it \bibinfo{journal}{Remote sensing of environment}\/},  {\it
  \bibinfo{volume}{199}\/}, \bibinfo{pages}{415--426}.
\bibitem[{Wang et~al.(2019)Wang, Azzari \& Lobell}]{Wang2019}
\bibinfo{author}{Wang, S.}, \bibinfo{author}{Azzari, G.}, \&
  \bibinfo{author}{Lobell, D.~B.} (\bibinfo{year}{2019}).
\newblock \bibinfo{title}{{Crop type mapping without field-level labels: Random
  forest transfer and unsupervised clustering techniques}}.
\newblock {\it \bibinfo{journal}{Remote Sensing of Environment}\/},  {\it
  \bibinfo{volume}{222}\/}, \bibinfo{pages}{303--317}.
  \DOIprefix\doi{10.1016/j.rse.2018.12.026}.
\bibitem[{Wei et~al.(2019)Wei, Zhang, Wang, Wang \& Xu}]{wei2019multi}
\bibinfo{author}{Wei, S.}, \bibinfo{author}{Zhang, H.}, \bibinfo{author}{Wang,
  C.}, \bibinfo{author}{Wang, Y.}, \& \bibinfo{author}{Xu, L.}
  (\bibinfo{year}{2019}).
\newblock \bibinfo{title}{Multi-temporal sar data large-scale crop mapping
  based on u-net model}.
\newblock {\it \bibinfo{journal}{Remote Sensing}\/},  {\it
  \bibinfo{volume}{11}\/}, \bibinfo{pages}{68}.
\bibitem[{Weigand et~al.(2020)Weigand, Staab, Wurm \&
  Taubenb{\"o}ck}]{weigand2020spatial}
\bibinfo{author}{Weigand, M.}, \bibinfo{author}{Staab, J.},
  \bibinfo{author}{Wurm, M.}, \& \bibinfo{author}{Taubenb{\"o}ck, H.}
  (\bibinfo{year}{2020}).
\newblock \bibinfo{title}{Spatial and semantic effects of lucas samples on
  fully automated land use/land cover classification in high-resolution
  sentinel-2 data}.
\newblock {\it \bibinfo{journal}{International Journal of Applied Earth
  Observation and Geoinformation}\/},  {\it \bibinfo{volume}{88}\/},
  \bibinfo{pages}{102065}.
\bibitem[{You \& Dong(2020)}]{you2020examining}
\bibinfo{author}{You, N.}, \& \bibinfo{author}{Dong, J.}
  (\bibinfo{year}{2020}).
\newblock \bibinfo{title}{Examining earliest identifiable timing of crops using
  all available sentinel 1/2 imagery and google earth engine}.
\newblock {\it \bibinfo{journal}{ISPRS Journal of Photogrammetry and Remote
  Sensing}\/},  {\it \bibinfo{volume}{161}\/}, \bibinfo{pages}{109--123}.

\end{thebibliography}

\appendix

\section*{Data dissemination}
\label{sec:app_data_dissemination}

The EU crop map masked with the non-vegetation classes and steep slopes is available for download and visualization.\footnote{
The map can be downloaded as a GeoTiff (6.2 Gb) from the FTP server : \url{https://jeodpp.jrc.ec.europa.eu/ftp/jrc-opendata/EUCROPMAP/2018/}.
The map can be accessed via a WMS service via:
\url{https://jeodpp.jrc.ec.europa.eu/jeodpp/services/ows/wms/landcover/eucropmap}.}


\onecolumn

\clearpage
\tableofcontents
\setlength{\cftfignumwidth}{1.2cm}
\setlength{\cfttabnumwidth}{1.5cm}
\listoffigures
\listoftables

\clearpage

\setcounter{figure}{0}
\setcounter{table}{0}
\setcounter{page}{1}

\section*{Supplementary Material}

\label{AppendixA}
\captionsetup{list=no}
\renewcommand{\thetable}{Supplementary Table S\arabic{table}}
\renewcommand{\thefigure}{Supplementary Fig. S\arabic{figure}}

\begin{table*}[ht]
\centering
\caption{The training data consist of 58,178 polygons derived from the LUCAS Copernicus module corresponding to 2,956,889 S1 10-m pixels and associated time series. The training data are distributed into 21 thematic classes and two geographical strata (Str1 and Str2).}
\label{tab:LUCAScopernicus_per_stratum}
\footnotesize
\begin{tabular}
{>{\bfseries}C{0.8cm} 
L{5cm} |
S[table-column-width=1cm]
S[table-column-width=1cm]
S[table-column-width=1cm] |
S[table-column-width=1.5cm] 
S[table-column-width=1.5cm]
S[table-column-width=1.5cm]
}

  \hline
  &  & \multicolumn{3}{c}{\textbf{\# polygons}} & \multicolumn{3}{c}{\textbf{\# pixels}} \\
  {\textbf{class}} & 
  {\textbf{label}} & 
  {\textbf{Str1}} &
  {\textbf{Str2}} &
  {\textbf{Sum}} &
  {\textbf{Str1 }} &
  {\textbf{Str2  }} & 
  {\textbf{Sum}} \\ 
  
  \hline
  100 & Artificial land & 157 & 56 & 213 & 2449 & 772 & 3221 \\ 
  211 & Common wheat & 4210 & 619 & 4829 & 261290 & 28826 & 290116 \\ 
  212 & Durum wheat & 198 & 388 & 586 & 10863 & 17962 & 28825 \\ 
  213 & Barley & 1615 & 930 & 2545 & 98670 & 44216 & 142886 \\ 
  214 & Rye & 528 & 75 & 603 & 31956 & 3259 & 35215 \\ 
  215 & Oats & 433 & 209 & 642 & 25021 & 8984 & 34005 \\ 
  216 & Maize & 2241 & 140 & 2381 & 119697 & 5947 & 125644 \\ 
  217 & Rice & 11 & 3 & 14 & 661 & 207 & 868 \\ 
  218 & Triticale & 297 & 32 & 329 & 17686 & 1588 & 19274 \\ 
  219 & Other cereals & 86 & 4 & 90 & 4625 & 99 & 4724 \\ 
  221 & Potatoes & 285 & 23 & 308 & 14965 & 850 & 15815 \\ 
  222 & Sugar beet & 396 & 7 & 403 & 22706 & 468 & 23174 \\ 
  223 & Other root crops & 75 & 19 & 94 & 4036 & 884 & 4920 \\ 
  230 & Other non perm. ind. crops & 141 & 106 & 247 & 7600 & 4400 & 12000 \\ 
  231 & Sunflower & 462 & 222 & 684 & 24369 & 10208 & 34577 \\ 
  232 & Rape and turnip rape & 1096 & 15 & 1111 & 63899 & 553 & 64452 \\ 
  233 & Soya & 154 & 1 & 155 & 7020 & 74 & 7094 \\ 
  240 & Dry pulse  vegetables and flowers & 423 & 285 & 708 & 22440 & 11627 & 34067 \\ 
  250 & Other fodder crops & 859 & 472 & 1331 & 40453 & 18600 & 59053 \\ 
  290 & Bare arable land & 628 & 683 & 1311 & 35167 & 28442 & 63609 \\ 
  300 & Woodland & 15507 & 6845 & 22352 & 932467 & 284063 & 1216530 \\ 
  500 & Grassland & 14021 & 3086 & 17107 & 624627 & 108337 & 732964 \\ 
  600 & Bare land & 75 & 60 & 135 & 2488 & 1368 & 3856 \\ 
   \hline
  \multicolumn{2}{l}{\textbf{Total}}&  43898 & 14280 & 58178 & 2375155 & 581734 & 2956889  \\  
   \hline
\end{tabular}
\end{table*}

\begin{table*}[ht]
\small
\addtolength{\tabcolsep}{-1pt}
\caption{Weighted area accuracy for all the classes}
\label{tab:LUCAS Confusion matrix}
\resizebox{\textwidth}{!}{
\begin{tabular}{llllllllllllllllllllllll}
\hline

 &  & 211 & 212 & 213 & 214 & 215 & 216 & 217 & 218 & 219 & 221 & 222 & 223 & 230 & 231 & 232 & 233 & 240 & 250 & 290 & 300 & 500 & U.A \\
 \hline

common wheat & 211 & 0.039 & 0.003 & 0.007 & 0.002 & 0.003 & 0.001 & 0 & 0.002 & 0 & 0 & 0 & 0 & 0 & 0 & 0.001 & 0 & 0 & 0.002 & 0.005 & 0.001 & 0.012 & 49.6 \\
durum wheat & 212 & 0 & 0.002 & 0 & 0 & 0 & 0 & 0 & 0 & 0 & 0 & 0 & 0 & 0 & 0 & 0 & 0 & 0 & 0 & 0 & 0 & 0.001 & 49.7 \\
barley & 213 & 0.003 & 0.001 & 0.015 & 0.001 & 0.001 & 0 & 0 & 0 & 0 & 0 & 0 & 0 & 0 & 0 & 0 & 0 & 0.001 & 0.001 & 0.002 & 0 & 0.003 & 51.6 \\
rye & 214 & 0 & 0 & 0 & 0.002 & 0 & 0 & 0 & 0.001 & 0 & 0 & 0 & 0 & 0 & 0 & 0 & 0 & 0 & 0 & 0 & 0 & 0 & 48.3 \\
oats & 215 & 0 & 0 & 0 & 0 & 0 & 0 & 0 & 0 & 0 & 0 & 0 & 0 & 0 & 0 & 0 & 0 & 0 & 0 & 0 & 0 & 0 & 37.3 \\
maize & 216 & 0.001 & 0 & 0.001 & 0 & 0 & 0.033 & 0 & 0 & 0.001 & 0.001 & 0.001 & 0 & 0 & 0 & 0 & 0.001 & 0.001 & 0.002 & 0.001 & 0.001 & 0.009 & 58.5 \\
rice & 217 & 0 & 0 & 0 & 0 & 0 & 0 & 0 & 0 & 0 & 0 & 0 & 0 & 0 & 0 & 0 & 0 & 0 & 0 & 0 & 0 & 0 &  \\
triticale & 218 & 0 & 0 & 0 & 0 & 0 & 0 & 0 & 0 & 0 & 0 & 0 & 0 & 0 & 0 & 0 & 0 & 0 & 0 & 0 & 0 & 0 & 50.0 \\
other cereals & 219 & 0 & 0 & 0 & 0 & 0 & 0 & 0 & 0 & 0 & 0 & 0 & 0 & 0 & 0 & 0 & 0 & 0 & 0 & 0 & 0 & 0 & 57.1 \\
potatoes & 221 & 0 & 0 & 0 & 0 & 0 & 0 & 0 & 0 & 0 & 0.002 & 0 & 0 & 0 & 0 & 0 & 0 & 0 & 0 & 0 & 0 & 0 & 73.5 \\
sugar beet & 222 & 0 & 0 & 0 & 0 & 0 & 0 & 0 & 0 & 0 & 0 & 0.003 & 0 & 0 & 0 & 0 & 0 & 0 & 0 & 0 & 0 & 0 & 75.0 \\
other roots crops & 223 & 0 & 0 & 0 & 0 & 0 & 0 & 0 & 0 & 0 & 0 & 0 & 0 & 0 & 0 & 0 & 0 & 0 & 0 & 0 & 0 & 0 &  \\
\begin{tabular}[c]{@{}l@{}}other non permanent \\ industrial crops\end{tabular} & 230 & 0 & 0 & 0 & 0 & 0 & 0 & 0 & 0 & 0 & 0 & 0 & 0 & 0.001 & 0 & 0 & 0 & 0 & 0 & 0 & 0 & 0 & 76.5 \\
sunflower & 231 & 0 & 0 & 0 & 0 & 0 & 0.001 & 0 & 0 & 0 & 0 & 0 & 0 & 0 & 0.007 & 0 & 0 & 0 & 0 & 0 & 0 & 0.001 & 62.2 \\
rape and turnip rape & 232 & 0 & 0 & 0 & 0 & 0 & 0 & 0 & 0 & 0 & 0 & 0 & 0 & 0 & 0 & 0.014 & 0 & 0 & 0 & 0.001 & 0 & 0.001 & 80.2 \\
soya & 233 & 0 & 0 & 0 & 0 & 0 & 0 & 0 & 0 & 0 & 0 & 0 & 0 & 0 & 0 & 0 & 0 & 0 & 0 & 0 & 0 & 0 & 88.9 \\
\begin{tabular}[c]{@{}l@{}}dry pulses, \\ vegetables and flowers\end{tabular} & 240 & 0 & 0 & 0 & 0 & 0 & 0 & 0 & 0 & 0 & 0 & 0 & 0 & 0 & 0 & 0 & 0 & 0.002 & 0 & 0 & 0 & 0.001 & 40.35cm \\
\begin{tabular}[c]{@{}l@{}}fodder crops \\ (cereals and leguminous)\end{tabular} & 250 & 0 & 0 & 0 & 0 & 0 & 0 & 0 & 0 & 0 & 0 & 0 & 0 & 0 & 0 & 0 & 0 & 0 & 0.001 & 0 & 0 & 0.001 & 28.4 \\
bare arable land & 290 & 0 & 0 & 0 & 0 & 0 & 0 & 0 & 0 & 0 & 0 & 0 & 0 & 0 & 0 & 0 & 0 & 0 & 0 & 0.005 & 0.001 & 0.002 & 49.1 \\
\begin{tabular}[c]{@{}l@{}}woodland and shrubland \\ vegetation (incl.   permanent crops)\end{tabular} & 300 & 0.002 & 0.001 & 0.002 & 0 & 0.001 & 0.003 & 0 & 0 & 0 & 0 & 0 & 0 & 0 & 0 & 0.001 & 0 & 0.001 & 0.004 & 0.004 & 0.49 & 0.078 & 83.3 \\
\begin{tabular}[c]{@{}l@{}}permanent and temporary \\ grasslands\end{tabular} & 500 & 0.004 & 0.001 & 0.002 & 0.001 & 0.001 & 0.001 & 0 & 0 & 0 & 0 & 0 & 0 & 0 & 0 & 0 & 0 & 0 & 0.009 & 0.001 & 0.013 & 0.151 & 82.0 \\
 &  &  &  &  &  &  &  &  &  &  &  &  &  &  &  &  &  &  &  &  &  &  &  \\
 \hline

 &  & 0.0504 & 0.0087 & 0.0286 & 0.0067 & 0.0077 & 0.0389 & 0.0009 & 0.0043 & 0.0015 & 0.0039 & 0.0058 & 0.0010 & 0.0027 & 0.0076 & 0.0168 & 0.0018 & 0.0069 & 0.0195 & 0.0209 & 0.5058 & 0.2595 &  \\
P.A. &  & 78.1 & 22.1 & 52.7 & 31.1 & 6.0 & 83.7 & 0.0 & 1.0 & 1.6 & 38.7 & 58.3 & 0.0 & 24.6 & 85.9 & 82.1 & 7.2 & 24.7 & 6.2 & 22.0 & 96.9 & 58.1 & \\
\\
O.A. & 0.766 &  &  &  &  &  &  &  &  &  & \\
\hline
\end{tabular}}
\end{table*}

\begin{table*}[ht]
\centering
\footnotesize
\caption{Hyperparameters of the RF models used in the final version }
\label{tab:appendix_Hyperparameters}
\begin{tabular}{@{}l|l|l@{}}
\hline
\textbf{} & \textbf{Stratum 1}                                                                                                                                         & \textbf{Stratum 2} \\
\hline
Level1    & \begin{tabular}[c]{@{}l@{}}n\_estimators=1200 \\ max\_features='sqrt' \\ min\_samples\_leaf=1,  \\ min\_samples\_split=2,\\ max\_depth=None, \\criterion='gini', \\bootstrap = True \end{tabular} &\begin{tabular}[c]{@{}l@{}}n\_estimators=900 \\ max\_features='sqrt' \\ min\_samples\_leaf=1,  \\ min\_samples\_split=3\, \\max\_depth=None, \\criterion='gini', \\bootstrap = True \end{tabular}                      \\
\hline
Level2    & \begin{tabular}[c]{@{}l@{}}n\_estimators=1200 \\ max\_features='log2' \\ min\_samples\_leaf=1 \\ min\_samples\_split=2,\\ max\_depth=None, \\criterion='gini', \\bootstrap = True \end{tabular}   &   \begin{tabular}[c]{@{}l@{}}n\_estimators=900 \\ max\_features='log2' \\ min\_samples\_leaf=1,  \\ min\_samples\_split=2,\\ max\_depth=None, \\criterion='gini', \\bootstrap = True \end{tabular}                  \\
\hline
\end{tabular}
\end{table*}

\begin{figure*}[ht]
  \centering
  \begin{subfigure}{.47\linewidth}
    \vspace{0.1cm}
    \centering
    \includegraphics[width = \linewidth]{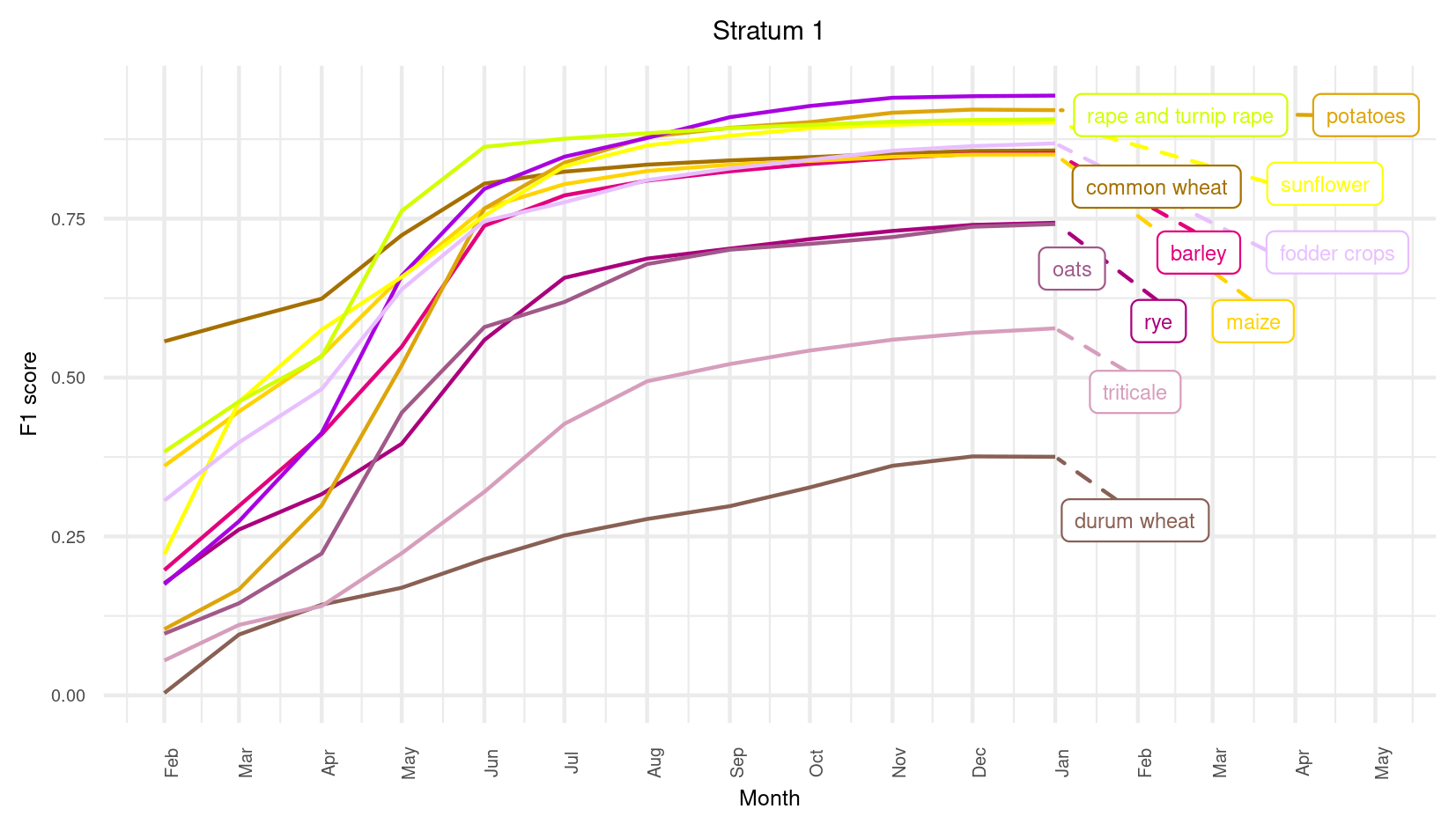}
    \caption{}
  \end{subfigure}
  \hspace{1em}
  \begin{subfigure}{.47\linewidth}
    \centering
    \includegraphics[width = \linewidth]{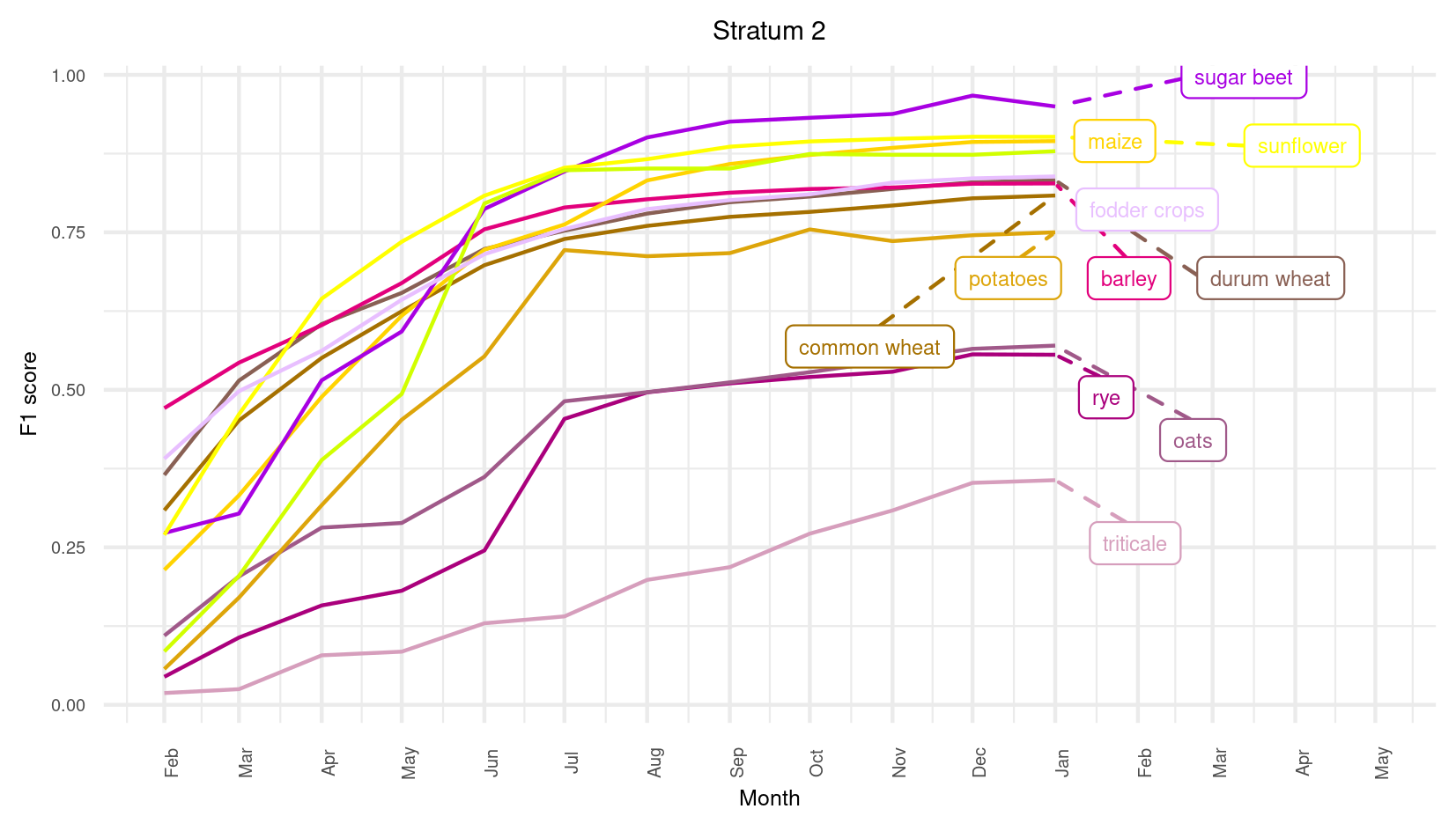}
    \caption{}
  \end{subfigure}
  \caption{Evolution of The F-score for each Stratum through time for the 12 mains crops over the EU-28. }
  \label{fig:benchmark_date_combined_Fscore_crop_strata}
\end{figure*}

\begin{table*}[ht]
\centering
\footnotesize
\caption{Masking the EU crop map for visualization and area estimation }
\label{tab:Appendix_Masking}
\begin{tabular}{l|l|l}
\hline
 & Classes labels & Class codes \\ \hline
Corine Land Cover 2018 &  &  \\
 & Inland and coastal wetlands & 411-412, 421-423 \\
 & Moors and Heatlands & 322 \\
 & Bare Lands & 331-333,335 \\
 & Port and airports & 123,124 \\
JRC GHSL European Settlement Map R2019 & Residential and non residential built-up area & 250, 255 \\
JRC Surface Water 2019 & 2019 permanent water bodies & 3 \\ \hline
\end{tabular}
\end{table*}

\begin{table*}[ht]
\centering
\footnotesize
\caption{Area (ha) of the EU crop type classes, with the projection EPSG 3035 }
\label{tab:Appendix_Area}
\begin{tabular}{@{}r S[table-column-width=4cm]@{}}
\toprule
\textbf{Class}                                 & \textbf{Area (ha)}  \\ \midrule
211                                            & 31325983.42         \\
212                                            & 1842598.45          \\
213                                            & 11822095.11         \\
214                                            & 1641600.28          \\
215                                            & 457025.1            \\
216                                            & 22188829.6          \\
217                                            & 3380.94             \\
218                                            & 33631.6             \\
219                                            & 13944.24            \\
221                                            & 785300.85           \\
222                                            & 1753491.13          \\
223                                            & 19124.88            \\
230                                            & 393616.4            \\
231                                            & 4551610.89          \\
232                                            & 6868240.33          \\
233                                            & 62272.83            \\
240                                            & 1731230.9           \\
250                                            & 1872904.25          \\
290                                            & 4375965.4           \\
\hline
\multicolumn{1}{l}{\textbf{Total arable land}} & \textbf{\num{91742847}} \\
\hline
300                                            & 223436295.7       \\
500                                            & 70941202.43       \\ \bottomrule
\end{tabular}
\end{table*}

\begin{table*}[ht]
\centering
\caption{Confusion matrix for the EU-28 region. The values represent sample counts. UA = User’s accuracy, PA=Producer Accuracy }
\label{tab:Appendix_ConfusionMatrixCount}
\resizebox{\textwidth}{!}{%
\begin{tabular}{@{}crrrrrrrrrrrrrrrrrrrrrrrr@{}}
 & \multicolumn{24}{c}{\textbf{Reference class (LUCAS point)}} \\
 \hline
 & \multicolumn{1}{c}{211} & \multicolumn{1}{c}{212} & \multicolumn{1}{c}{213} & \multicolumn{1}{c}{214} & \multicolumn{1}{c}{215} & \multicolumn{1}{c}{216} & \multicolumn{1}{c}{217} & \multicolumn{1}{c}{218} & \multicolumn{1}{c}{219} & \multicolumn{1}{c}{221} & \multicolumn{1}{c}{222} & \multicolumn{1}{c}{223} & \multicolumn{1}{c}{230} & \multicolumn{1}{c}{231} & \multicolumn{1}{c}{232} & \multicolumn{1}{c}{233} & \multicolumn{1}{c}{240} & \multicolumn{1}{c}{250} & \multicolumn{1}{c}{290} & \multicolumn{1}{c}{300} & \multicolumn{1}{c}{500} & \multicolumn{1}{c}{total} & \multicolumn{1}{c}{\textbf{UA}} & \multicolumn{1}{c}{\textbf{OA}} \\
 \hline

211 & 4624 & 367 & 833 & 287 & 348 & 99 & 7 & 269 & 25 & 8 & 11 & 12 & 52 & 4 & 93 & 3 & 47 & 205 & 598 & 69 & 1369 & 9330 & \textbf{49.6} & \textbf{74.0} \\
212 & 68 & 402 & 61 & 2 & 37 & 3 & 1 & 3 & 0 & 0 & 0 & 1 & 2 & 0 & 0 & 0 & 10 & 60 & 43 & 5 & 111 & 809 & \textbf{49.7} & \textbf{} \\
213 & 372 & 121 & 2117 & 112 & 206 & 47 & 6 & 38 & 4 & 3 & 0 & 10 & 24 & 3 & 23 & 0 & 78 & 164 & 255 & 49 & 467 & 4099 & \textbf{51.6} & \textbf{} \\
214 & 34 & 3 & 41 & 236 & 11 & 2 & 0 & 99 & 1 & 1 & 0 & 0 & 2 & 0 & 12 & 0 & 1 & 1 & 19 & 0 & 26 & 489 & \textbf{48.3} & \textbf{} \\
215 & 26 & 3 & 14 & 3 & 56 & 0 & 0 & 1 & 0 & 0 & 0 & 0 & 1 & 0 & 0 & 0 & 1 & 15 & 8 & 1 & 21 & 150 & \textbf{37.3} & \textbf{} \\
216 & 128 & 16 & 105 & 21 & 44 & 3464 & 28 & 16 & 81 & 124 & 149 & 38 & 36 & 45 & 36 & 120 & 133 & 169 & 139 & 98 & 936 & 5926 & \textbf{58.5} & \textbf{} \\
217 & 0 & 0 & 0 & 0 & 0 & 0 & 0 & 0 & 0 & 0 & 0 & 0 & 0 & 0 & 0 & 0 & 0 & 0 & 0 & 0 & 0 & 0 & \textbf{} & \textbf{} \\
218 & 2 & 0 & 1 & 1 & 0 & 0 & 0 & 4 & 0 & 0 & 0 & 0 & 0 & 0 & 0 & 0 & 0 & 0 & 0 & 0 & 0 & 8 & \textbf{50.0} & \textbf{} \\
219 & 0 & 0 & 0 & 0 & 0 & 0 & 0 & 0 & 4 & 1 & 0 & 0 & 0 & 0 & 1 & 0 & 0 & 0 & 0 & 0 & 1 & 7 & \textbf{57.1} & \textbf{} \\
221 & 0 & 0 & 0 & 0 & 0 & 1 & 0 & 0 & 0 & 180 & 35 & 4 & 0 & 5 & 2 & 2 & 6 & 0 & 5 & 0 & 5 & 245 & \textbf{73.5} & \textbf{} \\
222 & 0 & 0 & 2 & 0 & 0 & 21 & 0 & 0 & 1 & 68 & 476 & 7 & 0 & 17 & 3 & 11 & 14 & 1 & 7 & 2 & 5 & 635 & \textbf{75.0} & \textbf{} \\
223 & 0 & 0 & 0 & 0 & 0 & 0 & 0 & 0 & 0 & 0 & 0 & 1 & 0 & 0 & 0 & 0 & 2 & 0 & 0 & 0 & 0 & 3 & \textbf{33.3} & \textbf{} \\
230 & 0 & 0 & 1 & 0 & 0 & 0 & 5 & 0 & 1 & 1 & 0 & 0 & 124 & 5 & 0 & 0 & 10 & 1 & 7 & 0 & 7 & 162 & \textbf{76.5} & \textbf{} \\
231 & 5 & 3 & 9 & 0 & 1 & 46 & 4 & 1 & 1 & 17 & 28 & 7 & 36 & 530 & 3 & 22 & 40 & 8 & 37 & 7 & 47 & 852 & \textbf{62.2} & \textbf{} \\
232 & 15 & 1 & 12 & 3 & 1 & 4 & 0 & 5 & 3 & 1 & 10 & 4 & 5 & 2 & 1415 & 1 & 34 & 11 & 135 & 8 & 94 & 1764 & \textbf{80.2} & \textbf{} \\
233 & 0 & 0 & 0 & 0 & 0 & 1 & 0 & 0 & 0 & 0 & 0 & 0 & 0 & 0 & 0 & 8 & 0 & 0 & 0 & 0 & 0 & 9 & \textbf{88.9} & \textbf{} \\
240 & 5 & 13 & 13 & 2 & 1 & 9 & 1 & 1 & 6 & 21 & 4 & 16 & 8 & 3 & 37 & 2 & 213 & 26 & 51 & 7 & 89 & 528 & \textbf{40.3} & \textbf{} \\
250 & 36 & 44 & 45 & 4 & 31 & 21 & 4 & 3 & 2 & 0 & 0 & 2 & 2 & 1 & 8 & 0 & 32 & 203 & 29 & 16 & 232 & 715 & \textbf{28.4} & \textbf{} \\
290 & 21 & 15 & 24 & 0 & 13 & 34 & 34 & 1 & 0 & 11 & 2 & 4 & 11 & 20 & 12 & 0 & 35 & 28 & 667 & 91 & 335 & 1358 & \textbf{49.1} & \textbf{} \\
300 & 137 & 93 & 117 & 26 & 47 & 179 & 8 & 18 & 11 & 10 & 9 & 2 & 14 & 14 & 62 & 9 & 70 & 275 & 251 & 31981 & 5082 & 38415 & \textbf{83.3} & \textbf{} \\
500 & 441 & 79 & 253 & 84 & 131 & 144 & 2 & 38 & 19 & 1 & 1 & 5 & 22 & 1 & 18 & 3 & 46 & 1039 & 163 & 1532 & 18327 & 22349 & \textbf{82.0} & \textbf{} \\
\hline
Total & \multicolumn{1}{c}{5914} & \multicolumn{1}{c}{1160} & \multicolumn{1}{c}{3648} & \multicolumn{1}{c}{781} & \multicolumn{1}{c}{927} & \multicolumn{1}{c}{4075} & \multicolumn{1}{c}{100} & \multicolumn{1}{c}{497} & \multicolumn{1}{c}{159} & \multicolumn{1}{c}{447} & \multicolumn{1}{c}{725} & \multicolumn{1}{c}{113} & \multicolumn{1}{c}{339} & \multicolumn{1}{c}{650} & \multicolumn{1}{c}{1725} & \multicolumn{1}{c}{181} & \multicolumn{1}{c}{772} & \multicolumn{1}{c}{2206} & \multicolumn{1}{c}{2414} & \multicolumn{1}{c}{33866} & \multicolumn{1}{c}{27154} &  &  &  \\
\hline

\multicolumn{1}{l}{\begin{tabular}[c]{@{}l@{}}Producer 's \\ accuracy\end{tabular}} & \multicolumn{1}{c}{\textbf{78.2}} & \multicolumn{1}{c}{\textbf{34.7}} & \multicolumn{1}{c}{\textbf{58.0}} & \multicolumn{1}{c}{\textbf{30.2}} & \multicolumn{1}{c}{\textbf{6.0}} & \multicolumn{1}{c}{\textbf{85.0}} & \multicolumn{1}{c}{\textbf{0.0}} & \multicolumn{1}{c}{\textbf{0.8}} & \multicolumn{1}{c}{\textbf{2.5}} & \multicolumn{1}{c}{\textbf{40.3}} & \multicolumn{1}{c}{\textbf{65.7}} & \multicolumn{1}{c}{\textbf{0.9}} & \multicolumn{1}{c}{\textbf{36.6}} & \multicolumn{1}{c}{\textbf{81.5}} & \multicolumn{1}{c}{\textbf{82.0}} & \multicolumn{1}{c}{\textbf{4.4}} & \multicolumn{1}{c}{\textbf{27.6}} & \multicolumn{1}{c}{\textbf{9.2}} & \multicolumn{1}{c}{\textbf{27.6}} & \multicolumn{1}{c}{\textbf{94.4}} & \multicolumn{1}{c}{\textbf{67.5}} & \multicolumn{1}{c}{\textbf{}} & \multicolumn{1}{c}{} & \multicolumn{1}{c}{} \\
\hline

\end{tabular}%
}
\end{table*}

\begin{table*}[ht]
\centering
\caption{Confusion matrix for the EU-28 region. The values represent the estimated area proportions. UA = User’s accuracy, PA=Producer Accuracy }
\label{tab:Appendix_ConfusionMatrixAreaProportion}
\resizebox{\textwidth}{!}{%
\begin{tabular}{@{}rrrrrrrrrrrrrrrrrrrrrrrrr@{}}
\hline

\multicolumn{1}{l}{} & \multicolumn{4}{l}{\textbf{Reference class (LUCAS point)}} & \multicolumn{1}{l}{\textbf{}} & \multicolumn{1}{l}{} & \multicolumn{1}{l}{} & \multicolumn{1}{l}{} & \multicolumn{1}{l}{} & \multicolumn{1}{l}{} & \multicolumn{1}{l}{} & \multicolumn{1}{l}{} & \multicolumn{1}{l}{} & \multicolumn{1}{l}{} & \multicolumn{1}{l}{} & \multicolumn{1}{l}{} & \multicolumn{1}{l}{} & \multicolumn{1}{l}{} & \multicolumn{1}{l}{} & \multicolumn{1}{l}{} & \multicolumn{1}{l}{} & \multicolumn{1}{l}{} & \multicolumn{1}{l}{} & \multicolumn{1}{l}{} \\
\multicolumn{1}{l}{\textbf{Map class}} & 211 & 212 & 213 & 214 & 215 & 216 & 217 & 218 & 219 & 221 & 222 & 223 & 230 & 231 & 232 & 233 & 240 & 250 & 290 & 300 & 500 & \multicolumn{1}{l}{UA} & \multicolumn{1}{l}{SE} & \multicolumn{1}{l}{OA} \\
\hline

211 & 0.04 & 0.003 & 0.007 & 0.002 & 0.003 & 0.001 & 0 & 0.002 & 0 & 0 & 0 & 0 & 0 & 0 & 0.001 & 0 & 0 & 0.002 & 0.005 & 0.001 & 0.012 & 49.6 & 1 & 76.3 \\
212 & 0 & 0.002 & 0 & 0 & 0 & 0 & 0 & 0 & 0 & 0 & 0 & 0 & 0 & 0 & 0 & 0 & 0 & 0 & 0 & 0 & 0.001 & 49.7 & 3.4 &  \\
213 & 0.003 & 0.001 & 0.016 & 0.001 & 0.002 & 0 & 0 & 0 & 0 & 0 & 0 & 0 & 0 & 0 & 0 & 0 & 0.001 & 0.001 & 0.002 & 0 & 0.003 & 51.6 & 1.5 &  \\
214 & 0 & 0 & 0 & 0.002 & 0 & 0 & 0 & 0.001 & 0 & 0 & 0 & 0 & 0 & 0 & 0 & 0 & 0 & 0 & 0 & 0 & 0 & 48.3 & 4.4 &  \\
215 & 0 & 0 & 0 & 0 & 0 & 0 & 0 & 0 & 0 & 0 & 0 & 0 & 0 & 0 & 0 & 0 & 0 & 0 & 0 & 0 & 0 & 37.3 & 7.8 &  \\
216 & 0.001 & 0 & 0.001 & 0 & 0 & 0.034 & 0 & 0 & 0.001 & 0.001 & 0.001 & 0 & 0 & 0 & 0 & 0.001 & 0.001 & 0.002 & 0.001 & 0.001 & 0.009 & 58.5 & 1.3 &  \\
217 & 0 & 0 & 0 & 0 & 0 & 0 & 0 & 0 & 0 & 0 & 0 & 0 & 0 & 0 & 0 & 0 & 0 & 0 & 0 & 0 & 0 & \multicolumn{1}{c}{} & \multicolumn{1}{c}{} & \multicolumn{1}{c}{} \\
218 & 0 & 0 & 0 & 0 & 0 & 0 & 0 & 0 & 0 & 0 & 0 & 0 & 0 & 0 & 0 & 0 & 0 & 0 & 0 & 0 & 0 & 50 & 37 &  \\
219 & 0 & 0 & 0 & 0 & 0 & 0 & 0 & 0 & 0 & 0 & 0 & 0 & 0 & 0 & 0 & 0 & 0 & 0 & 0 & 0 & 0 & 57.1 & 39.6 &  \\
221 & 0 & 0 & 0 & 0 & 0 & 0 & 0 & 0 & 0 & 0.001 & 0 & 0 & 0 & 0 & 0 & 0 & 0 & 0 & 0 & 0 & 0 & 73.5 & 5.5 &  \\
222 & 0 & 0 & 0 & 0 & 0 & 0 & 0 & 0 & 0 & 0 & 0.003 & 0 & 0 & 0 & 0 & 0 & 0 & 0 & 0 & 0 & 0 & 75 & 3.4 &  \\
223 & 0 & 0 & 0 & 0 & 0 & 0 & 0 & 0 & 0 & 0 & 0 & 0 & 0 & 0 & 0 & 0 & 0 & 0 & 0 & 0 & 0 & 33.3 & 65.3 &  \\
230 & 0 & 0 & 0 & 0 & 0 & 0 & 0 & 0 & 0 & 0 & 0 & 0 & 0.001 & 0 & 0 & 0 & 0 & 0 & 0 & 0 & 0 & 76.5 & 6.5 &  \\
231 & 0 & 0 & 0 & 0 & 0 & 0.001 & 0 & 0 & 0 & 0 & 0 & 0 & 0 & 0.007 & 0 & 0 & 0.001 & 0 & 0.001 & 0 & 0.001 & 62.2 & 3.3 &  \\
232 & 0 & 0 & 0 & 0 & 0 & 0 & 0 & 0 & 0 & 0 & 0 & 0 & 0 & 0 & 0.014 & 0 & 0 & 0 & 0.001 & 0 & 0.001 & 80.2 & 1.9 &  \\
233 & 0 & 0 & 0 & 0 & 0 & 0 & 0 & 0 & 0 & 0 & 0 & 0 & 0 & 0 & 0 & 0 & 0 & 0 & 0 & 0 & 0 & 88.9 & 21.8 &  \\
240 & 0 & 0 & 0 & 0 & 0 & 0 & 0 & 0 & 0 & 0 & 0 & 0 & 0 & 0 & 0 & 0 & 0.002 & 0 & 0 & 0 & 0.001 & 40.3 & 4.2 &  \\
250 & 0 & 0 & 0 & 0 & 0 & 0 & 0 & 0 & 0 & 0 & 0 & 0 & 0 & 0 & 0 & 0 & 0 & 0.001 & 0 & 0 & 0.002 & 28.4 & 3.3 &  \\
290 & 0 & 0 & 0 & 0 & 0 & 0 & 0 & 0 & 0 & 0 & 0 & 0 & 0 & 0 & 0 & 0 & 0 & 0 & 0.006 & 0.001 & 0.003 & 49.1 & 2.7 &  \\
300 & 0.002 & 0.001 & 0.002 & 0 & 0.001 & 0.003 & 0 & 0 & 0 & 0 & 0 & 0 & 0 & 0 & 0.001 & 0 & 0.001 & 0.004 & 0.004 & 0.482 & 0.077 & 83.3 & 0.4 &  \\
500 & 0.004 & 0.001 & 0.002 & 0.001 & 0.001 & 0.001 & 0 & 0 & 0 & 0 & 0 & 0 & 0 & 0 & 0 & 0 & 0 & 0.009 & 0.001 & 0.013 & 0.151 & 82 & 0.5 &  \\
\hline

\multicolumn{1}{l}{\textbf{PA(\%)}} & \textbf{78} & \textbf{25.5} & \textbf{53.4} & \textbf{30.2} & \textbf{5.6} & \textbf{83.8} & \textbf{0} & \textbf{1} & \textbf{1.4} & \textbf{37.6} & \textbf{57.5} & \textbf{1.6} & \textbf{26.8} & \textbf{86.7} & \textbf{82.3} & \textbf{7.5} & \textbf{25} & \textbf{6.9} & \textbf{25} & \textbf{96.9} & \textbf{58} & \textbf{} & \multicolumn{1}{l}{\textbf{}} & \multicolumn{1}{l}{\textbf{}} \\
\hline
\multicolumn{1}{l}{\textbf{SE(\%)}} & \textbf{1} & \textbf{1.9} & \textbf{1.4} & \textbf{2.6} & \textbf{1.2} & \textbf{1.2} & \textbf{0} & \textbf{0.7} & \textbf{1} & \textbf{3.3} & \textbf{3.2} & \textbf{3.1} & \textbf{3.2} & \textbf{2.2} & \textbf{1.7} & \textbf{2} & \textbf{2.5} & \textbf{0.8} & \textbf{1.4} & \textbf{0.1} & \textbf{0.5} & \textbf{} & \multicolumn{1}{l}{\textbf{}} & \multicolumn{1}{l}{\textbf{}}\\
\hline

\end{tabular}%
}
\end{table*}

\begin{table*}[ht]
\centering
\caption{Confusion matrix for main crop type group for the EU-28 region. The values represent sample counts. UA = User’s accuracy, PA=Producer Accuracy }
\label{tab:Appendix_ConfusionMatrixCount_grouped}
\resizebox{\textwidth}{!}{%
\begin{tabular}{@{}lllrlrlrlrlrlrlrlrlrllll@{}}
\rowcolor[HTML]{F2F2F2} 
 & \multicolumn{2}{l}{\cellcolor[HTML]{F2F2F2}\textbf{}} & \multicolumn{21}{l}{\cellcolor[HTML]{F2F2F2}\textbf{Reference class (LUCAS point)}} \\
 & \multicolumn{2}{l}{\textbf{Map Class}} & \multicolumn{2}{l}{210} & \multicolumn{2}{l}{220} & \multicolumn{2}{l}{230} & \multicolumn{2}{l}{240} & \multicolumn{2}{l}{250} & \multicolumn{2}{l}{290} & \multicolumn{2}{l}{300} & \multicolumn{2}{l}{500} & \multicolumn{2}{l}{Total} & \multicolumn{2}{l}{UA (\%)} & OA(\%) \\
\rowcolor[HTML]{F2F2F2} 
\textbf{210} & \multicolumn{2}{l}{\cellcolor[HTML]{F2F2F2}Cereals} & \multicolumn{2}{r}{\cellcolor[HTML]{F2F2F2}14904} & \multicolumn{2}{r}{\cellcolor[HTML]{F2F2F2}358} & \multicolumn{2}{r}{\cellcolor[HTML]{F2F2F2}240} & \multicolumn{2}{r}{\cellcolor[HTML]{F2F2F2}270} & \multicolumn{2}{r}{\cellcolor[HTML]{F2F2F2}614} & \multicolumn{2}{r}{\cellcolor[HTML]{F2F2F2}1062} & \multicolumn{2}{r}{\cellcolor[HTML]{F2F2F2}222} & \multicolumn{2}{r}{\cellcolor[HTML]{F2F2F2}2931} & \multicolumn{2}{r}{\cellcolor[HTML]{F2F2F2}20601} & \multicolumn{2}{l}{\cellcolor[HTML]{F2F2F2}\textbf{72.3}} & \textbf{79.2} \\
\textbf{220} & \multicolumn{2}{l}{Root Crops} & \multicolumn{2}{r}{25} & \multicolumn{2}{r}{771} & \multicolumn{2}{r}{13} & \multicolumn{2}{r}{22} & \multicolumn{2}{r}{1} & \multicolumn{2}{r}{12} & \multicolumn{2}{r}{2} & \multicolumn{2}{r}{10} & \multicolumn{2}{r}{856} & \multicolumn{2}{l}{\textbf{90.1}} & \textbf{} \\
\rowcolor[HTML]{F2F2F2} 
\textbf{230} & \multicolumn{2}{l}{\cellcolor[HTML]{F2F2F2}Non permanent industrial crops} & \multicolumn{2}{r}{\cellcolor[HTML]{F2F2F2}8} & \multicolumn{2}{r}{\cellcolor[HTML]{F2F2F2}1} & \multicolumn{2}{r}{\cellcolor[HTML]{F2F2F2}132} & \multicolumn{2}{r}{\cellcolor[HTML]{F2F2F2}10} & \multicolumn{2}{r}{\cellcolor[HTML]{F2F2F2}1} & \multicolumn{2}{r}{\cellcolor[HTML]{F2F2F2}7} & \multicolumn{2}{r}{\cellcolor[HTML]{F2F2F2}0} & \multicolumn{2}{r}{\cellcolor[HTML]{F2F2F2}7} & \multicolumn{2}{r}{\cellcolor[HTML]{F2F2F2}166} & \multicolumn{2}{l}{\cellcolor[HTML]{F2F2F2}\textbf{79.5}} & \textbf{} \\
\textbf{240} & \multicolumn{2}{l}{Dry pulses, Vegetables and Flowers} & \multicolumn{2}{r}{51} & \multicolumn{2}{r}{41} & \multicolumn{2}{r}{10} & \multicolumn{2}{r}{213} & \multicolumn{2}{r}{26} & \multicolumn{2}{r}{51} & \multicolumn{2}{r}{7} & \multicolumn{2}{r}{89} & \multicolumn{2}{r}{488} & \multicolumn{2}{l}{\textbf{43.6}} & \textbf{} \\
\rowcolor[HTML]{F2F2F2} 
\textbf{250} & \multicolumn{2}{l}{\cellcolor[HTML]{F2F2F2}Fodder Crops} & \multicolumn{2}{r}{\cellcolor[HTML]{F2F2F2}190} & \multicolumn{2}{r}{\cellcolor[HTML]{F2F2F2}2} & \multicolumn{2}{r}{\cellcolor[HTML]{F2F2F2}2} & \multicolumn{2}{r}{\cellcolor[HTML]{F2F2F2}32} & \multicolumn{2}{r}{\cellcolor[HTML]{F2F2F2}203} & \multicolumn{2}{r}{\cellcolor[HTML]{F2F2F2}29} & \multicolumn{2}{r}{\cellcolor[HTML]{F2F2F2}16} & \multicolumn{2}{r}{\cellcolor[HTML]{F2F2F2}232} & \multicolumn{2}{r}{\cellcolor[HTML]{F2F2F2}706} & \multicolumn{2}{l}{\cellcolor[HTML]{F2F2F2}\textbf{28.8}} & \textbf{} \\
\textbf{290} & \multicolumn{2}{l}{Bare Arable Land} & \multicolumn{2}{r}{142} & \multicolumn{2}{r}{17} & \multicolumn{2}{r}{11} & \multicolumn{2}{r}{35} & \multicolumn{2}{r}{28} & \multicolumn{2}{r}{667} & \multicolumn{2}{r}{91} & \multicolumn{2}{r}{335} & \multicolumn{2}{r}{1326} & \multicolumn{2}{l}{\textbf{50.3}} & \textbf{} \\
\rowcolor[HTML]{F2F2F2} 
\textbf{300} & \multicolumn{2}{l}{\cellcolor[HTML]{F2F2F2}Tree and Shrub Cover} & \multicolumn{2}{r}{\cellcolor[HTML]{F2F2F2}636} & \multicolumn{2}{r}{\cellcolor[HTML]{F2F2F2}21} & \multicolumn{2}{r}{\cellcolor[HTML]{F2F2F2}23} & \multicolumn{2}{r}{\cellcolor[HTML]{F2F2F2}70} & \multicolumn{2}{r}{\cellcolor[HTML]{F2F2F2}275} & \multicolumn{2}{r}{\cellcolor[HTML]{F2F2F2}251} & \multicolumn{2}{r}{\cellcolor[HTML]{F2F2F2}31981} & \multicolumn{2}{r}{\cellcolor[HTML]{F2F2F2}5082} & \multicolumn{2}{r}{\cellcolor[HTML]{F2F2F2}38339} & \multicolumn{2}{l}{\cellcolor[HTML]{F2F2F2}\textbf{83.4}} & \textbf{} \\
\textbf{500} & \multicolumn{2}{l}{Grassland} & \multicolumn{2}{r}{1191} & \multicolumn{2}{r}{7} & \multicolumn{2}{r}{25} & \multicolumn{2}{r}{46} & \multicolumn{2}{r}{1039} & \multicolumn{2}{r}{163} & \multicolumn{2}{r}{1532} & \multicolumn{2}{r}{18327} & \multicolumn{2}{r}{22330} & \multicolumn{2}{l}{\textbf{82.1}} & \textbf{} \\
\rowcolor[HTML]{F2F2F2} 
\textbf{} & \multicolumn{2}{l}{\cellcolor[HTML]{F2F2F2}Total} & \multicolumn{2}{r}{\cellcolor[HTML]{F2F2F2}17147} & \multicolumn{2}{r}{\cellcolor[HTML]{F2F2F2}1218} & \multicolumn{2}{r}{\cellcolor[HTML]{F2F2F2}456} & \multicolumn{2}{r}{\cellcolor[HTML]{F2F2F2}698} & \multicolumn{2}{r}{\cellcolor[HTML]{F2F2F2}2187} & \multicolumn{2}{r}{\cellcolor[HTML]{F2F2F2}2242} & \multicolumn{2}{r}{\cellcolor[HTML]{F2F2F2}33851} & \multicolumn{2}{r}{\cellcolor[HTML]{F2F2F2}27013} & \multicolumn{2}{r}{\cellcolor[HTML]{F2F2F2}\textbf{}} & \multicolumn{2}{l}{\cellcolor[HTML]{F2F2F2}\textbf{}} & \textbf{} \\
\textbf{} & \multicolumn{2}{l}{PA (\%)} & \multicolumn{2}{l}{\textbf{86.9}} & \multicolumn{2}{l}{\textbf{63.3}} & \multicolumn{2}{l}{\textbf{28.9}} & \multicolumn{2}{l}{\textbf{30.5}} & \multicolumn{2}{l}{\textbf{9.3}} & \multicolumn{2}{l}{\textbf{29.8}} & \multicolumn{2}{l}{\textbf{94.5}} & \multicolumn{2}{l}{\textbf{67.8}} & \multicolumn{2}{l}{\textbf{}} & \multicolumn{2}{l}{\textbf{}} & \textbf{}
\end{tabular}%
}
\end{table*}

\begin{table*}[ht]
\centering
\footnotesize
\caption{F-score by country for the 6 main crops} 
\label{tab:fscore_country_6maincrops}
\begin{tabular}{lrrrrrrr}
  \hline
Country & \# points & Fscore 211 & Fscore 213 & Fscore 216 & Fscore 222 & Fscore 231 & Fscore 232 \\ 
  \hline
AT & 1750 & 0.63 & 0.54 & 0.70 & 0.46 & 0.50 & 0.96 \\ 
  BE & 946 & 0.69 & 0.51 & 0.72 & 0.65 &  & 0.83 \\ 
  BG & 1751 & 0.68 & 0.27 & 0.71 &  & 0.97 & 0.87 \\ 
  CY & 487 & 0.32 & 0.15 &  &  &  &  \\ 
  CZ & 2084 & 0.78 & 0.65 & 0.85 & 0.98 &  & 0.96 \\ 
  DE & 9170 & 0.67 & 0.62 & 0.85 & 0.75 & 0.22 & 0.90 \\ 
  DK & 1069 & 0.57 & 0.69 & 0.77 & 0.74 &  & 0.87 \\ 
  EE & 666 & 0.46 & 0.57 &  &  &  & 0.70 \\ 
  EL & 2991 & 0.09 & 0.26 & 0.33 &  & 0.31 &  \\ 
  ES & 13242 & 0.53 & 0.60 & 0.42 &  & 0.46 & 0.51 \\ 
  FI & 3183 & 0.18 & 0.54 &  &  &  & 0.41 \\ 
  FR & 15024 & 0.74 & 0.57 & 0.72 & 0.80 & 0.78 & 0.87 \\ 
  HR & 983 & 0.36 &  & 0.74 & 0.22 & 0.84 & 0.13 \\ 
  HU & 1428 & 0.58 & 0.16 & 0.83 & 0.36 & 0.86 & 0.78 \\ 
  IE & 1329 & 0.56 & 0.81 & 0.21 &  &  & 0.67 \\ 
  IT & 6280 & 0.55 & 0.17 & 0.64 & 0.47 & 0.55 & 0.26 \\ 
  LT & 1138 & 0.56 & 0.20 & 0.33 & 0.57 &  & 0.88 \\ 
  LU & 106 & 0.71 & 0.75 & 0.67 &  &  & 1.00 \\ 
  LV & 1313 & 0.67 & 0.28 & 0.12 &  &  & 0.88 \\ 
  MT &  11 &  &  &  &  &  &  \\ 
  NL & 1437 & 0.58 & 0.23 & 0.69 & 0.58 &  &  \\ 
  PL & 7810 & 0.51 & 0.32 & 0.70 & 0.75 &  & 0.88 \\ 
  PT & 2230 & 0.05 & 0.29 & 0.48 &  &  &  \\ 
  RO & 3528 & 0.26 & 0.08 & 0.62 & 0.30 & 0.76 & 0.47 \\ 
  SE & 4756 & 0.48 & 0.50 & 0.16 & 0.40 &  & 0.74 \\ 
  SI & 634 & 0.55 &  & 0.81 &  & 0.67 &  \\ 
  SK & 818 & 0.73 & 0.54 & 0.46 &  & 0.75 & 0.92 \\ 
  UK & 5037 & 0.58 & 0.64 & 0.27 & 0.38 &  & 0.48 \\ 
   \hline
\end{tabular}
\end{table*}

\begin{figure*}[ht]
  \centering 
         \includegraphics[width=0.7\textwidth]{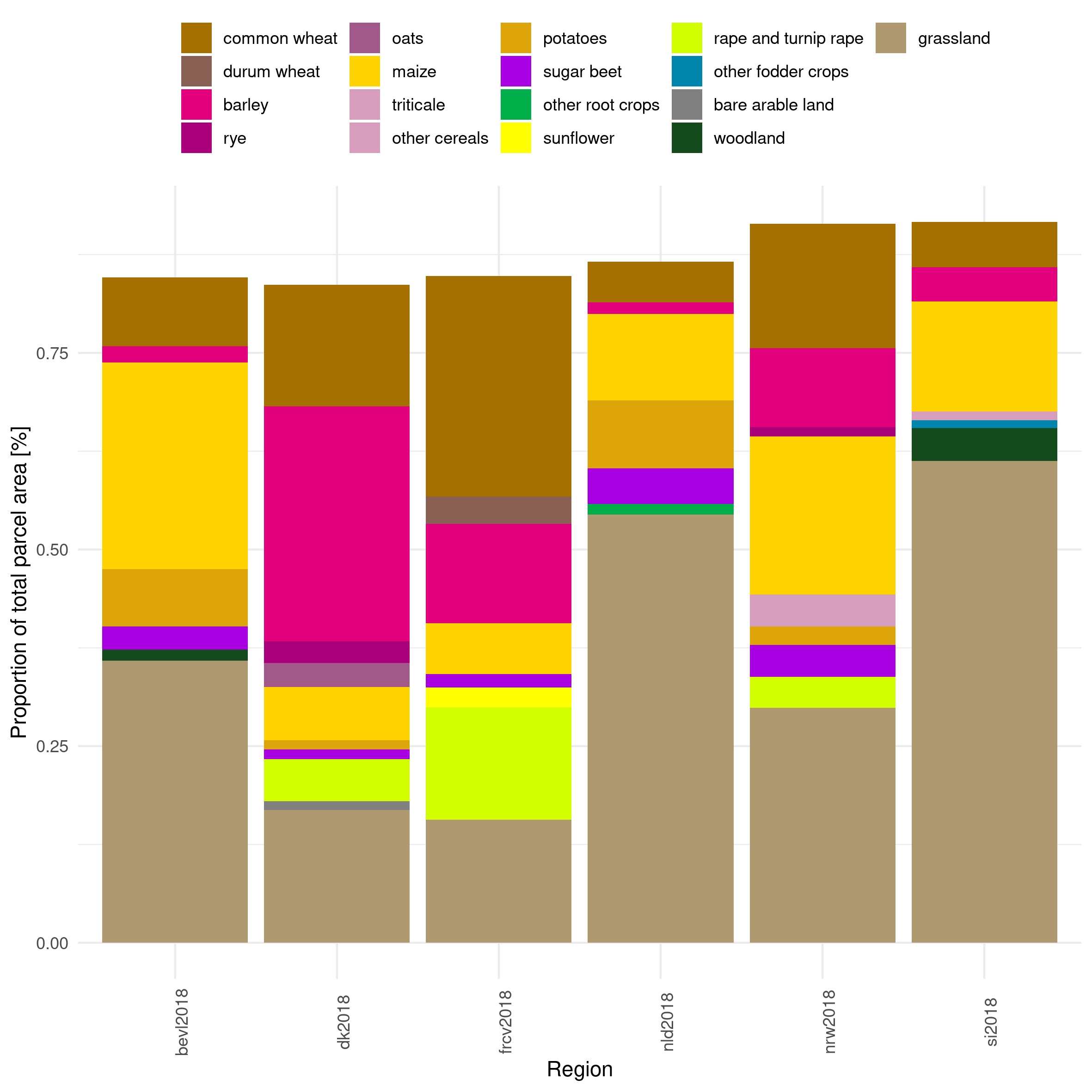}
         \caption{Proportion of total GSAA declared area in 2018 by crop in the six regions of interest. Crop type covering an area smaller than 1\% of the total are neglected. }
         \label{fig:LPIS_distribution_area}
 \end{figure*}

\begin{figure*}[ht]
  \centering 
         \includegraphics[width=0.7\textwidth]{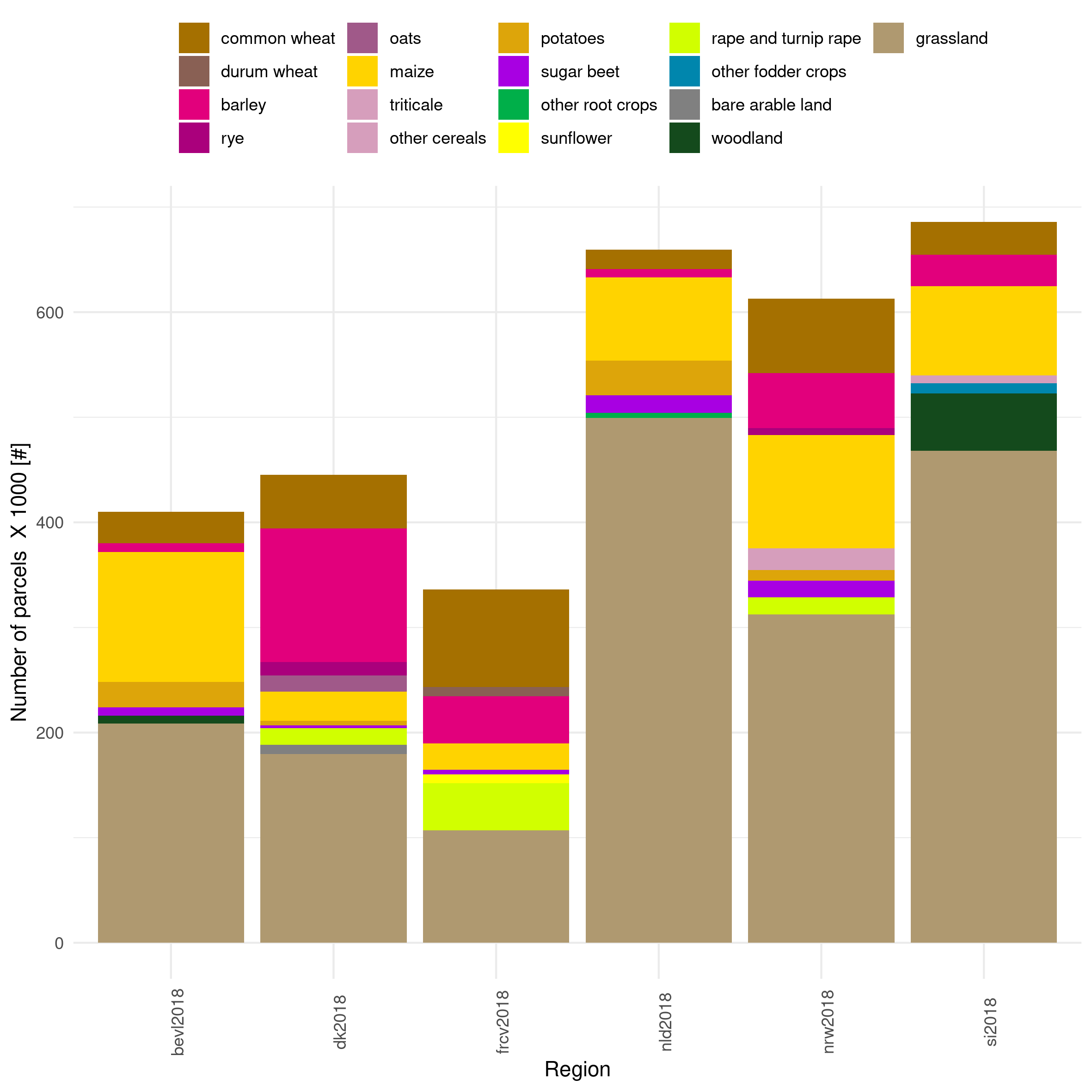}
         \caption{Area of total GSAA declared area in 2018 by crop in the six regions of interest. Crop type covering an area smaller than 1\% of the total are neglected.}
         \label{fig:LPIS_distribution_count}
 \end{figure*}

\begin{figure*}[ht]
  \centering 
         \includegraphics[width=1\textwidth]{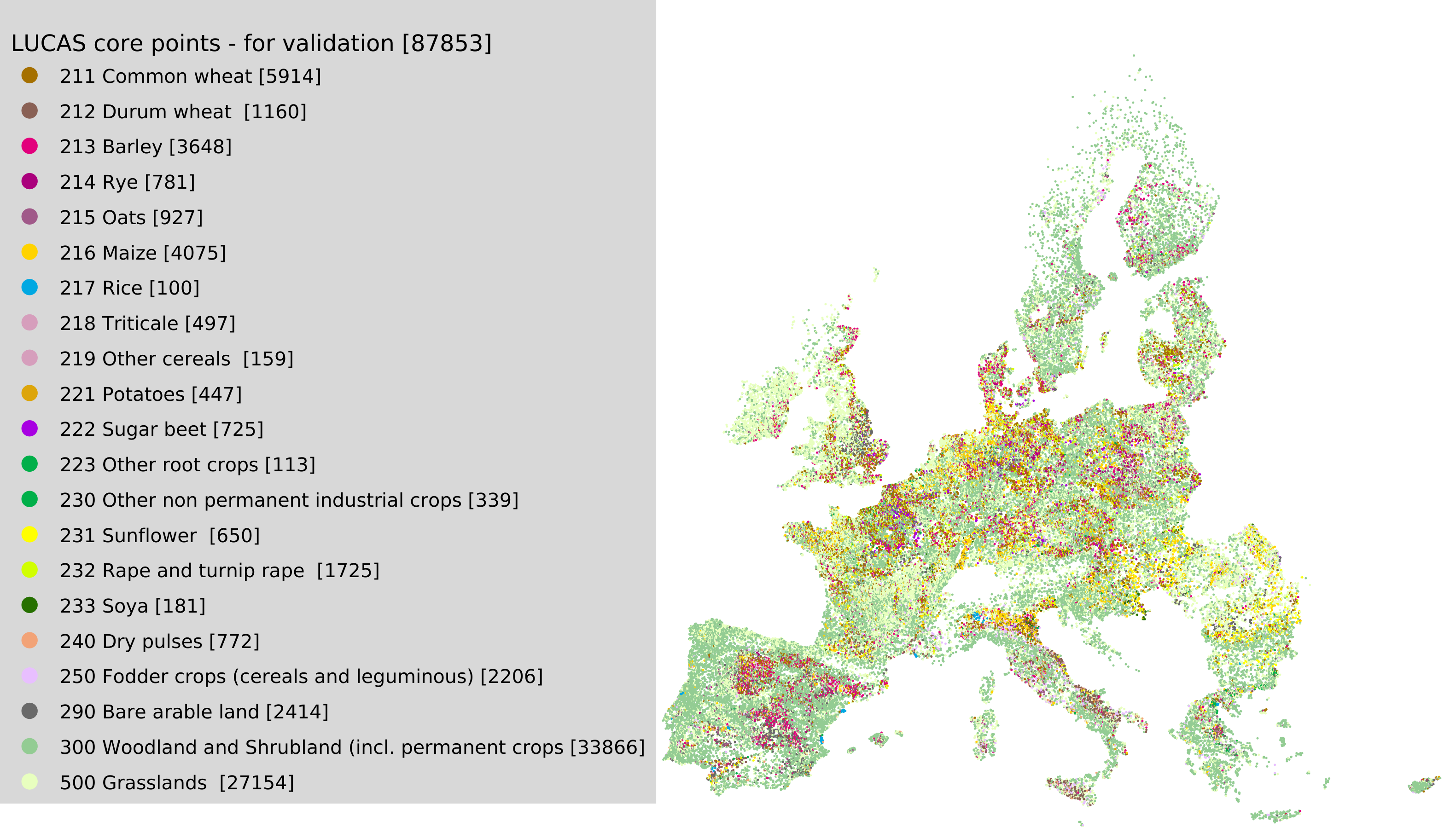}
         \caption{Validation points (87,853) are high-quality point of the LUCAS core survey 2018 not used for training}
         \label{fig:ValidationPOint}
 \end{figure*}

\begin{table*}[ht]
\centering
\caption{Summary of GSAA legend matching data used in the study.} 
\label{tab:GSAA-LUCAS_legendMatching}
\tiny
\begin{tabular}{lllllllllllllllllll}
  \hline
Region & Crop code & 211 & 212 & 213 & 214 & 215 & 216 & 218 & 219 & 221 & 222 & 223 & 231 & 232 & 250 & 290 & 300 & 500 \\ 
  \hline
bevl2018 & hfdtlt & 311 & 0 & 321 & 0 & 0 & 201,202 & 0 & 0 & 901,904 & 91 & 0 & 0 & 0 & 0 & 0 & 9711 & 60,700 \\ 
  dk2018 & afgkode & 2,11 & 0 & 1,10 & 15 & 3 & 216 & 0 & 0 & 151 & 160 & 0 & 0 & 22 & 0 & 318 & 0 & 101,252,254,260,263,276,308 \\ 
  frcv2018 & cropcode & 801 & 1 & 809 & 0 & 0 & 5,6 & 0 & 807 & 0 & 1 & 0 & 1 & 1 & 208 & 0 & 100,699 & 201,203,204 \\ 
  nld2018 & gws\_gewasc & 233 & 0 & 236 & 0 & 0 & 259 & 0 & 0 & 2014,2015,2017 & 256 & 262 & 0 & 0 & 0 & 0 & 0 & 265,266,331,335,336 \\ 
  nrw2018 & mon10\_cr\_1 & 115 & 0 & 131,132 & 121 & 0 & 171,411 & 156 & 0 & 602 & 603 & 0 & 0 & 311 & 0 & 0 & 0 & 424,459 \\ 
  si2018 & sifra\_kmrs & 1 & 10 & 6,7 & 0 & 0 & 4,5 & 0 & 1 & 0 & 189 & 0 & 39 & 37 & 1 & 0 & 2 & 149,150,151 \\ 
   \hline
\end{tabular}
\end{table*}

\begin{table*}[ht]
\footnotesize
\caption{Confusion matrix for bevl2018} 
\label{tab:bevl2018}
\centering
\begin{tabular}{rlllllll}
  \hline
 & 211 & 213 & 216 & 221 & 222 & TOTAL & UA \\ 
  \hline
  211 & 26456 & 118 & 262 & 4 & 6 & 26846 & 0.9855 \\ 
  213 & 1168 & 5401 & 107 & 0 & 1 & 6677 & 0.8089 \\ 
  216 & 734 & 196 & 102799 & 107 & 42 & 103878 & 0.9896 \\ 
  221 & 37 & 10 & 10086 & 11555 & 1332 & 23020 & 0.502 \\ 
  222 & 16 & 7 & 2170 & 982 & 4804 & 7979 & 0.6021 \\ 
  TOTAL & 28411 & 5732 & 115424 & 12648 & 6185 & 168400 &  \\ 
  PA & 0.9312 & 0.9423 & 0.8906 & 0.9136 & 0.7767 &  &  \\ 
   \hline
\end{tabular}

\end{table*}

\begin{table*}[ht]
\caption{Confusion matrix for dk2018} 
\label{tab:dk2018}
\footnotesize
\centering
\begin{tabular}{rlllllllllll}
  \hline
 & 211 & 213 & 214 & 215 & 216 & 221 & 222 & 232 & 290 & TOTAL & UA \\ 
  \hline
211 & 41660 & 3267 & 56 & 7 & 167 & 3 & 0 & 30 & 0 & 45190 & 0.9219 \\ 
  213 & 4593 & 100065 & 103 & 11 & 1679 & 37 & 39 & 97 & 7 & 106631 & 0.9384 \\ 
  214 & 2423 & 2088 & 7524 & 0 & 30 & 1 & 0 & 4 & 0 & 12070 & 0.6234 \\ 
  215 & 1546 & 7060 & 14 & 70 & 386 & 5 & 0 & 15 & 3 & 9099 & 0.0077 \\ 
  216 & 213 & 513 & 3 & 2 & 25374 & 10 & 1 & 1 & 0 & 26117 & 0.9716 \\ 
  221 & 9 & 26 & 0 & 0 & 1788 & 2197 & 21 & 0 & 0 & 4041 & 0.5437 \\ 
  222 & 11 & 52 & 0 & 0 & 679 & 28 & 1929 & 0 & 0 & 2699 & 0.7147 \\ 
  232 & 135 & 312 & 4 & 0 & 66 & 0 & 1 & 14467 & 1 & 14986 & 0.9654 \\ 
  290 & 5 & 8 & 0 & 1 & 20 & 0 & 0 & 0 & 0 & 34 & 0 \\ 
  TOTAL & 50595 & 113391 & 7704 & 91 & 30189 & 2281 & 1991 & 14614 & 11 & 220867 &  \\ 
  PA & 0.8234 & 0.8825 & 0.9766 & 0.7692 & 0.8405 & 0.9632 & 0.9689 & 0.9899 & 0 &  &  \\ 
   \hline
\end{tabular}
\end{table*}

\begin{table*}[ht]
\caption{Confusion matrix for si2018} 
\label{tab:si2018}
\footnotesize
\centering
\begin{tabular}{rlllllll}
  \hline
 & 211 & 213 & 216 & 219 & 250 & TOTAL & UA \\ 
  \hline
211 & 18662 & 155 & 6689 & 0 & 31 & 25537 & 0.7308 \\ 
  213 & 8539 & 4118 & 7979 & 0 & 98 & 20734 & 0.1986 \\ 
  216 & 608 & 31 & 69898 & 0 & 13 & 70550 & 0.9908 \\ 
  219 & 3633 & 72 & 1765 & 0 & 10 & 5480 & 0 \\ 
  250 & 184 & 27 & 1373 & 0 & 23 & 1607 & 0.0143 \\ 
  TOTAL & 31626 & 4403 & 87704 & 0 & 175 & 123908 &  \\ 
  PA & 0.5901 & 0.9353 & 0.797 & NaN & 0.1314 &  &  \\ 
   \hline
\end{tabular}
\end{table*}

\begin{table*}[ht]
\caption{Confusion matrix for frcv2018} 
\label{tab:frcv2018}
\centering
\begin{tabular}{rlllllllll}
  \hline
 & 211 & 212 & 213 & 216 & 222 & 231 & 232 & TOTAL & UA \\ 
  \hline
211 & 88981 & 0 & 233 & 224 & 10 & 10 & 122 & 89580 & 0.9933 \\ 
  212 & 8413 & 0 & 88 & 31 & 3 & 0 & 8 & 8543 & 0 \\ 
  213 & 7186 & 0 & 34060 & 636 & 13 & 9 & 75 & 41979 & 0.8114 \\ 
  216 & 217 & 0 & 49 & 21902 & 10 & 11 & 54 & 22243 & 0.9847 \\ 
  222 & 25 & 0 & 7 & 173 & 3723 & 51 & 33 & 4012 & 0.928 \\ 
  231 & 59 & 0 & 9 & 1317 & 74 & 6768 & 25 & 8252 & 0.8202 \\ 
  232 & 124 & 0 & 24 & 49 & 1 & 11 & 43836 & 44045 & 0.9953 \\ 
  TOTAL & 105005 & 0 & 34470 & 24332 & 3834 & 6860 & 44153 & 218654 &  \\ 
  PA & 0.8474 & - & 0.9881 & 0.9001 & 0.971 & 0.9866 & 0.9928 &  &  \\ 
   \hline
\end{tabular}
\end{table*}

\begin{table*}[ht]
\caption{Confusion matrix for nld2018} 
\label{tab:nld2018}
\footnotesize
\centering
\begin{tabular}{rllllllll}
  \hline
 & 211 & 213 & 216 & 221 & 222 & 223 & TOTAL & UA \\ 
  \hline
211 & 17352 & 148 & 178 & 6 & 4 & 0 & 17688 & 0.981 \\ 
  213 & 1873 & 4224 & 473 & 4 & 3 & 1 & 6578 & 0.6421 \\ 
  216 & 564 & 420 & 73207 & 57 & 14 & 16 & 74278 & 0.9856 \\ 
  221 & 71 & 38 & 7823 & 20472 & 3565 & 9 & 31978 & 0.6402 \\ 
  222 & 32 & 25 & 5229 & 969 & 10416 & 7 & 16678 & 0.6245 \\ 
  223 & 306 & 311 & 2674 & 10 & 108 & 1210 & 4619 & 0.262 \\ 
  TOTAL & 20198 & 5166 & 89584 & 21518 & 14110 & 1243 & 151819 &  \\ 
  PA & 0.8591 & 0.8177 & 0.8172 & 0.9514 & 0.7382 & 0.9735 &  &  \\ 
   \hline
\end{tabular}
\end{table*}

\begin{table*}[ht]
\caption{Confusion matrix for nrw2018} 
\label{tab:nrw2018}
\footnotesize
\centering
\begin{tabular}{rllllllllll}
  \hline
 & 211 & 213 & 214 & 216 & 218 & 221 & 222 & 232 & TOTAL & UA \\ 
  \hline
211 & 69259 & 274 & 164 & 268 & 0 & 4 & 15 & 45 & 70029 & 0.989 \\ 
  213 & 10833 & 37528 & 220 & 556 & 0 & 1 & 7 & 125 & 49270 & 0.7617 \\ 
  214 & 1455 & 754 & 3918 & 48 & 0 & 1 & 1 & 17 & 6194 & 0.6325 \\ 
  216 & 824 & 135 & 43 & 102231 & 0 & 17 & 23 & 61 & 103334 & 0.9893 \\ 
  218 & 12692 & 2064 & 4921 & 92 & 8 & 2 & 0 & 42 & 19821 & 4e-04 \\ 
  221 & 96 & 15 & 2 & 2204 & 0 & 4365 & 2564 & 29 & 9275 & 0.4706 \\ 
  222 & 41 & 13 & 0 & 2039 & 0 & 231 & 13238 & 5 & 15567 & 0.8504 \\ 
  232 & 94 & 17 & 0 & 28 & 0 & 0 & 0 & 16175 & 16314 & 0.9915 \\ 
  TOTAL & 95294 & 40800 & 9268 & 107466 & 8 & 4621 & 15848 & 16499 & 289804 &  \\ 
  PA & 0.7268 & 0.9198 & 0.4227 & 0.9513 & 1 & 0.9446 & 0.8353 & 0.9804 &  &  \\ 
   \hline
\end{tabular}
\end{table*}

\begin{table*}[ht]
\caption{Legend convergence between Eurostat and EU crop map classes} 
\label{tab:legend_convergence_Estat}
\centering
\begin{tabular}{llrl}
  \hline
estat\_crop\_class & estat\_crop\_label & eucropmap\_class & eucropmap\_label \\ 
  \hline
C1110 & Common wheat and spelt & 211 & common wheat \\ 
  C1120 & Durum wheat & 212 & durum wheat \\ 
  C1200 & Rye and winter cereal mixtures (maslin) & 214 & rye \\ 
  C1300 & Barley & 213 & barley \\ 
  C1410 & Oats & 215 & oats \\ 
  C1500 & Grain maize and corn-cob-mix & 216 & maize \\ 
  C1600 & Triticale & 218 & triticale \\ 
  C2000 & Rice & 217 & rice \\ 
  G3000 & Green maize & 216 & maize \\ 
  R1000 & Potatoes (including seed potatoes) & 221 & potatoes \\ 
  R2000 & Sugar beet (excluding seed) & 222 & sugar beet \\ 
  I1110 & Rape and turnip rape seeds & 232 & rape and turnip rape \\ 
  I1120 & Sunflower seed & 231 & sunflower \\ 
  I1130 & Soya & 233 & soya \\ 
   \hline
\end{tabular}
\end{table*}

\begin{figure*}[ht]
  \centering 
         \includegraphics[width=\textwidth]{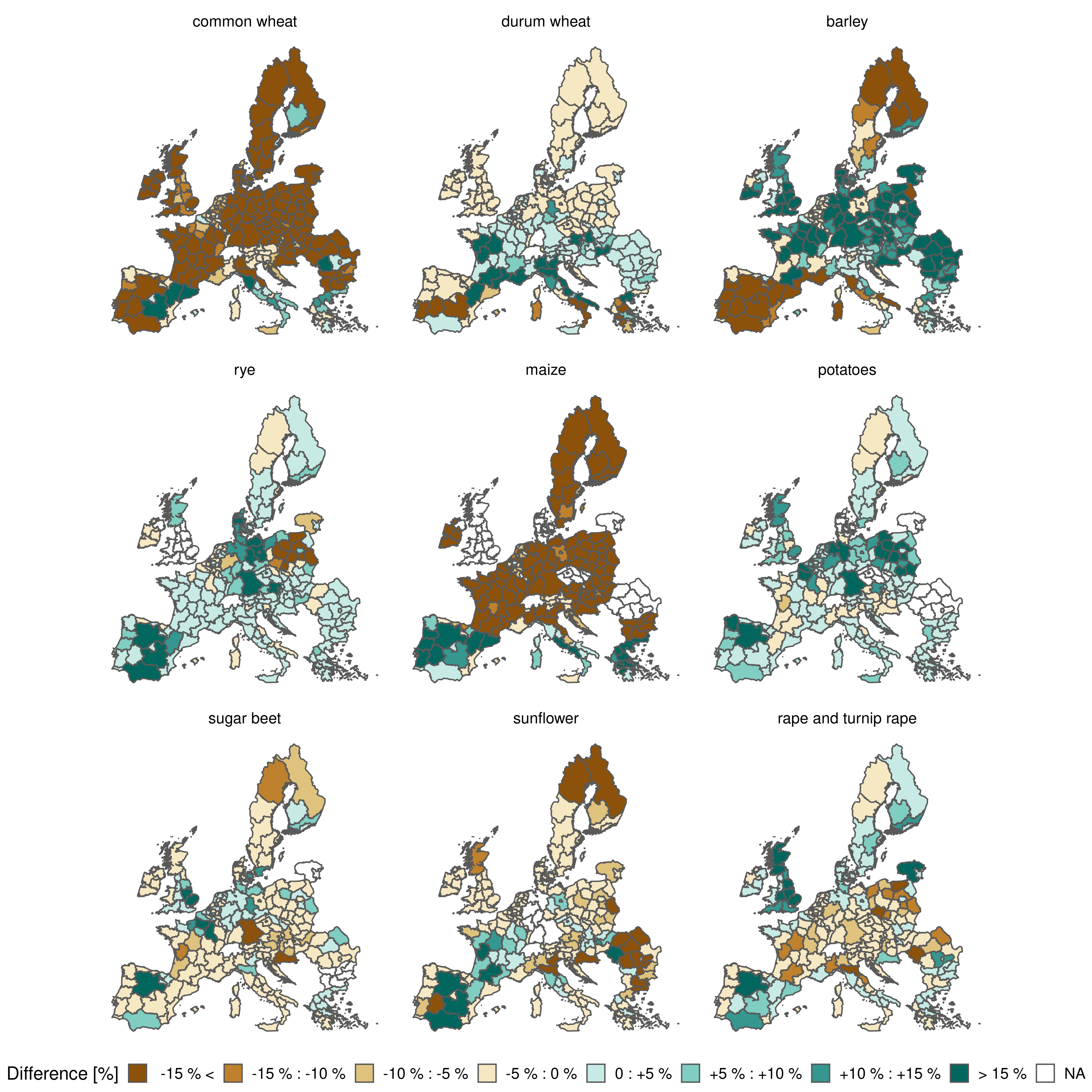}
         \caption{Relative difference of area [\%] reported by Eurostat and retrieved from the EU crop map. The difference is the Eurostat reported value minus the value estimated from the EU crop map. Green values means that the EU crop map underestimates the area compared to the data reported from Eurostat while the brown values means that there is an overestimation.}
         \label{fig:Nuts2Comparison_map}
 \end{figure*}

\begin{figure*}[ht]
  \centering 
         \includegraphics[width=\textwidth]{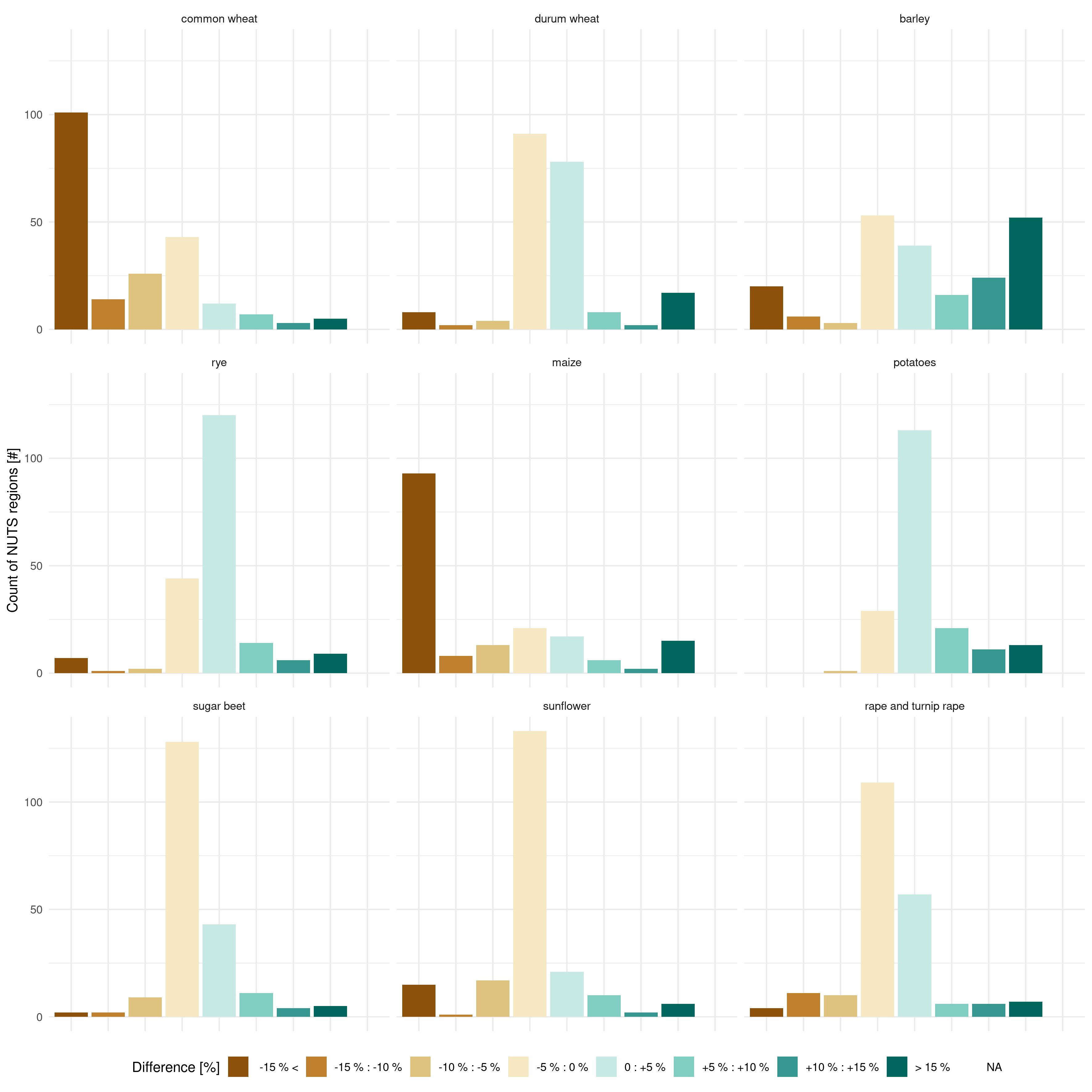}
         \caption{The areas reported by Eurostat at NUTS2 level (except for UK and DE at NUTS1) are compared with the area retrieved from the EUcropmap. The difference is the Eurostat reported value minus the value estimated from the EU crop map. Green values means that the EU crop map underestimates the area compared to the data reported from Eurostat while the brown values means that there is an overestimation.}
         \label{fig:Nuts2Comparison_hist}
 \end{figure*}

\begin{figure*}[ht]
  \centering 
         \includegraphics[width=0.7\textwidth]{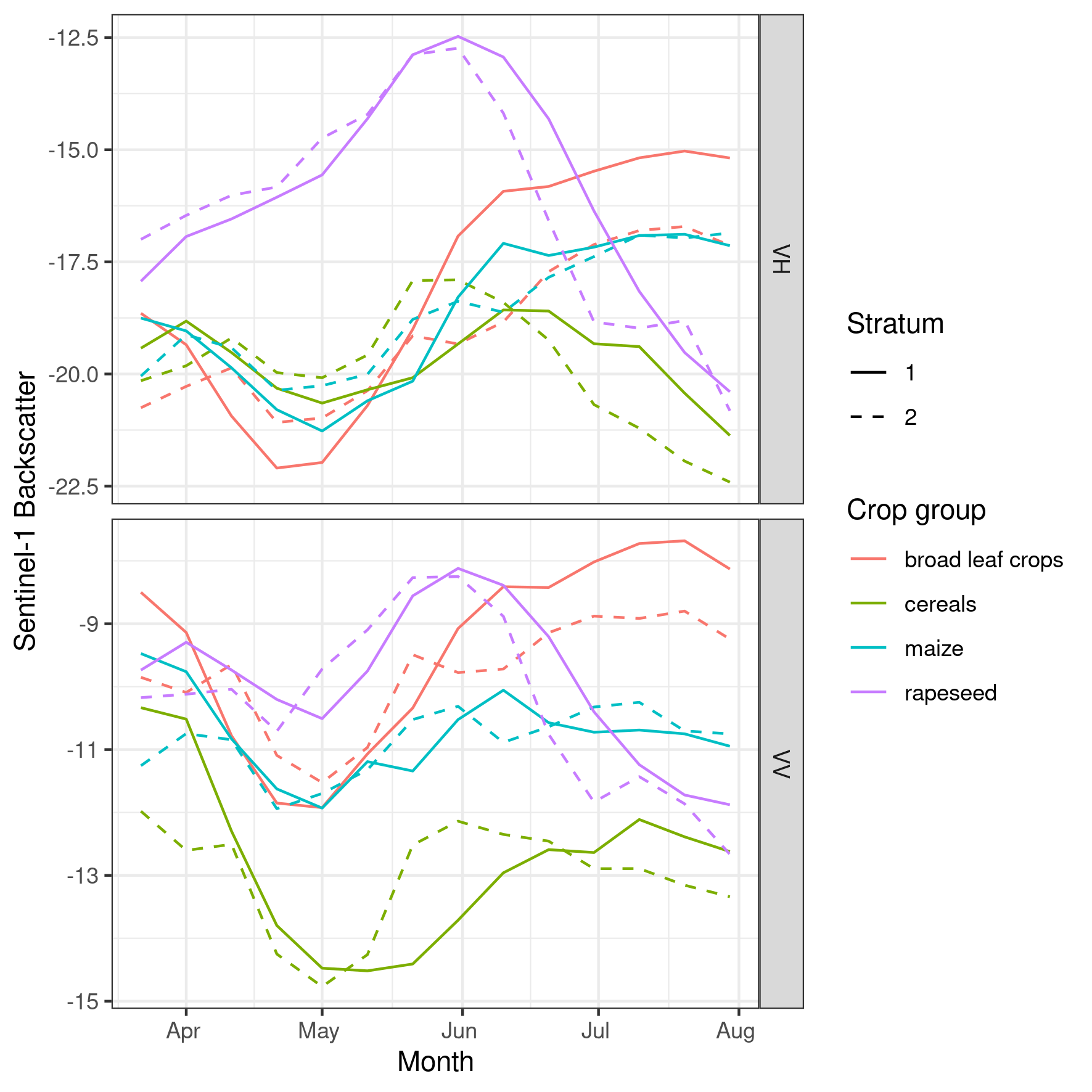}
         \caption{Sentinel-1 backscattering coefficient VV and VH are grouped for broad leaf crops (potato, sugar beet, sunflower), cereals (common wheat, durum wheat, barley, rye, oats, triticale), maize and rapeseed. This grouping better illustrates the differences observed in the north (stratum 1) and in the south (stratum 2). }
         \label{fig:S1TScropGroupStratumComparison}
 \end{figure*}

\end{document}